\documentclass[twoside,11pt]{article}
\usepackage{jmlr2e}
\usepackage{microtype}
\usepackage{amsmath}
\usepackage{amssymb}
\usepackage{bm}
\usepackage{graphics}
\usepackage{graphicx}
\usepackage{multirow}
\usepackage{psfrag}
\usepackage{epsfig}
\usepackage{subfigure}
\usepackage{wrapfig}
\usepackage[ruled]{algorithm2e}
\usepackage[usenames]{color}
\usepackage{tikz}
\usetikzlibrary{arrows,shapes,snakes,automata,backgrounds,fit,petri}
\definecolor{myred}{rgb}{0.38,0.08,0.12}
\definecolor{mygray}{rgb}{0.1,0.1,0.1}
\definecolor{imp2}{rgb}{1.0000,0.7647,0.4824}
\definecolor{imp3}{rgb}{0.0471,0.2549,0.235}
\definecolor{imp4}{rgb}{0.7059,0.1647,0.001}
\usepackage[colorlinks=true, urlcolor=myred, citecolor=myred, linkcolor=myred]{hyperref}
\def\B{{\bf B}}

\def\C{{\bf C}}

\def\D{{\bf D}}

\def\H{{\bf H}}

\def\I{{\bf I}}

\def\K{{\bf K}}

\def\L{{\bf L}}

\def\M{{\bf M}}
\def\m{{\bf m}}

\def\P{{\bf P}}

\def\Q{{\bf Q}}

\def\R{{\bf R}}

\def\S{{\bf S}}

\def\U{{\bf U}}
\def\u{{\bf u}}
\def\V{{\bf V}}

\def\W{{\bf W}}

\def\X{{\bf X}}
\def\x{{\bf x}}
\def\Y{{\bf Y}}
\def\y{{\bf y}}
\def\Z{{\bf Z}}
\def\z{{\bf z}}

\def\PL{{\bf P}_\mathrm{r}}

\def\PC{{\bf P}_\mathrm{f}}

\def\VR{{\cal V}}

\def\0{{\bf 0}}
\def\1{{\bf 1}}

\def\Lik{\mathcal{L}}

\def\<{\, \langle \,}
\def\>{\, \rangle \,}
\def\inv{^{-1}}
\def\ts{^\top}

\def\topc{\mathrm{top}}

\def\sign{\mathrm{sign}}

\def\min{\mathrm{min}}

\def\diag{\mathrm{diag}}

\def\FM{\mathrm{FM}}
\def\DAG{\mathrm{DAG}}

\def\rep{\mathrm{rep}}

\def\bPsi{\bm{\Psi}}
\def\bpi{\bm{\pi}}
\def\bSigma{\bm{\Sigma}}
\def\bepsilon{\bm{\epsilon}}

\def\bupsilon{\bm{\upsilon}}

\def\bTheta{\bm{\Theta}}

\def\DN{\mathcal{N}}

\def\IG{\mathrm{IG}}
\def\La{\mathrm{Laplace}}
\def\Exp{\mathrm{Exponential}}

\def\Be{\mathrm{Beta}}

\def\Ber{\mathrm{Bernoulli}}
\def\Ga{\mathrm{Gamma}}

\def\GP{\mathrm{GP}}

\def\MOD{\mathcal{M}}

\newcommand{\argmax}{\operatornamewithlimits{argmax}}

\newcommand{\deff}{\overset{\underset{\mathrm{def}}{}}{=}}

\newcommand{\ie}{i.e.\ }
\newcommand{\eg}{e.g.\ }

\newcommand{\iid}{i.i.d.\ }

\newcommand{\quotes}[1]{``#1''}

\def\ctilde{\kern -.04em\lower .7ex\hbox{\~{}}\kern .04em}

\jmlrheading{12}{2011}{663-705}{10/09; Revised 10/10}{3/11}{Ricardo Henao and Ole Winther}
\ShortHeadings{Sparse Linear Identifiable Multivariate Modeling}{Henao and Winther} \firstpageno{663}
\begin{document}
\title{Sparse Linear Identifiable Multivariate Modeling}
\author{\name Ricardo Henao \email  rhenao@binf.ku.dk \\
        \name Ole Winther \email owi@imm.dtu.dk \\
        \addr DTU Informatics \\
              Richard Petersens Plads, Building 321 \\
              Technical University of Denmark \\
              DK-2800 Lyngby, Denmark
        \AND
        \addr Bioinformatics Centre \\
              University of Copenhagen \\
              Ole Maaloes Vej 5 \\
              DK-2200 Copenhagen N, Denmark}
\editor{Aapo Hyv\"{a}rinen}
\maketitle
\begin{abstract}
In this paper we consider sparse and identifiable linear latent variable (factor) and linear Bayesian network models for parsimonious analysis of multivariate data. We propose a computationally efficient method for joint parameter and model inference, and model comparison. It consists of a fully Bayesian hierarchy for sparse models using slab and spike priors (two-component $\delta$-function and continuous mixtures), non-Gaussian latent factors and a stochastic search over the ordering of the variables. The framework, which we call SLIM (Sparse Linear Identifiable Multivariate modeling), is validated and bench-marked on artificial and real biological data sets. SLIM is closest in spirit to LiNGAM \citep{shimizu06}, but differs substantially in inference, Bayesian network structure learning and model comparison. Experimentally, SLIM performs equally well or better than LiNGAM with comparable computational complexity. We attribute this mainly to the stochastic search strategy used, and to parsimony (sparsity and identifiability), which is an explicit part of the model. We propose two extensions to the basic \iid linear framework: non-linear dependence on observed variables, called SNIM (Sparse Non-linear Identifiable Multivariate modeling) and allowing for correlations between latent variables, called CSLIM (Correlated SLIM), for the temporal and/or spatial data. The source code and scripts are available from \url{http://cogsys.imm.dtu.dk/slim/}.
\end{abstract}
\section{Introduction}
Modeling and interpretation of multivariate data are central themes in machine learning. Linear latent variable models (or factor analysis) and linear directed acyclic graphs (DAGs) are prominent examples of models for continuous multivariate data. In factor analysis, data is modeled as a linear combination of independently distributed factors thus allowing for capture of a rich underlying co-variation structure. In the DAG model, each variable is expressed as regression on a subset of the remaining variables with the constraint that total connectivity is acyclic in order to have a properly defined joint distribution. Parsimonious (interpretable) modeling, using sparse factor loading matrix or restricting the number of parents of a node in a DAG, are good prior assumptions in many applications. Recently, there has been a great deal of interest in detailed modeling of sparsity in factor models, for example in the context of gene expression data analysis \citep{west03,lucas06,knowles07,thibaux07,carvalho08,rai08}. Sparsity arises for example in gene regulation because the latent factors represent driving signals for gene regulatory sub-networks and/or transcription factors, each of which only includes/affects a limited number of genes. A parsimonious DAG is particularly attractable from an interpretation point of view but the restriction to only having observed variables in the model may be a limitation because one rarely measures all relevant variables. Furthermore, linear relationships might be unrealistic for example in gene regulation, where it is generally accepted that one cannot replace the driving signal (related to concentration of a transcription factor protein in the cell nucleus) with the measured concentration of corresponding mRNA. Bayesian networks represent a very general class of models, encompassing both observed and latent variables. In many situations it will thus be relevant to learn parsimonious Bayesian networks with both latent variables and a non-linear DAG parts. Although attractive, by being closer to what one may expect in practice, such modeling is complicated by difficult inference (\citet{chickering96} showed that DAG structure learning is NP-hard) and by potential non-identifiability. Identifiability means that each setting of the parameters defines a unique distribution of the data. Clearly, if the model is not identifiable in the DAG and latent parameters, this severely limits the interpretability of the learned model.

\citet{shimizu06} provided the important insight that every DAG has a factor model representation, \ie the connectivity matrix of a DAG gives rise to a triangular mixing matrix in the factor model. This provided the motivation for the Linear Non-Gaussian Acyclic Model (LiNGAM) algorithm which solves the identifiable factor model using Independent Component Analysis \citep[ICA,][]{hyvarinen01} followed by iterative permutation of the solutions towards triangular, aiming to find a suitable ordering for the variables. As final step, the resulting DAG is pruned based on different statistics, \eg Wald, Bonferroni, $\chi^2$ second order model fit tests. Model selection is then performed using some pre-chosen significance level, thus LiNGAM select from models with different sparsity levels and a fixed deterministically found ordering. There is a possible number of extensions to their basic model, for instance \citet{hoyer08a} extend it to allow for latent variables, for which they use a probabilistic version of ICA to obtain the variable ordering, pruning to make the model sparse and bootstrapping for model selection. Although the model seems to work well in practice, as commented by the authors, it is restricted to very small problems (3 or 4 observed and 1 latent variables). Non-linear DAGs are also a possibility, however finding variable orderings in this case is known to be far more difficult than in the linear case. These methods inspired by \citet{friedman00}, mainly consist of two steps: performing non-linear regression for a set of possible orderings, and then testing for independence to prune the model, see for instance \citet{hoyer08} and \citet{zhang09a}. For tasks where exhaustive order enumeration is not feasible, greedy approaches like DAG-search \citep[see \quotes{ideal parent} algorithm,][]{elidan07} or PC \citep[Prototypical Constraint, see kernel PC,][]{tillman09} can be used as computationally affordable alternatives.

Factor models have been successfully employed as exploratory tools in many multivariate analysis applications. However, interpretability using sparsity is usually not part of the model, but achieved through post-processing. Examples of this include, bootstrapping, rotating the solutions to maximize sparsity (varimax, procrustes), pruning or thresholding. Another possibility is to impose sparsity in the model through $L_1$ regularization to obtain a maximum a-posteriori estimate \citep{jolliffe03,zou06}. In fully Bayesian sparse factor modeling, two approaches have been proposed: parametric models with bimodal sparsity promoting priors \citep{west03,lucas06,carvalho08,henao09}, and non-parametric models where the number of factors is potentially infinite \citep{knowles07,thibaux07,rai08}. It turns out that most of the parametric sparse factor models can be seen as finite versions of their non-parametric counterparts, for instance \citet{west03} and \citet{knowles07}. The model proposed by \citet{west03} is, as far as the authors know, the first attempt to encode sparsity in a factor model explicitly in the form of a prior. The remaining models improve the initial setting by dealing with the optimal number of factors in \citet{knowles07}, improved hierarchical specification of the sparsity prior in \citet{lucas06,carvalho08,thibaux07}, hierarchical structure for the loading matrices in \citet{rai08} and identifiability without restricting the model in \citet{henao09}.

Many algorithms have been proposed to deal with the NP-hard DAG structure learning task. LiNGAM, discussed above, is the first fully identifiable approach for continuous data. All other approaches for continuous data use linearity and (at least implicitly) Gaussianity assumptions so that the model structure learned is only defined up to equivalence classes. Thus in most cases the directionality information about the edges in the graph must be discarded. Linear Gaussian-based models have the added advantage that they are computationally affordable for the many variables case. The structure learning approaches can be roughly divided into stochastic search and score \citep{cooper92,heckerman00,friedman03}, constraint-based (with conditional independence tests) \citep{spirtes01} and two stage; like LiNGAM,  \citep{tsamardinos06,friedman99,teyssier05,schmidt07,shimizu06}. In the following, we discuss in more detail previous work in the last category, as it is closest to the work in this paper and can be considered representative of the state-of-the-art. The Max-Min Hill-Climbing algorithm \citep[MMHC,][]{tsamardinos06} first learns the skeleton using conditional independence tests similar to PC algorithms \citep{spirtes01} and then the order of the variables is found using a Bayesian-scoring hill-climbing search. The Sparse Candidate (SC) algorithm \citep{friedman99} is in the same spirit but restricts the skeleton to within a predetermined link candidate set of bounded size for each variable. The Order Search algorithm \citep{teyssier05} uses hill-climbing first to find the ordering, and then looks for the skeleton with SC. $L_1$ regularized Markov Blanket \citep{schmidt07} replaces the skeleton learning from MMHC with a dependency network \citep{heckerman00} written as a set of local conditional distributions represented as regularized linear regressors. Since the source of identifiability in Gaussian DAG models is the direction of the edges in the graph, a still meaningful approach consists of entirely focusing on inferring the skeleton of the graph by keeping the edges undirected as in \citet{dempster72,dawid93,giudici99,rajaratnam08}.
\begin{figure}[th]
	\centering
	\begin{tikzpicture}[ bend angle = 45, >= latex, font = \fontsize{6}{6}\selectfont ]

		\tikzstyle{omat} = [ rectangle, draw = black!100, fill = imp2, minimum width = 6mm, minimum height = 6mm, inner sep = 2pt ]
		\tikzstyle{umat} = [ rectangle, draw = black!100, fill = black!0, minimum width = 6mm, minimum height = 6mm, inner sep = 2pt]
		\tikzstyle{bmat} = [ rectangle, draw = black!100, fill = black!10, minimum width = 40mm, minimum height = 8mm, inner sep = 2pt, thick ]
		\tikzstyle{epty} = [ draw = black!100, fill = black!0, minimum width = 50mm, minimum height = 16mm, inner sep = 2pt]
		\tikzstyle{obs} = [ circle, thick, draw = black!80, fill = imp2, minimum size = 1mm, inner sep = 1pt]
		\tikzstyle{lat} = [ circle, thick, draw = black!80, fill = black!0, minimum size = 1mm, inner sep = 1pt]
		\tikzstyle{dmy} = [ circle, thick, draw = black!0, fill = black!0, minimum size = 1mm, inner sep = 1pt]
		\tikzstyle{tbox} = [ rectangle, draw = black!100, fill = black!0, minimum width = 6mm, minimum height = 3mm, inner sep = 2pt ]
		
		\tikzstyle{every label} = [black!100]
		%
		\begin{scope}[ node distance = 8mm, xshift = 85mm, yshift = -13mm ]
			\node [obs] (x1)  [] {$x_1$};
			\node [obs] (x2) [ below of = x1 ] {$x_2$};
			\node [dmy] (x3) [ below of = x2, node distance = 6mm ] {$\vdots$};
			\node [obs] (x4) [ below of = x3 ] {$x_d$};
			\node [lat] (z1) [ right of = x1, xshift = 8mm] {$z_1$}
				edge [post] (x1)
				edge [post] (x2)
				edge [post] (x3)
				edge [post] (x4);
			\node [lat] (z2) [ right of = x2, xshift = 8mm] {$z_2$}
				edge [post] (x2)
				edge [post] (x3)
				edge [post] (x4);
			\node [dmy] (z3) [ right of = x3, xshift = 8mm] {$\vdots$};
			\node [lat] (z4) [ right of = x4, xshift = 8mm] {$z_m$}
				edge [post] (x2)
				edge [post] (x4);
		\end{scope}
		\begin{scope}[ node distance = 8mm, xshift = 85mm, yshift = -4mm ]
			\node [obs] (x1) [] {$x_1$};
			\node [obs] (x2) [ above of = x1 ] {$x_2$}
				edge [pre] (x1);
			\node [obs] (x3) [ right of = x1 ] {$x_3$}
				edge [pre] (x1)
				edge [pre] (x2);
			\node [dmy] (x4) [ above of = x3 ] {$\ldots$}
				edge [pre] (x2);
			\node [obs] (x5) [ right of = x4 ] {$x_d$}
				edge [pre] (x3)
				edge [pre] (x4);
		\end{scope}
		%
		\begin{scope}[ node distance = 8mm, xshift = 85mm, yshift = 13mm ]
			\node [lat] (l1) [] {$l_1$};
			\node [obs] (x1) [ above of = l1 ] {$x_1$}
				edge [pre] (l1);
			\node [obs] (x2) [ right of = l1 ] {$x_2$}
				edge [pre] (l1);
			\node [dmy] (x3) [ above of = x2 ] {$\ldots$}
				edge [pre] (x1);
			\node [obs] (x4) [ right of = x3 ] {$x_d$}
				edge [pre] (x2)
				edge [pre] (x3);
			\node [obs] (x5) [ above of = x3 ] {$x_3$}
				edge [pre] (x3)
				edge [post] (x4);
			\node [lat] (l3) [ below of = x4 ] {$l_m$}
				edge [post] (x2)
				edge [post] (x4);
			\node [lat] (l2) [ above of = x1 ] {$l_2$}
				edge [post] (x3)
				edge [post] (x5);
		\end{scope}
		\begin{scope}[node distance = 7mm,->]
			\node [omat] (X)  [] {$\{\X,\X^\star\}$};
			\node [umat] (P)  [ right of = X, node distance = 14mm ] {$\P$};
				\path (X.east) edge (P.west);
				\path (X)+(42mm,0mm) node (FM) [bmat, label = 90:{linear Bayesian network} ] {};
				\path (P.east) edge (FM.west);
			\node [umat] (B)  [ right of = FM, xshift = -23mm ] {$\B$};
			\node [omat] (X)  [ right of = B ] {$\X$};
			\node [ right of = X, node distance = 4.5mm ] (plus1) {$+$};
			\node [umat] (C)  [ right of = plus1, node distance = 4.5mm ] {$\C$};
			\node [umat] (Z)  [ right of = C ] {$\Z$};
			\node [ right of = Z, node distance = 4.5mm ] (plus2) {$+$};
			\node [umat] (E)  [ right of = plus2, node distance = 4.5mm ] {$\bepsilon$};
			\node [ right of = E, xshift = 18mm, yshift = 20mm ] (dmy1) {};
			\node [ right of = E, xshift = 18mm, yshift = 0mm ] (dmy2) {};
			\node [ right of = E, xshift = 18mm, yshift = -20mm ] (dmy3) {};
				\path (FM.east)+(0,2mm) edge node [fill=black!0] {$\langle\B\rangle,\langle\C_L\rangle$} (dmy1);
				\path (FM.east) edge node [fill=black!0] {$\langle\B\rangle$} (dmy2);
				\path (FM.east)+(0,-2mm) edge node [fill=black!0] {$\langle\C_L\rangle$} (dmy3);
			\node [ below of = FM, yshift = -2mm, xshift = -16mm ] (dmy5) {};
				\path (FM.south)+(-16mm,0) edge (dmy5);
			\node [ below of = P, yshift = -2mm, xshift = 0mm ] (dmy6) {};
				\path (dmy6) edge (P.south);
			\node [ below of = FM, yshift = -2mm, xshift = 7mm ] (dmy7) {};
				\path [pre] (FM.south)+(7mm,0) edge (dmy7);
			\node [tbox] (sslim) [ below of = FM, yshift = -16.5mm, xshift = 7mm ] {CSLIM};
			\node [tbox] (slim) [ below of = FM, yshift = -5.5mm, xshift = 7mm ] {SLIM};
			\node [tbox] (snim) [ below of = FM, yshift = -11mm, xshift = 7mm ] {SNIM};
			\node [ right of = dmy1, xshift = 14mm ] (dmy8) {};
			\node [ right of = dmy2, xshift = 14mm ] (dmy9) {};
			\node [ right of = dmy3, xshift = 14mm ] (dmy10) {};
			\node [ right of = dmy9, xshift = 6.5mm ] (dmy11) {};
				\path (dmy8) edge node [fill=black!0] {$\langle\Lik\rangle$} ([yshift=2mm]dmy11);
				\path (dmy9) edge node [fill=black!0] {$\langle\Lik\rangle$} ([xshift=0.2mm]dmy11);
				\path (dmy10) edge node [fill=black!0] {$\langle\Lik\rangle$} ([yshift=-2mm]dmy11);
		\end{scope}
		%
		\begin{scope}[ node distance = 8mm, xshift = 15mm, yshift = -20mm, , font = \fontsize{5}{5}\selectfont ]
			\draw[ -> ] (0mm,0mm) -- coordinate (x axis mid) (12mm,0);
			\draw[ -> ] (0mm,0mm) -- coordinate (y axis mid) (0mm,10mm);
				\node[ below, xshift = -0.5mm, yshift = 0.5mm ] at (x axis mid) {candidates};
				\node[ left, rotate = 90, yshift = 1.5mm, xshift = 3.5mm ] at (y axis mid) {usage};
			\draw[ color = imp3, thick ] (1mm,0mm) -- (1mm,8mm);
			\draw[ color = imp3, thick ] (2mm,0mm) -- (2mm,6mm);
			\draw[ color = imp3, thick ] (3mm,0mm) -- (3mm,4mm);
			\draw[ color = imp3, thick ] (4mm,0mm) -- (4mm,3mm);
			\draw[ color = imp3, thick ] (5mm,0mm) -- (5mm,2mm);
			\draw[ color = imp4, thick ] (6mm,0mm) -- (6mm,1mm);
			\draw[ color = imp4 ] (7mm,0.5mm) -- (7mm,0.6mm);
			\draw[ color = imp4 ] (8mm,0.5mm) -- (8mm,0.6mm);
			\draw[ color = imp4 ] (9mm,0.5mm) -- (9mm,0.6mm);
			\draw[ color = imp4, thick ] (10mm,0mm) -- (10mm,1mm);
			\filldraw [line width = 0.5pt, draw = black!80, fill = black!0, fill opacity = 0.1, rounded corners = 2mm, dashed ] (20mm,12mm) rectangle (-10mm,-8mm);
			\draw[ color = black!100 ] (5.0mm,-5mm) node[ rotate = 0 ] {$\{\P^{(1)},\ldots,\P^{(m_\topc)},\ldots\}$};
			\draw[ color = black!100 ] (5.0mm,-9.5mm) node[ rotate = 0 ] {Order search};
		\end{scope}
		%
		\begin{scope}[ node distance = 8mm, xshift = 49mm, yshift = -20mm, font = \fontsize{5}{5}\selectfont ]
			\filldraw [line width = 0.5pt, draw = black!80, fill = black!0, fill opacity = 0.1, rounded corners = 2mm, dashed ] (8mm,12mm) rectangle (-8mm,-8mm);
			\draw[ color = black!100 ] (0.0mm,-9.5mm) node[ rotate = 0 ] {Latent variables};
		\end{scope}
		\begin{scope}[ node distance = 8mm, xshift = 124mm, yshift = -2.5mm, font = \fontsize{5}{5}\selectfont ]
			\draw[ -> ] (0mm,0mm) -- coordinate (x axis mid) (12mm,0);
			\draw[ -> ] (0mm,0mm) -- coordinate (y axis mid) (0mm,10mm);
				\node[ below, xshift = -0.5mm, yshift = 0.5mm ] at (x axis mid) {likelihood};
				\node[ left, rotate = 90, yshift = 1.5mm, xshift = 3.5mm ] at (y axis mid) {density};
			\draw[ color = imp3, thick ] (1mm,0mm) -- (1mm,0.5mm);
			\draw[ color = imp3, thick ] (2mm,0mm) -- (2mm,1mm);
			\draw[ color = imp3, thick ] (3mm,0mm) -- (3mm,4mm);
			\draw[ color = imp3, thick ] (4mm,0mm) -- (4mm,8mm);
			\draw[ color = imp3, thick ] (5mm,0mm) -- (5mm,4mm);
			\draw[ color = imp3, thick ] (6mm,0mm) -- (6mm,1mm);
			\draw[ color = imp3, thick ] (7mm,0mm) -- (7mm,1mm);
			\draw[ color = imp3, thick ] (8mm,0mm) -- (8mm,0.5mm);
			\draw[ color = imp3, thick ] (9mm,0mm) -- (9mm,0mm);
			\draw[ color = imp3, thick ] (10mm,0mm) -- (10mm,0mm);
			\draw[ color = imp4, thick ] (2.5mm,0mm) -- (2.5mm,0.5mm);
			\draw[ color = imp4, thick ] (3.5mm,0mm) -- (3.5mm,1mm);
			\draw[ color = imp4, thick ] (4.5mm,0mm) -- (4.5mm,2mm);
			\draw[ color = imp4, thick ] (5.5mm,0mm) -- (5.5mm,6mm);
			\draw[ color = imp4, thick ] (6.5mm,0mm) -- (6.5mm,8mm);
			\draw[ color = imp4, thick ] (7.5mm,0mm) -- (7.5mm,3mm);
			\draw[ color = imp4, thick ] (8.5mm,0mm) -- (8.5mm,1mm);
			\draw[ color = imp4, thick ] (9.5mm,0mm) -- (9.5mm,0.5mm);
			\filldraw [line width = 0.5pt, draw = black!80, fill = black!0, fill opacity = 0.1, rounded corners = 2mm, dashed ]
					(16mm,14mm) rectangle (-6mm,-8mm);
			\draw[ color = black!100 ] (5.0mm,-5.5mm) node[ rotate = 0 ] {$\{\Lik^{(1)},\Lik^{(2)},\ldots,\}$};
			\draw[ color = black!100 ] (5.0mm,-9.5mm) node[ rotate = 0 ] {Model comparison};
		\end{scope}
	\end{tikzpicture}
	\caption{SLIM in a nutshell. Starting from a training-test set partition of data $\{\X,\X^\star\}$, our framework produces factor models $\C$ and DAG candidates $\B$ with and without latent variables $\Z$ that can be compared in terms of how well they fit the data using test likelihoods $\Lik$. The variable ordering $\P$ needed by the DAG is obtained as a byproduct of a factor model inference. Besides, changing the prior over latent variables $\Z$ produces two variants of SLIM called CSLIM and SNIM.} \label{fg:slimfig}
\end{figure}
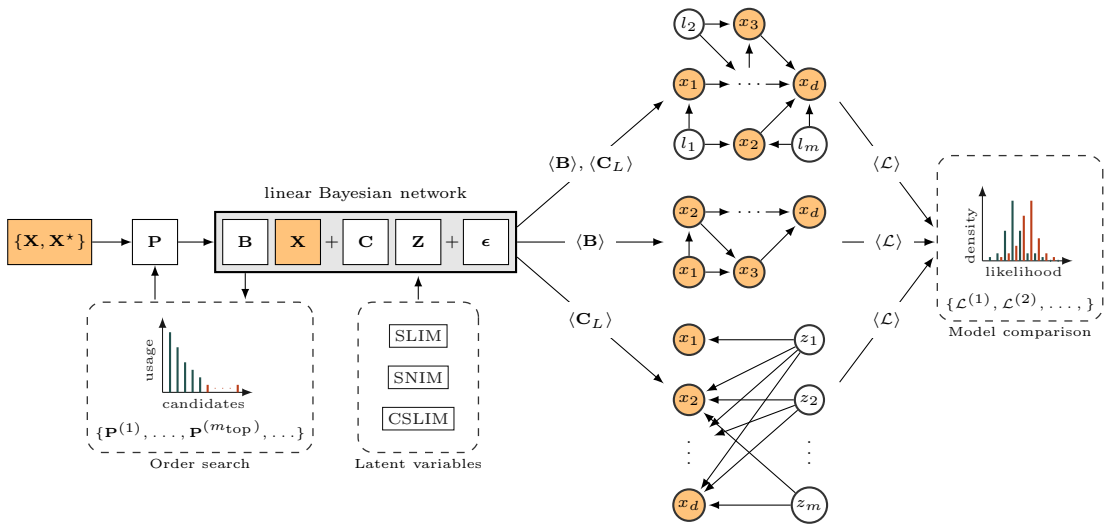

In this paper we propose a framework called SLIM (Sparse Linear Identifiable Multivariate modeling, see Figure \ref{fg:slimfig}) in which we learn models from a rather general class of Bayesian networks and perform quantitative model comparison between them\footnote{A preliminary version of our approach appears in NIPS 2009: Henao and Winther, Bayesian sparse factor models and DAGs inference and comparison.}. Model comparison may be used for model selection or serve as a hypothesis-generating tool. We use the likelihood on a test set as a computationally simple quantitative proxy for model comparison and as an alternative to the marginal likelihood. The other two key ingredients in the framework are the use of sparse and identifiable model components \citep[][respectively]{carvalho08,kagan73} and the stochastic search for the correct order of the variables needed by the DAG representation. Like LiNGAM, SLIM exploits the close relationship between factor models and DAGs. However, since we are interested in the factor model by itself, we will not constrain the factor loading matrix to have triangular form, but allow for sparse solutions so pruning is not needed. Rather we may ask whether there exists a permutation of the factor-loading matrix agreeing to the DAG assumption (in a probabilistic sense). The slab and spike prior biases towards sparsity so it makes sense to search for a permutation in parallel with factor model inference. We propose to use stochastic updates for the permutation using a Metropolis-Hastings acceptance ratio based on likelihoods with the factor-loading matrix being masked. In practice this approach gives good solutions up to at least fifty dimensions. Given a set of possible variable orderings inferred by this method, we can then learn DAGs using slab and spike priors for their connectivity matrices. The so-called slab and spike prior is a two-component mixture of a continuous distribution and degenerate $\delta$-function point mass at zero. This type of model implicitly defines a prior over structures and is thus a computationally attractive alternative to combinatorial structure search since parameter and structure inference are performed simultaneously. A key to effective learning in these intractable models is Markov Chain Monte Carlo (MCMC) sampling schemes that mix well. For non-Gaussian heavy-tailed distributions like the Laplace and $t$-distributions, Gibbs sampling can be efficiently defined using appropriate infinite scale mixture representations of these distributions \citep{andrews74}. We also show that our model is very flexible in the sense that it can be easily extended by only changing the prior distribution of a set of latent variables, for instance to allow for time series data (CSLIM, Correlated SLIM) and non-linearities in the DAG structure (SNIM, Sparse non-Linear Identifiable Multivariate modeling) through Gaussian process priors.

The rest of the paper is organized as follows: Section \ref{sc:lin} describes the model and its identifiability properties. Section \ref{sc:noi} provides all prior specification including sparsity, latent variables and driving signals, order search and extensions for correlated data (CSLIM) and non-linearities (SNIM). Section \ref{sc:ms} elaborates on model comparison. Section \ref{sc:inf} and Appendix \ref{ap:inf} provide an overview of the model and practical details on the MCMC-based inference, proposed workflow and computational cost requirements. Section \ref{sc:res} contains the experiments. We show simulations based on artificial data to illustrate all the features of the model proposed. Real biological data experiments illustrate the advantages of considering different variants of Bayesian networks. For all data sets we compare with some of the most relevant existing methods.  Section \ref{sc:dis} concludes with a discussion, open questions and future directions.
\section{Linear Bayesian networks} \label{sc:lin}
A Bayesian network is essentially a joint probability distribution defined via a directed acyclic graph, where each node in the graph represents a random variable $x$. Due to the acyclic property of the graph, its node set ${x_1,\ldots,x_d}$ can be partitioned into $d$ subsets $\{V_1,V_2,\ldots,V_d\}\equiv\VR$, such that if $x_j\rightarrow x_i$ then $x_j\in V_i$, \ie $V_i$ contains all \emph{parents} of $x_i$. We can then write the joint distribution as a product of conditionals of the form
\begin{align*} 
	P(x_1,\ldots,x_d)=\prod_{i=1}^d P(x_i|V_i) \ ,
\end{align*}
thus $x_i$ is conditionally independent of $\{x_j|x_i\notin V_j\}$ given $V_i$ for $i\neq j$. This means that $p(x_1,\ldots,x_d)$ can be used to describe the joint probability of any set of variables once $\VR$ is given. The problem is that $\VR$ is usually unknown and thus needs to be (at least partially) inferred from observed data.

We consider a model for a fairly general class of linear Bayesian networks by putting together a linear DAG, $\x=\B\x+\z$, and a factor model, $\x=\C\z+\bepsilon$. Our goal is to explain each one of $d$ observed variables $\x$ as a linear combination of the remaining ones, a set of $d+m$ independent latent variables $\z$ and additive noise $\bepsilon$. We have then
\begin{align} \label{eq:PBxCz}
	\x=(\R\odot\B)\x + (\Q\odot\C)\z + \bepsilon \ ,
\end{align}
where $\odot$ is the element-wise product and we can further identify the following elements:
\begin{list}{\labelitemi}{\leftmargin=1em}
	\item $\z$ is partitioned into two subsets, $\z_D$ is a set of $d$ driving signals for each observed variable in $\x$ and $\z_L$ is a set of $m$ shared general purpose latent variables. $\z_D$ is used here to describe the intrinsic behavior of the observed variables that cannot regarded as \quotes{external} noise.
	\item $\R$ is a $d\times d$ binary connectivity matrix that encodes whether there is an edge between observed variables, by means of $r_{ij}=1$ if $x_i\to x_j$. Since every non-zero element in $\R$ is an edge of a DAG, $r_{ii}=0$ and $r_{ij}=0$ if $r_{ji}\neq0$ to avoid self-interactions and bi-directional edges, respectively. This also implies that there is at least one permutation matrix $\P$ such that $\P\ts\R\P$ is strictly lower triangular where we have used that $\P$ is orthonormal then $\P\inv = \P\ts$.
	\item $\Q=[ \Q_D \ \Q_L ]$ is a $d\times (d+m)$ binary connectivity matrix, this time for the conditional independence relations between observed and latent variables. We assume that each observed variable has a dedicated latent variable, thus the first $d$ columns of $\Q_D$ are the identity. The remaining $m$ columns can be arbitrarily specified, by means of $q_{ij}\neq0$ if there is an edge between $x_i$ and $z_j$ for $d<j\leq m$.
	\item $\B$ and $\C=[\C_L \ \C_D]$ are respectively, $d\times d$ and $d\times(d+m)$ weight matrices containing the edge strengths for the Bayesian network. Their elements are constrained to be non-zero only if their corresponding connectivities are also non-zero.
\end{list}
The model \eqref{eq:PBxCz} has two important special cases, (i) if all elements in $\R$ and $\Q_D$ are zero it becomes a standard factor model (FM) and (ii) if $m=0$ or all elements in $\Q_L$ are zero it is a pure DAG. The model is not a completely general linear Bayesian network because connections to latent variables are absent \citep[see for example][]{silva10}. However, this restriction is mainly introduced to avoid compromising the identifiability of the model. In the following we will only write $\Q$ and $\R$ explicitly when we specify the sparsity modeling.
\subsection{Identifiability} \label{sc:idf}
We will split the identifiability of the model in equation \eqref{eq:PBxCz} in three parts addressing first the factor model, second the pure DAG and finally the full model. By identifiability we mean that each different setting of the parameters $\B$ and $\C$ gives a unique distribution of the data. In some cases the model is only unique up to some symmetry of the model. We discuss these symmetries and their effect on model interpretation in the following.

Identifiability in factor models $\x=\C_L\z_L+\bepsilon$ can be obtained in a number of ways \citep[see Chapter 10,][]{kagan73}. Probably the easiest way is to assume sparsity in $\C_L$ and restrict its number of free parameters, for example by restricting the dimensionality of $\z$, namely $m$, according to the Ledermann bound $m\leq(2d+1-(8d+1)^{1/2})/2$ \citep{bekker97}. The Ledermann bound guarantees the identification of $\bepsilon$ and follows just from counting the number of free parameters in the covariance matrices of $\x$, $\bepsilon$ and in $\C_L$, assuming Gaussianity of $\z$ and $\bepsilon$. Alternatively, identifiability is achieved using non-Gaussian distributions for $\z$. \citet[Theorem 10.4.1,][]{kagan73} states that when at least $m-1$ latent variables are non-Gaussian, $\C_L$ is identifiable up to scale and permutation of its columns, i.e.\ we can identify $\widehat{\C}_L = \C_L \S_\mathrm{f} \PC$, where $\S_\mathrm{f}$ and $\PC$ are arbitrary scaling and permutation matrices, respectively. \citet{comon94} provided an alternative well-known proof for the particular case of $m-1=d$. The $\S_\mathrm{f}$ and $\PC$ symmetries are inherent in the factor model definition in all cases and will usually not affect interpretability. However, some researchers prefer to make the model completely identifiable, \eg by making $\C_L$ triangular with non-negative diagonal elements \citep{lopes04}. In addition, if all components of $\bepsilon$ are Gaussian and the rank of $\C_L$ is $m$, then the distributions of $\z$ and $\bepsilon$ are uniquely defined to within common shift in mean \citep[Theorem 10.4.3,][]{kagan73}. In this paper, we use the non-Gaussian $\z$ option for two reasons, (i) restricting the number of latent variables severely limits the usability of the model and (ii) non-Gaussianity is a more realistic assumption in many application areas such as for example biology.

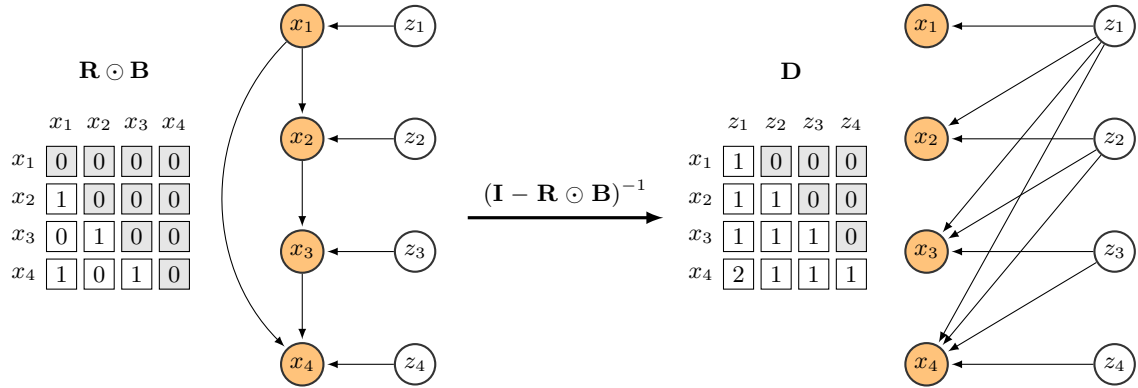
\begin{figure}[tb]
	\begin{tikzpicture}[ bend angle = 45, >=latex, font = \footnotesize ]
		\tikzstyle{obs} = [ circle, thick, draw = black!80, fill = imp2, minimum size = 1mm, inner sep = 2pt ]
		\tikzstyle{lat} = [ circle, thick, draw = black!80, fill = black!0, minimum size = 1mm, inner sep = 2pt ]
		\tikzstyle{cellf} = [ rectangle, draw = black!100, fill = black!0, minimum width = 4mm, minimum height = 4mm, inner sep = 2pt ]
		\tikzstyle{celle} = [ rectangle, draw = black!100, fill = black!10, minimum width = 4mm, minimum height = 4mm, inner sep = 2pt ]
		\tikzstyle{lab} = [ rectangle, draw = black!0, fill = black!0, minimum width = 4mm, minimum height = 4mm, inner sep = 2pt ]
		\tikzstyle{every label} = [black!100]
		\begin{scope}[ node distance = 1.5cm and 1.5cm, rounded corners = 4pt, xshift = 3.2cm, yshift = 1.8cm ]
			\node [obs] (x1)  [] {$x_1$};
			\node [lat] (z1) [ right of = x1 ] {$z_1$}
				edge [post] (x1);
			\node [obs] (x2) [ below of = x1 ] {$x_2$}
				edge [pre] (x1);
			\node [lat] (z2) [ right of = x2 ] {$z_2$}
				edge [post] (x2);
			\node [obs] (x3) [ below of = x2 ] {$x_3$}
				edge [pre] (x2);
			\node [lat] (z3) [ right of = x3 ] {$z_3$}
				edge [post] (x3);
			\node [obs] (x4) [ below of = x3] {$x_4$}
				edge [pre] (x3)
				edge [pre, bend left] (x1);
			\node [lat] (z4) [ right of = x4 ] {$z_4$}
				edge [post] (x4);
		\end{scope}
		\begin{scope}[ node distance = 1.5cm, rounded corners = 4pt, xshift = 11.5 cm, yshift = 1.8cm ]
			\node [obs] (x1)  [] {$x_1$};
			\node [obs] (x2) [ below of = x1 ] {$x_2$};
			\node [obs] (x3) [ below of = x2 ] {$x_3$};
			\node [obs] (x4) [ below of = x3] {$x_4$};
			\node [lat] (z1) [ right of = x1, xshift = 10mm ] {$z_1$}
				edge [post] (x1)
				edge [post] (x2)
				edge [post] (x3)
				edge [post] (x4);
			\node [lat] (z2) [ right of = x2, xshift = 10mm ] {$z_2$}
				edge [post] (x2)
				edge [post] (x3)
				edge [post] (x4);
			\node [lat] (z3) [ right of = x3, xshift = 10mm ] {$z_3$}
				edge [post] (x3)
				edge [post] (x4);
			\node [lat] (z4) [ right of = x4, xshift = 10mm ] {$z_4$}
				edge [post] (x4);
		\end{scope}
		\begin{scope}[ node distance = 5mm ]
			\node [celle] (e11)  [] {$0$};
			\node [celle] (e12)  [ right of = e11 ] {$0$};
			\node [celle] (e13)  [ right of = e12 ] {$0$};
			\node [celle] (e14)  [ right of = e13 ] {$0$};
			\node [cellf] (e21)  [ below of = e11 ] {$1$};
			\node [celle] (e22)  [ right of = e21 ] {$0$};
			\node [celle] (e23)  [ right of = e22 ] {$0$};
			\node [celle] (e24)  [ right of = e23 ] {$0$};
			\node [cellf] (e31)  [ below of = e21 ] {$0$};
			\node [cellf] (e32)  [ right of = e31 ] {$1$};
			\node [celle] (e33)  [ right of = e32 ] {$0$};
			\node [celle] (e34)  [ right of = e33 ] {$0$};
			\node [cellf] (e41)  [ below of = e31 ] {$1$};
			\node [cellf] (e42)  [ right of = e41 ] {$0$};
			\node [cellf] (e43)  [ right of = e42 ] {$1$};
			\node [celle] (e44)  [ right of = e43 ] {$0$};
			\node [lab] (lt1)  [ above of = e11 ] {$x_1$};
			\node [lab] (lt2)  [ above of = e12 ] {$x_2$};
			\node [lab] (lt3)  [ above of = e13 ] {$x_3$};
			\node [lab] (lt4)  [ above of = e14 ] {$x_4$};
			\node [lab] (ls1)  [ left of = e11 ] {$x_1$};
			\node [lab] (ls2)  [ left of = e21 ] {$x_2$};
			\node [lab] (ls3)  [ left of = e31 ] {$x_3$};
			\node [lab] (ls4)  [ left of = e41 ] {$x_4$};
			\draw (0.7,1.2) node {$\R\odot\B$};
		\end{scope}
		\begin{scope}[node distance = 5mm, xshift=9cm]
			\node [cellf] (e11)  [] {$1$};
			\node [celle] (e12)  [ right of = e11 ] {$0$};
			\node [celle] (e13)  [ right of = e12 ] {$0$};
			\node [celle] (e14)  [ right of = e13 ] {$0$};
			\node [cellf] (e21)  [ below of = e11 ] {$1$};
			\node [cellf] (e22)  [ right of = e21 ] {$1$};
			\node [celle] (e23)  [ right of = e22 ] {$0$};
			\node [celle] (e24)  [ right of = e23 ] {$0$};
			\node [cellf] (e31)  [ below of = e21 ] {$1$};
			\node [cellf] (e32)  [ right of = e31 ] {$1$};
			\node [cellf] (e33)  [ right of = e32 ] {$1$};
			\node [celle] (e34)  [ right of = e33 ] {$0$};
			\node [cellf] (e41)  [ below of = e31 ] {$2$};
			\node [cellf] (e42)  [ right of = e41 ] {$1$};
			\node [cellf] (e43)  [ right of = e42 ] {$1$};
			\node [cellf] (e44)  [ right of = e43 ] {$1$};
			\node [lab] (lt1)  [ above of = e11 ] {$z_1$};
			\node [lab] (lt2)  [ above of = e12 ] {$z_2$};
			\node [lab] (lt3)  [ above of = e13 ] {$z_3$};
			\node [lab] (lt4)  [ above of = e14 ] {$z_4$};
			\node [lab] (ls1)  [ left of = e11 ] {$x_1$};
			\node [lab] (ls2)  [ left of = e21 ] {$x_2$};
			\node [lab] (ls3)  [ left of = e31 ] {$x_3$};
			\node [lab] (ls4)  [ left of = e41 ] {$x_4$};
			\draw (0.7,1.2) node {$\D$};
		\end{scope}
		\draw [-to,very thick,->]
			(5.4cm,-0.75cm) -- (8cm,-0.75cm)
			node [above=0mm,midway,text width=3cm,text centered] {$(\I-\R\odot\B)\inv$};
	\end{tikzpicture}
	\caption{FM-DAG equivalence illustration. In the left side, a DAG model with four variables with corresponding connectivity matrix $\R$, $b_{ij}=1$ when $r_{ij}=1$ and $\C_D=\I$. In the right hand side, the equivalent factor model with mixing matrix $\D$. Note that the factor model is sparse even if its corresponding DAG is dense. The gray boxes in $\D$ and $\R\odot\B$ represent elements that must be zero by construction.} \label{fg:DAGtoFA} \vspace{-5mm}
\end{figure}

For pure DAG models $\x = \B \x + \C_D \z_D$, identifiability can be obtained using the factor model result from \citet{kagan73} by rewriting the DAG into an equivalent factor model $\x = \D \z$ with $\D=(\I - \B)^{-1} \C_D$, see Figure \ref{fg:DAGtoFA}. From the factor model result it only follows that $\D$ is identifiable up to a scaling and permutation. However, as mentioned above, due to the acyclicity there is at least one permutation matrix $\P$ such that $\P\ts\B\P$ is strictly lower triangular. Now, if $\x$ admits DAG representation, the same $\P$ makes the permuted $\widehat{\D}=(\I - \P\ts \B \P)^{-1} \C_D$, triangular with $\C_D$ on its diagonal. The constraint on the number of non-zero elements in $\D$ due to triangularity removes the permutation freedom $\PC$ such that we can subsequently identify $\P$, $\B$ and $\C_D$. It also implies that any valid permutation $\P$ will produce exactly the same distribution for $\x$.

In the general case in equation \eqref{eq:PBxCz}, $\D=(\I-\B)^{-1}\C$ is of size $d\times(d+m)$. What we will show is that even if $\D$ is still identifiable, we can no longer obtain $\B$ and $\C$ uniquely unless we \quotes{tag} the model by requiring the distributions of driving signals $\z_D$ and latent signals $\z_L$ to differ. In order to illustrate why we get non-identifiability, we can write $\x=\D\z$ inverting $\D$ explicitly. For simplicity we consider $m=1$ and $\P=\I$ but generalizing to $m>1$ is straight forward

{\small
\begin{align*}
	\left[\begin{array}{c}
		x_1 \\ x_2 \\ x_3 \\ \vdots \\ x_d
	\end{array}\right] =
	\left[\begin{array}{ccccc}
		c_{11} & 0 & 0 & \cdots & c_{1L} \\
		b_{21}c_{11} & c_{22} & 0 & \cdots & b_{21}c_{1L} + c_{2L} \\
		b_{31}c_{11} + b_{32}b_{21}c_{11} & b_{32}c_{22} & c_{33} & \cdots & b_{31}c_{1L} + b_{32}b_{21}c_{1L} + a_{32}c_{2L} + c_{3L} \\
		\vdots & \vdots & \vdots & \ddots & \vdots \\
		c_{11} + \sum_{k=1}^{i-1}b_{ik}d_{k1} & \cdots & \cdots & \cdots & c_{iL} + \sum_{k=1}^{i-1}b_{ik}d_{kL}
	\end{array}\right]
	\left[\begin{array}{c}
		z_1 \\ z_2 \\ z_3 \\ \vdots \\ z_{d+1}
	\end{array}\right] .
\end{align*}
}

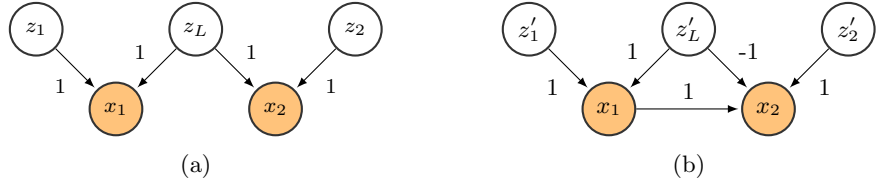
\begin{figure}[tb]
	\centering
	\subfigure[]{\begin{tikzpicture}[ bend angle = 45, >=latex, font = \scriptsize ]
		\tikzstyle{obs} = [ circle, thick, draw = black!80, fill = imp2, minimum size = 7mm, inner sep = 2pt ]
		\tikzstyle{lat} = [ circle, thick, draw = black!80, fill = black!0, minimum size = 7mm, inner sep = 2pt ]
		\tikzstyle{dmy} = [ circle, thick, draw = black!0, fill = black!0, minimum size = 7mm, inner sep = 1pt]
		\tikzstyle{every label} = [black!100]
		\begin{scope}[ node distance = 1.5cm and 1.5cm, rounded corners = 4pt ]
			\node [lat] (L)  [] {$z_L$};
			\node [obs] (x1) [ below left of = L ] {$x_1$}
				edge [pre] node[ above left ] {1} (L);
			\node [obs] (x2) [ below right of = L ] {$x_2$}
				edge [pre] node[ above right ] {1} (L);
			\node [lat] (z1) [ above left of = x1 ] {$z_1$}
				edge [post] node[ below left ] {1} (x1);
			\node [lat] (jnk) [ above right of = x2 ] {$z_2$}
				edge [post] node[ below right ] {1} (x2);
		\end{scope}
	\end{tikzpicture} \label{fg:toyDAGL1} } \hspace{12mm}
	\subfigure[]{\begin{tikzpicture}[ bend angle = 45, >=latex, font = \footnotesize ]
		\tikzstyle{obs} = [ circle, thick, draw = black!80, fill = imp2, minimum size = 7mm, inner sep = 2pt ]
		\tikzstyle{lat} = [ circle, thick, draw = black!80, fill = black!0, minimum size = 7mm, inner sep = 2pt ]
		\tikzstyle{dmy} = [ circle, thick, draw = black!0, fill = black!0, minimum size = 7mm, inner sep = 1pt]
		\tikzstyle{every label} = [black!100]
		\begin{scope}[ node distance = 1.5cm and 1.5cm, rounded corners = 4pt ]
			\node [lat] (L)  [] {$z'_L$};
			\node [obs] (x1) [ below left of = L ] {$x_1$}
				edge [pre] node[ above left ] {1} (L);
			\node [obs] (x2) [ below right of = L ] {$x_2$}
				edge [pre] node[ above right ] {-1} (L)
				edge [pre] node[ above ] {1} (x1);
			\node [lat] (z1) [ above left of = x1 ] {$z'_1$}
				edge [post] node[ below left ] {1} (x1);
			\node [lat] (jnk) [ above right of = x2 ] {$z'_2$}
				edge [post] node[ below right ] {1} (x2);
		\end{scope}
	\end{tikzpicture} \label{fg:toyDAGL2} }
	\caption{Two DAGs with latent variables. They are equivalent if $\z$ has the same distribution as $\z'$.} \label{fg:toyDAGL} \vspace{-5mm}
\end{figure}

We see from this equation that if all latent variables have the same distribution and $c_{1L}$ is non-zero then we may exchange the first and last column in $\D$ to get two equivalent distributions with different elements for $\B$ and $\C$. The model is thus non-identifiable. If the first $i$ elements in latent column of $\C$ are zero then the $(i+1)$-th and last column can be exchanged. \citet{hoyer08a} made the same basic observation through a number of examples. Interestingly, we also see from the triangularity requirement of the \quotes{driving signal} part of $\D$ that $\P$ is actually identifiable despite the fact that $\B$ and $\C$ are not. To illustrate that the non-identifiability may lead to quite severe confusion about inferences, consider a model with only two observed variables $\x=[x_1,x_2]\ts$ and $c_{11}=c_{22}=1$. Two different hypothesis $\{b_{21},c_{1L},c_{2L}\}=\{0,1,1\}$ and $\{b_{21},c_{1L},c_{2L}\}=\{1,1,-1\}$ with graphs shown in Figure \ref{fg:toyDAGL} have equivalent factor models written as
\begin{align*}
	\left[\begin{array}{c} x_1 \\ x_2 \end{array}\right] = \left[\begin{array}{ccc} 1 & 0 & 1 \\ 0 & 1 & 1 \end{array}\right]\left[\begin{array}{c} z_1 \\ z_2 \\ z_L \end{array}\right] \ \mathrm{and} \  \left[\begin{array}{c} x_1 \\ x_2 \end{array}\right] = \left[\begin{array}{ccc} 1 & 0 & 1 \\ 1 & 1 & 0 \end{array}\right]\left[\begin{array}{c} z'_1 \\ z'_2 \\ z'_L \end{array}\right] \ .
\end{align*}
The two models above have the same mixing matrix $\D$, up to permutation of columns $\PC$. In general we expect the number of solutions with equivalent distribution may be as large as $2^m$, corresponding to the number of times a column of $\D$ from its latent part (last $m$ columns) con be exchanged with a column from its observed part (first $d$ columns). This readily assumes that the sparsity pattern in $\D$ is identified, which follows from the results of \citet{kagan73}.

One way to get identifiability is to change the distributions $\z_D$ and $\z_L$ such that they differ and cannot be exchanged. Here it is not enough to change the scale of the variables, \ie variance for continuous variables, because this effect can be countered by rescaling $\C$ with $\S_{\mathrm{f}}$. So we need distributions that differ beyond rescaling. In our examples we use Laplace and the more heavy-tailed Cauchy for $\z_D$ and $\z_L$, respectively. This specification is not unproblematic in practical situations however it can be sometimes restrictive and prone to model mismatch issues. We nevertheless show one practical example which leads to sensible inferences.

In time series applications for example, it is natural to go beyond an \iid model for $\z$. One may for example use a Gaussian process prior for each factor to get smoothness over time, \ie $z_{j1},\ldots,z_{jN}|\nu_j \sim \DN(0,\K_{\nu_j})$, where $\K_{\nu_j}$ is the covariance matrix with elements $k_{j,nn'}=k_{\upsilon_j,n}(n,n')$ and $k_{\upsilon_j,n}(\cdot)$ is the covariance function. For the \iid Gaussian model the source distribution is only identifiable up to an arbitrary rotation matrix $\U$, \ie the rotated factors $\U\z$ are still \iid. We can show that contrary to the \iid Gaussian model, the Gaussian process factor model is identifiable if the covariance functions differ. We need to show that $\widehat{\Z}= \U \Z$ has a different covariance structure than $\Z=[\z_1 \ \ldots \ \z_N]$. We get $\overline{\z_n \z_{n'}\ts}= {\rm diag}(k_{1,nn'},\ldots,k_{d+m,nn'})$ and $\overline{\widehat{\z}_n\widehat{\z}_{n'}\ts } = \U \overline{\z_n \z_{n'}\ts} \U\ts = \U {\rm diag}(k_{1,nn'},\ldots,k_{d+m,nn'}) \U\ts$ for the original and rotated variables, respectively. The covariances are indeed different and the model is thus identifiable if no covariance functions $k_{\upsilon_j,n}(n,n')$, $j=1,\ldots,d+m$ are the same.
\section{Prior specification} \label{sc:noi}
In this section we provide a detailed description of the priors used for each one of the elements of our sparse linear identifiable model already defined in equation \eqref{eq:PBxCz}. We start with $\bepsilon$, the noise term that allow us to quantify the mismatch between a set of $N$ observations $\X=[\x_1 \ \ldots \ \x_N]$ and the model itself. For this purpose, we use uncorrelated Gaussian noise components $\bepsilon \sim \DN(\bepsilon|\0,\bPsi)$ with conjugate inverse gamma priors for their variances as follows
\begin{align*}
	\X|\m,\bPsi & \sim \prod_{n=1}^N\DN(\x_n|\m,\bPsi) \ , \\ 
	\bPsi\inv|s_s,s_r & \sim \prod_{i=1}^d\Ga(\psi_i\inv|s_s,s_r) \ , 
\end{align*}
where we have already marginalized out $\bepsilon$, $\bPsi$ is a diagonal covariance matrix denoting uncorrelated noise across dimensions and $\m$ is the mean vector such that $\m_\FM=\C\z_n$ and $\m_\DAG=\B\x_n+\C\z_n$. In the noise covariance hyperprior, $s_s$ and $s_r$ are the shape and rate, respectively. The selection of hyperparameters for $\bPsi$ should not be very critical as long as both \quotes{signal and noise} hypotheses are supported, \ie diffuse enough to allow for small values of $\psi_i$ as well as for $\psi_i=1$ (assuming that the data is standardized in advance).  We set $s_s=20$ and $s_r=1$ in the experiments for instance. Another issue to consider when selecting $s_s$ and $s_r$ is the Bayesian analogue of the Heywood problem in which likelihood functions are bounded below away from zero as $\psi_i$ tends to zero, hence inducing multi-modality in the posterior of $\psi_i$ with one of the modes at zero. The latter can be avoided by specifying $s_s$ and $s_r$ such that the prior decays to zero at the origin, as we did above. It is well known, for example, that Heywood problems cannot be avoided using improper reference priors, $p(\psi_i)\propto1/\psi_i$ \citep{martin75}.

The remaining components of the model are described as it follows in five parts named sparsity, latent variables and driving signals, order search, allowing for correlated data and allowing for non-linearities. The first part addresses the interpretability of the model by means of parsimonious priors for $\C$ and $\D$. The second part describes the type of non-Gaussian distributions used on $\z$ in order to keep the model identifiable. The third part considers how a search over permutations of the observed variables can be used in order to handle the constraints imposed on matrix $\R$. The last two parts describe how introducing Gaussian process process priors in the model can be used to model non-independent observations and non-linear dependencies in the DAGs.
\subsection{Sparsity} \label{sc:sse}
The use of sparse models will in many cases give interpretable results and is often motivated by the principle of parsimony. Also, in many application domains it is also natural from a prediction point of view to enforce sparsity because the number of explanatory variables may exceed the number of examples by orders of magnitude. In regularized maximum likelihood type formulations of learning (maximum a-posteriori) it has become popular to use one-norm ($L_1$) regularization for example to achieve sparsity \citep{tibshirani96}. In the fully Bayesian inference setting (with averaging over variables), the corresponding Laplace prior will not lead to sparsity because it is very unlikely for a posterior summary like the mean, median or mode to be estimated as exactly zero even asymptotically. The same effect can be expected from any continuous distribution used for sparsity like Student's $t$,  $\alpha$-stable and bimodal priors \citep[continuous slab and spike priors,][]{ishwaran05}. Exact zeros can only be achieved by placing a point mass at zero, \ie explicitly specifying that the variable at hand is zero or not with some probability. This has motivated the introduction of many variants over the years of so-called slab and spike priors consisting of two component mixtures of a continuous part and a $\delta$-function at zero \citep{lempers71,mitchell88,george93,geweke96,west03}. In this paradigm, the columns of matrices $\C$ or $\B$ encode respectively, the connectivity of a factor or the set of parents associated to an observed variable. It is natural then to share information across elements in column $j$ by assuming a common sparsity level $1-\nu_j$, suggesting the following hierarchy
\begin{align} \label{eq:hdss1}
	\begin{aligned}
		c_{ij}|q_{ij},\cdot \ \sim & \ (1-q_{ij})\delta(c_{ij}) + q_{ij}{\rm Cont}(c_{ij}|\cdot) \ , \\
		q_{ij}|\nu_j \ \sim & \ \Ber(q_{ij}|\nu_j) \ , \\
		\nu_j|\beta_m,\beta_p \ \sim & \ \Be(\nu_j|\beta_p\beta_m,\beta_p(1-\beta_m)) \ ,
	\end{aligned}
\end{align}
where $\Q$, the binary matrix in equation \eqref{eq:PBxCz} appears naturally, $\delta(\cdot)$ is a Dirac $\delta$-function, ${\rm Cont}(\cdot)$ is the continuous slab component, $\Ber(\cdot)$ and $\Be(\cdot)$ are Bernoulli and beta distributions, respectively. Reparameterizing the beta distribution as $\Be(\nu_j|\alpha\beta/m,\beta)$ and taking the number of columns $m$ of $\Q\odot\C$ to infinity, leads to the non-parametric version of the slab and spike model with a so-called Indian buffet process prior over the (infinite) masking matrix $\Q=\{q_{ij}\}$ \citep{ghahramani06}. Note also that $q_{ij}|\nu_j$ is mainly used for clarity to make the binary indicators explicit, nevertheless in practice we can work directly with $c_{ij}|\nu_j,\cdot \sim (1-\nu_j)\delta(c_{ij}) + \nu_j{\rm Cont}(c_{ij}|\cdot)$ because $q_{ij}$ can be marginalized out.

As illustrated and pointed out by \citet{lucas06} and \citet{carvalho08} the model with a shared beta-distributed sparsity level per factor introduces the undesirable side-effect that there is strong co-variation between the elements in each column of the masking matrix. For example, in high dimensions we might expect that only a finite number of elements are non-zero, implying a prior favoring a very high sparsity rate $1-\nu_j$. Because of the co-variation, even the parameters that are clearly non-zero will have a posterior probability of being non-zero, $p(q_{ij}=1|\x,\cdot)$, quite spread over the unit interval. Conversely, if our priors do not favor sparsity strongly, then the opposite situation will arise and the solution will become completely dense. In general, it is difficult to set the hyperparameters to achieve a sensible sparsity level. Ideally, we would like to have a model with a high sparsity level with high certainty about the non-zero parameters. We can achieve this by introducing a sparsity parameter $\eta_{ij}$ for each element of $\C$ which has a mixture distribution with exactly this property
\begin{align} \label{eq:hdss2}
	\begin{aligned}
		q_{ij}|\eta_{ij} \ \sim & \ \Ber(q_{ij}|\eta_{ij}) \ , \\
		\eta_{ij}|\nu_j,\alpha_p,\alpha_m \ \sim & \ (1-\nu_j)\delta(\eta_{ij})+\nu_j\Be(\eta_{ij}|\alpha_p\alpha_m,\alpha_p(1-\alpha_m)) \ .
	\end{aligned}
\end{align}
The distribution over $\eta_{ij}$ expresses that we expect parsimony: either $\eta_{ij}$ is zero exactly (implying that $q_{ij}$ and $c_{ij}$ are zero) or non-zero drawn from a beta distribution favoring high values, \ie $q_{ij}$ and $c_{ij}$ are non-zero with high probability. We use $\alpha_p=10$ and $\alpha_m=0.95$ which has mean $\alpha_m=0.95$ and variance $\alpha_m(1-\alpha_m)/(1+\alpha_p)\approx0.086$. The expected sparsity rate of the modified model is $(1-\alpha_m)(1-\nu_j)$. This model has the additional advantage that the posterior distribution of $\eta_{ij}$ directly measures the distribution of $p(q_{ij}=1|\x,\cdot)$. This is therefore the statistic for ranking/selection purposes. Besides, we may want to reject interactions with high uncertainty levels when the probability of $p(q_{ij}=1|\x,\cdot)$ is less or very close to the expected value, $\alpha_m(1-\nu_j)$. 

\begin{figure}[t]
	\centering
	\begin{psfrags}
		 \psfrag{mag}[c][c][0.6]{Density}\psfrag{aij}[c][c][0.7]{$c_{ij}$}\psfrag{eij}[c][c][0.7]{$\eta_{ij}$}\psfrag{a88}[c][c][0.7]{$c_{64}$}\psfrag{b88}[c][c][0.7]{$c_{54}$}\psfrag{c88}[c][c][0.7]{$\eta_{64}$}\psfrag{d88}[c][c][0.7]{$\eta_{54}$}
		\subfigure[]{\includegraphics[scale = 0.49]{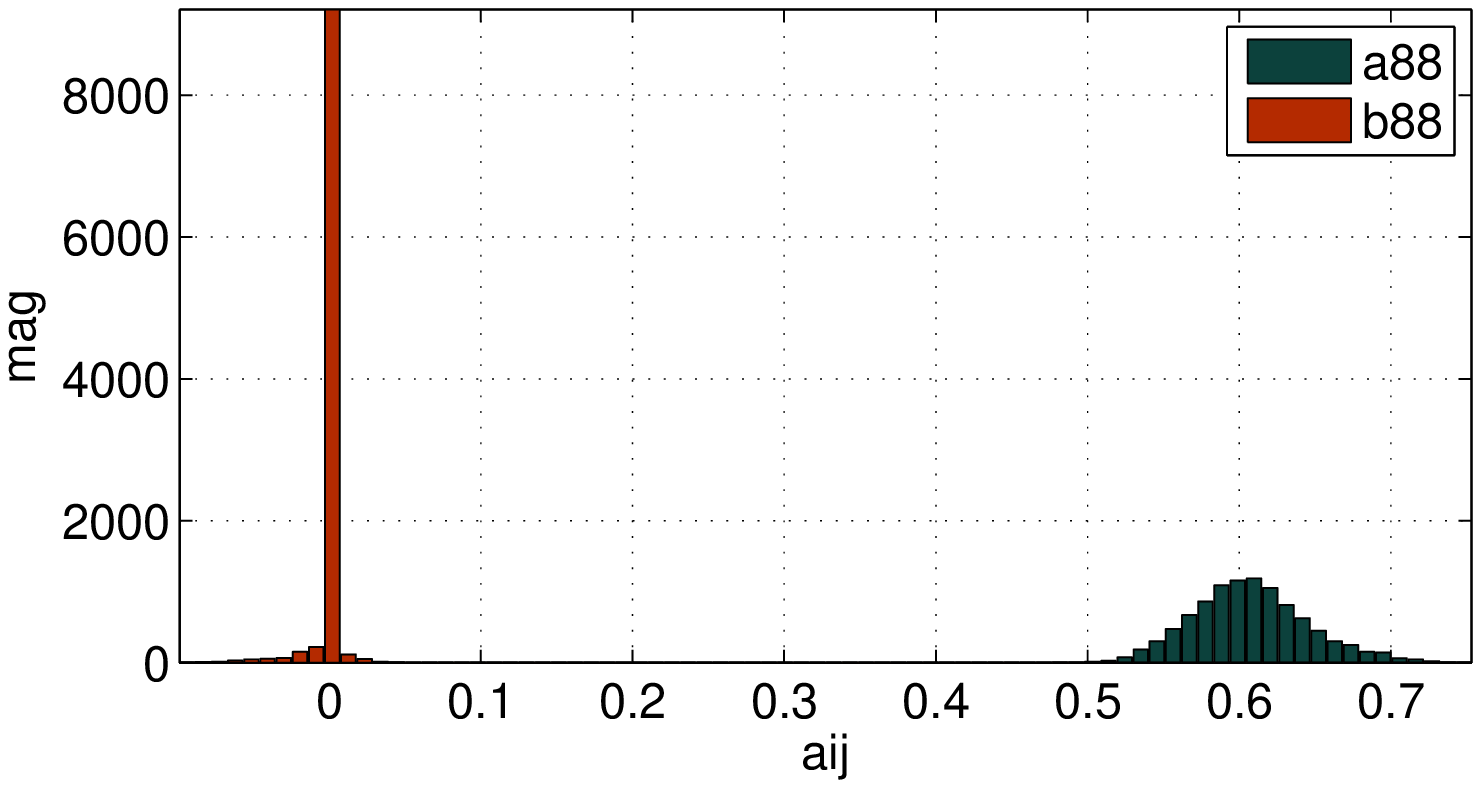}\label{fg:single_a}}
		\subfigure[]{\includegraphics[scale = 0.49]{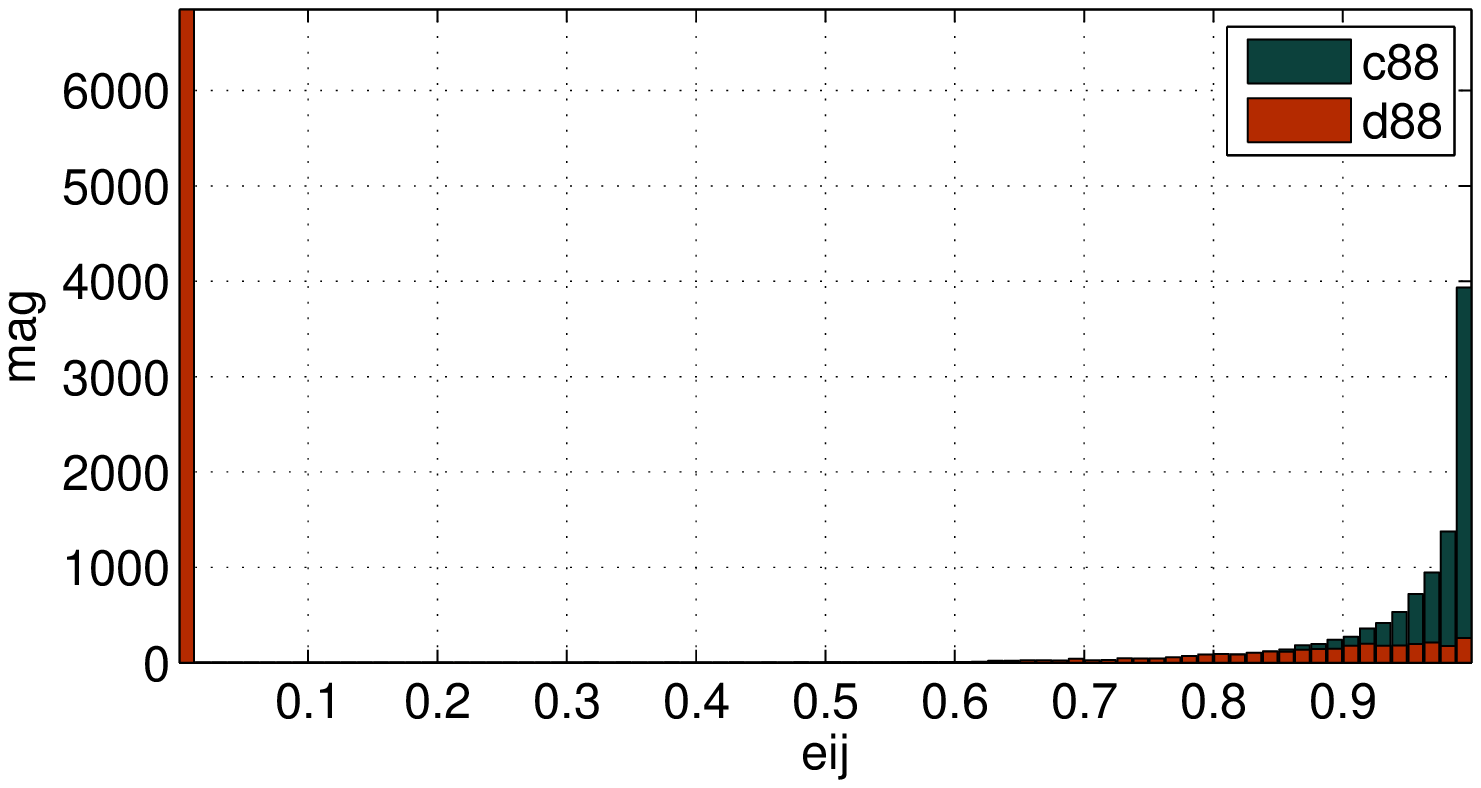}\label{fg:single_eta}}
		\vspace{-0.5cm}
	\end{psfrags}
	\caption{Slab and spike prior example. (a) Posterior unnormalized densities for the magnitude of two particular elements of $\C$. (b) Posterior density for $\eta_{ij}=p(c_{ij}\neq0|\x,\cdot)$. Here, $c_{64}\neq0$ and $c_{54}=0$ correspond to elements of the mixing matrix from the experiment shown in Figure \ref{fg:singlenets}.} \label{fg:single_aeta} \vspace{-5mm}
\end{figure}

To complete the specification of the prior, we let the continuous slab part in equation \eqref{eq:hdss1} be Gaussian distributed with inverse gamma prior on its variance. In addition, we scale the variances with $\psi_i$ as
\begin{align} \label{eq:hdss3}
	\begin{aligned}
		{\rm Cont}(c_{ij}|\psi_i,\tau_{ij}) \ = & \ \DN(c_{ij}|0,\psi_i\tau_{ij}) \ , \\
		\tau_{ij}\inv|t_s,t_r \ \sim & \ \Ga(\tau_{ij}\inv|t_s,t_r) \ . \\
	\end{aligned}
\end{align}
This scaling makes the model easier to specify and tend to have better mixing properties \citep[see][]{casella08}. The slab and spike for $\B$ (DAG) is obtained from equations \eqref{eq:hdss1}, \eqref{eq:hdss2} and \eqref{eq:hdss3}  by simply replacing $c_{ij}$ with $b_{ij}$ and $q_{ij}$ with $r_{ij}$. As already mentioned, we use $\alpha_p=10$ and $\alpha_m=0.95$ for the hierarchy in equation \eqref{eq:hdss2}. For the column-shared parameter $\nu_j$ defined in equation \eqref{eq:hdss1} we set the precision to $\beta_p=100$ and consider the mean values for factor models and DAGs separately. For the factor model we set a diffuse prior by making $\beta_m=0.9$ to reflect that some of the factors can be in general nearly dense or empty. For the DAG we consider two settings, if we expect to obtain dense graphs we set $\beta_m=0.99$, otherwise we set $\beta_m=0.1$. Both settings can produce sparse graphs, however smaller values of $\beta_m$ increase the overall sparsity rate and the gap between $p(r_{ij}=0)$ and $p(r_{ij}=1)$. A large separation between these two probabilities makes interpretation easier and also helps to spot non-zeros (edges) with high uncertainty. The hyperparameters for the variance of the non-zero elements of $\B$ and $\C$ are set to get a diffuse prior distribution bounded away from zero ($t_s=2$ and $t_r=1$), to allow for a better separation between slab and spike components. For the particular case of $\C_L$, in principle the prior should not have support on zero at all, \ie the driving signal should not vanish, however for simplicity we allow this anyway as it has not given any problems in practice. Figure \ref{fg:single_aeta} shows a particular example of the posterior, $p(c_{ij},\eta_{ij}|\x,\cdot)$ for two elements of $\C$ under the prior just described. In the example, $c_{64}\neq0$ with high probability according to $\eta_{ij}$, whereas $c_{54}$ is almost certainly zero since most of its probability mass is located exactly at zero, with some residual mass on the vicinity of zero, in Figure \ref{fg:single_a}. In the one level hierarchy equation \eqref{eq:hdss1} sparsity parameters are shared, $\eta_{64}=\eta_{54}=\nu_4$. The result would then be less parsimonious with the posterior density of $\nu_4$ being spread in the unit interval with a single mode located close to $\beta_m$.
\subsection{Latent variables and driving signals} \label{sc:smog}
We consider two different non-Gaussian --- heavy-tailed priors for $\z$, in order to obtain identifiable factor models and DAGs. A wide class of continuous, unimodal and symmetric distributions in one dimension can be represented as infinite scale mixtures of Gaussians, which are very convenient for Gibbs-sampling-based inference. We focus on Student's $t$ and Laplace distributions which have the following mixture representation \citep{andrews74}
\begin{align}
\La(z|\mu,\lambda) \ = & \ \int_0^\infty\DN(z|\mu,\upsilon)\Exp(\upsilon|\lambda^2)d\upsilon \ , \label{eq:sLad} \\
t(z|\mu,\theta,\sigma^2) \ = & \ \int_0^\infty\DN(z|\mu,\upsilon\sigma^2)\Ga\left(\upsilon\inv\left|\frac{\theta}{2},\frac{\theta}{2}\right.\right)d\upsilon \ , \label{eq:sStd}
\end{align}
where $\lambda>0$ is the rate, $\sigma^2>0$ the scale, $\theta>0$ is the degrees of freedom, and the distributions have exponential and gamma mixing densities accordingly. For varying degrees of freedom $\theta$, the $t$ distribution can interpolate between very heavy-tailed (power law and Cauchy when $\theta=1$) and very light tailed, \ie it becomes Gaussian when the degrees of freedom approaches infinity. The Laplace (or bi-exponential) distribution has tails which are intermediate between a $t$ (with finite degrees of freedom) and a Gaussian. In this sense, the $t$ distribution is more flexible but requires more careful selection of its hyperparameters because the model may become non-identifiable in the large $\theta$ limit (Gaussian).

An advantage of the Laplace distribution is that we can fix its parameter $\lambda=1$ and let the model learn the appropriate scaling from $\C$ in equation \eqref{eq:PBxCz}. If we use the pure DAG model, we will need to have a hyperprior for $\lambda^2$ in order to learn the variances of the latent variables/driving signals, as in \citet{henao09}. A hierarchical prior for the degrees of freedom in the $t$ distribution is not easy to specify because there is no conjugate prior available with a standard closed form. Although a conjugate prior exists, is not straightforward to sample from it, since numerical integration must be used to compute its normalization constant. Another possibility is to treat $\theta$ as a discrete variable so computing the normalizing constant becomes straight forward.

Laplace and Student's $t$ are not the only distributions admitting scale mixture representation. This mean that any other compatible type can be used as well, if the application requires it, and without considerable additional effort. Some examples include the logistic distribution \citep{andrews74}, the stable family \citep{west87} and skewed versions of heavy-tailed distributions \citep{branco01}. Another natural extension to the mixtures scheme could be, for example, to set the mean of each component to arbitrary values and let the number of components be an infinite sum, thus ending up providing each factor with a Dirichlet process prior. This might be useful for cases when the latent factors are expected to be scattered in clusters due to the presence of subgroups in the data, as was shown by \citet{carvalho08}.
\subsection{Order search} 
We need to infer the order of the variables in the DAG to meet the constraints imposed on $\R$ in Section \ref{sc:lin}. The most obvious way is to try to solve this task by inferring all parameters $\{\P,\B,\C,\z,\bepsilon\}$ by a Markov chain Monte Carlo (MCMC) method such as Gibbs sampling. However, algorithms for searching over variable order prefer to work with models for which parameters other than $\P$ can be marginalized analytically \citep[see][]{friedman03,teyssier05}. For our model, where we cannot marginalize analytically over $\B$ (due to $\R$ being binary), estimating $\P$ and $\B$ by Gibbs sampling would mean that we had to propose a new $\P$ for fixed $\B$. For example, exchanging the order of two variables would mean that they also exchange parameters in the DAG. Such a proposal would have very low acceptance, mainly as a consequence of the size of the search space and thus very poor mixing. In fact, for a given $d$ number of variables there are $d!$ possible orderings $\P$, while there are $d!2^{(d(d+2m-1))/2}$ possible structures for $\{\P,\B,\C\}$. We therefore opt for an alternative strategy by exploiting the equivalence between factor models and DAGs shown in Section \ref{sc:idf}. In particular for $m=0$, since $\B$ can be permuted to strictly lower triangular, then $\D=(\I-\B)\inv\C_D$ can be permuted to triangular. This means that we can perform inference for the factor model to obtain $\D$ while searching in parallel for a set of permutations $\P$ that are in good agreement (in a probabilistic sense) with the triangular requirement of $\D$. Such a set of orderings is found during the inference procedure of the factor model. To set up the stochastic search, we need to modify the factor model slightly by introducing separate data (row) and factor (column) permutations, $\P$ and $\PC$ to obtain $\x = \P\ts \D\PC \z + \bepsilon$. The reason for using two different permutation matrices, rather than only one like in the definition of the DAG model, is that we need to account for the permutation freedom of the factor model (see Section \ref{sc:idf}). Using the same permutation for row and column would thus require an additional step to identify the columns in the factor model. We make inference for the unrestricted factor model, but propose $\P^\star$ and $\PC^\star$ independently according to $q(\P^\star|\P)q(\PC^\star|\PC)$. Both distributions draw a new permutation matrix by exchanging two randomly chosen elements, \eg the order may change as $[x_1,x_2,x_3,x_4]\ts \to [x_1,x_4,x_3,x_2]\ts$. In other words, the proposals $q(\P^\star|\P)$ and $q(\PC^\star|\PC)$ are uniform distributions over the space of transpositions for $\P$ and $\PC$. Assuming we have no a-priori preferred ordering, we may use a Metropolis-Hastings (M-H) acceptance probability $\min(1,\xi_{\rightarrow\star})$ with $\xi_{\rightarrow\star}$ as a simple ratio of likelihoods with the permuted $\D$ masked to match the triangularity assumption. Formally, we use the binary mask $\M$ (containing zeros above the diagonal of its $d$ first columns) and write
\begin{align} \label{eq:rlik}
	 \xi_{\rightarrow\star}=\frac{{\cal N}(\X|(\P^\star)\ts(\M\odot \P^\star \D (\PC^\star)\ts)\PC^\star\Z,\bPsi)}{{\cal N}(\X|\P\ts(\M\odot \P\D\PC\ts)\PC\Z,\bPsi)} \ ,
\end{align}
where $\M\odot\D$ is the masked $\D$ and $\Z=[\z_1 \ \ldots \z_N]$. The procedure can be seen as a simple approach for generating hypotheses about good orderings, producing close to triangular versions of $\D$, in a model where the slab and spike prior provide the required bias towards sparsity. Once the inference is done, we end up having an estimate for the desired distribution over permutations $\P=\sum_i^{d!}\pi_i\delta_{\P_i}$, where $\bpi=[\pi_1 \ \pi_2 \ \ldots]$ is a sparse vector containing the probability for $\P=\P_i$, which in our case is proportional to the number of times permutation $\P_i$ was accepted by the M-H update during inference. Note that $\PC$ is just a nuisance variable that does not need to be stored or summarized.
\subsection{Allowing for correlated data (CSLIM)} \label{sc:cslim}
For the case where independence of observed variables cannot be assumed, for instance due to (time) correlation or smoothness, the priors discussed before for the latent variables and driving signals do not really apply anymore, however the only change we need to make is to allow elements in rows of $\Z$ to correlate. We can assume then independent Gaussian process (GP) priors for each latent variable instead of scale mixtures of Gaussians, to obtain what we have called correlated sparse linear identifiable modeling (CSLIM). For a set of $N$ realizations of variable $j$ we set
\begin{align}
	z_{j1},\ldots,z_{jN}|\upsilon_j \ & \sim \ \GP(z_{j1},\ldots,z_{jN}|k_{\upsilon_j,n}(\cdot)) \ , \label{eq:GPj}
\end{align}
where the covariance function has the form $k_{\upsilon_j,n}(n,n')=\exp( -\upsilon_j(n-n')^2 )$, $\{n,n'\}$ is a pair of observation indices or time points and $\upsilon_j$ is the length scale controlling the overall level of correlation allowed for each variable (row) in $\Z$. Conceptually, equation \eqref{eq:GPj} implies that each latent variable $j$ is sampled from a function and the GP acts as a prior over continuous functions. Since such a length scale is very difficult to set just by looking at the data, we further place priors on $\upsilon_j$ as
\begin{align}
	\upsilon_j|u_s,\kappa \ \sim \ \Ga(\upsilon_j|u_s,\kappa) \ , \quad \kappa|k_s,k_r \ \sim \ \Ga(\kappa|k_s,k_r) \ . \label{eq:GPhyp}
\end{align}
Given that the conditional distribution of $\bupsilon=[\upsilon_1,\ldots,\upsilon_m]$ is not of any standard form, Metropolis-Hastings updates are used. In the experiments we use that $u_s=k_s=2$ and $k_r=0.02$. The details concerning inference for this model are given in Appendix \ref{ap:inf}.

It is also possible to easily expand the possible applications of GP priors in this context by, for instance, using more structured covariance functions through scale mixture of Gaussian representations to obtain a prior distribution for continuous functions with heavy-tailed behavior --- a $t$-processes \citep{yu07}, or learning the covariance function as well using inverse Wishart hyperpriors.
\subsection{Allowing for non-linearities (SNIM)} \label{sc:snim}
Provided that we know the true ordering of the variables, \ie $\P$ is known then $\B$ is surely strictly lower triangular. It is very easy to allow for non-linear interactions in the DAG model from equation \eqref{eq:PBxCz} by rewriting it as
\begin{align} \label{eq:PBfxCz}
	\P\x=(\R\odot\B)\P\y + (\Q\odot\C)\z + \bepsilon \ ,
\end{align}
where $\y=[y_1,\ldots,y_d]\ts$ and $y_{i1},\ldots,y_{iN}|\upsilon_i\sim \GP(y_{i1},\ldots,y_{iN}|k_{\upsilon_i,x}(\cdot))$ has a Gaussian process prior with for instance, but not limited to, a stationary covariance function like $k_{\upsilon_i,x}(\x,\x')=\exp(-\upsilon_i(\x-\x')^2)$, similar to equation \eqref{eq:GPj} and with the same hyperprior structure as in equation \eqref{eq:GPhyp}. This is a straight forward extension that we call sparse non-linear multivariate modeling (SNIM) that is in spirit similar to \citet{friedman00,hoyer08,zhang09,zhang09a,tillman09}, however instead of treating the inherent multiple regression problem in equation \eqref{eq:PBfxCz} and the conditional independence of the observed variables independently, we proceed within our proposed framework by letting the multiple regressor be sparse, thus the conditional independences are encoded through $\R$. The main limitation of the model in equation \eqref{eq:PBfxCz} is that if the true ordering of the variables is unknown, the exhaustive enumeration of $\P$ is needed. This means that this could be done for very small networks, \eg up to 5 or 6 variables. In principle, an ordering search procedure for the non-linear model only requires the latent variables $\z$ to have Gaussian process priors as well. The main difficulty is that in order to build covariance functions for $\z$ we need a set of observations that are not available because $\z$ is latent.
\section{Model comparison} \label{sc:ms}
Quantitative model comparison between factor models and DAGs is a key ingredient in SLIM. The joint probability of data $\X$ and parameters for the factor model part in equation \eqref{eq:PBxCz} is
\begin{equation*}
p(\X,\C,\Z,\bepsilon,\cdot)=p(\X|\C,\Z,\bepsilon)p(\C|\cdot)p(\Z|\cdot)p(\bepsilon)p(\cdot) \ ,
\end{equation*}
where $(\cdot)$ indicates additional parameters in the hierarchical model. Formally the Bayesian model selection yardstick, the marginal likelihood for model $\MOD$
$$p(\X|\MOD)=\int p(\X|\bTheta,\Z) p(\bTheta|\MOD) p(\Z|\MOD) d\bTheta d\Z \ ,$$
can be obtained by marginalizing the joint over the parameters $\bTheta$ and latent variables $\Z$. Computationally this is a difficult task because the marginal likelihood cannot be written as an average over the posterior distribution in a simple way. It is still possible using MCMC methods, for example by partitioning of the parameter space and multiple chains or thermodynamic integration \citep[see][]{chib95,neal01,murray07,friel08}, but in general it must be considered as computationally expensive and non-trivial. On the other hand, evaluating the likelihood on a test set $\X^\star$, using predictive densities $p(\X^\star|\X,\MOD)$ is simpler from a computational point of view because it can be written in terms of an average over the posterior of the {\it intensive variables}, $p(\C,\bepsilon,\cdot|\X)$ and the prior distribution of the {\it extensive variables} associated with the test points\footnote{Intensive means not scaling with the sample size. Extensive means scaling with sample size in this case the size of the test sample.}, $p(\Z^\star|\cdot)$ as
\begin{equation} \label{eq:tFA}
\mathcal{L}_\FM \deff p(\X^\star|\X,\MOD_\FM) = \int p(\X^\star|\Z^\star,\bTheta_\FM,\cdot) p(\Z^\star|\cdot) p(\bTheta_\FM,\cdot|\X) d\Z^\star d\bTheta_\FM d(\cdot) \ ,
\end{equation}
where $\bTheta_\FM=\{\C,\bepsilon\}$. This average can be approximated by a combination of standard sampling and exact marginalization using the scale mixture representation of the heavy-tailed distributions presented in Section \ref{sc:smog}. For the full DAG model in equation \eqref{eq:PBxCz}, we will not average over permutations $\P$ but rather calculate the test likelihood for a number of candidates $\P^{(1)},\ldots,\P^{(c)},\ldots$ as
\begin{align}\label{eq:tBN}
\mathcal{L}_\DAG & \deff p(\X^\star|\P^{(c)},\X,\MOD_\DAG) \ , \nonumber \\
& = \int p(\X^\star|\P^{(c)},\X,\Z^\star,\bTheta_\DAG,\cdot) p(\Z^\star|\cdot) p(\bTheta_\DAG,\cdot|\X) d\Z^\star d\bTheta_\DAG d(\cdot) \ ,
\end{align}
where $\bTheta_\DAG=\{\B,\C,\bepsilon\}$. We use sampling to compute the test likelihoods in equations \eqref{eq:tFA} and \eqref{eq:tBN}. With Gibbs, we draw samples from the posterior distributions $p(\bTheta_\FM,\cdot|\X)$ and $p(\bTheta_\DAG,\cdot|\X)$, where  $(\cdot)$ is shorthand for example for the degrees of freedom $\theta$, if Student $t$ distributions are used. The average over the extensive variables associated with the test points $p(\Z^\star|\cdot)$ is slightly more complicated because naively drawing samples from $p(\Z^\star|\cdot)$ results in an estimator with high variance --- for $\psi_i\ll\upsilon_{jn}$. Instead we exploit the infinite mixture representation to marginalize exactly $\Z^\star$ and then draw samples in turn for the scale parameters. Omitting the permutation matrices for clarity, in general we get
\begin{align*}
	p(\X^\star|\bTheta,\cdot) = & \int p(\X^\star|\Z^\star,\bTheta,\cdot)p(\Z^\star|\cdot)d\Z^\star \ , \\
	= & \prod_{n} \int \DN(\x_n^\star|\m_n,\bSigma_n)\prod_{j} p(\upsilon_{jn}|\cdot)d\upsilon_{jn} 
	\approx \frac{1}{N_\rep}\prod_{n} \sum_{r}^{N_\rep} \DN(\x_n^\star|\m_n,\bSigma_n) \ ,
\end{align*}
where $N_\rep$ is the number of samples generated to approximate the intractable integral ($N_\rep=500$ in the experiments). For the factor model $\m_n=\0$ and $\bSigma_n=\C_D\U_n\C_D\ts+\bPsi$. For the DAG, $\m_n=\B\x_n^\star$ and $\bSigma_n=\C\U_n\C\ts+\bPsi$. The covariance matrix $\U_n=\diag(\upsilon_{1n},\ldots,\upsilon_{(d+m)n})$ with elements $\upsilon_{jn}$, is sampled directly from the prior, accordingly. Once we have computed  $p(\X^\star|\bTheta_\FM,\cdot)$ for the factor model and $p(\X^\star|\bTheta_\DAG,\cdot)$ for the DAG, we can use them to average over $p(\bTheta_\FM,\cdot|\X,)$ and $p(\bTheta_\DAG,\cdot|\X)$ to obtain the predictive densities $p(\X^\star|\X,\MOD_\FM)$ and $p(\X^\star|\X,\MOD_\DAG)$, respectively.

For the particular case in which $\X$ and consequently $\Z$ are correlated variables --- CSLIM, we use a slightly different procedure for model comparison. Instead of using a test set, we randomly remove some proportion of the elements of $\X$ and perform inference with missing values, then we summarize the likelihood on the missing values. In particular, for the factor model we use $\M_{\rm miss}\odot\X=\M_{\rm miss}\odot(\Q_L\odot\C_L\Z+\bepsilon)$ where $\M_{\rm miss}$ is a binary masking matrix with zeros corresponding to test points, \ie the missing values. See details in Appendix \ref{ap:inf}. Note that this scheme is not exclusive to CSLIM thus can be also used with SLIM or when the observed data contain actual missing values.
\section{Model overview and practical details} \label{sc:inf}
The three models described in the previous section namely SLIM, CSLIM and SNIM can be summarized as a graphical model and as a probabilistic hierarchy as follows

\vspace{2mm}
\hspace{-5mm}\parbox{0.5\textwidth}{
\begin{align*}
	\x_n|\W,\y_n,\z_n,\bPsi \ \sim & \ \DN(\x_n|\W[\y_n \ \z_n]\ts,\bPsi) \ , \ \ \W=[\B \ \C] \ ,\\
	\psi\inv_i|s_s,s_r \ \sim & \ \Ga(\psi_i\inv|s_s,s_r) \ , \\
	w_{ik}|h_{ik},\psi_i,\tau_{ik} \ \sim & \ (1-h_{ik})\delta_0(w_{ik}) + h_{ik}\DN(w_{ik}|0,\psi_i\tau_{ik}) \ , \\
	h_{ik}|\eta_{ik} \ \sim & \ \Ber(h_{ik}|\eta_{ik}) \ , \ \ \H=[\R \ \Q] \ , \\
	\eta_{ik}|\nu_k,\alpha_p,\alpha_m \ \sim & \ (1-\nu_k)\delta(\eta_{ik})+\nu_k\Be(\eta_{ik}|\alpha_p\alpha_m,\alpha_p(1-\alpha_m)) \ , \\
	\nu_k|\beta_m,\beta_p \ \sim & \ \Be(\nu_k|\beta_p\beta_m,\beta_p(1-\beta_m)) \ , \\
	\tau_{ik}\inv|t_s,t_r \ \sim & \ \Ga(\tau_{ik}\inv|t_s,t_r) \ , \\
	z_{j1},\ldots,z_{jN}|\upsilon \ \sim & \begin{cases} \prod_n\DN(z_{jn}|0,\upsilon_{jn}) \ , & {\rm (SLIM)} \\ \GP(z_{j1},\ldots,z_{jN}|k_{\upsilon_j,n}(\cdot)) \ , & {\rm (CSLIM)} \end{cases} \\
	y_{i1},\ldots,y_{iN}|\upsilon \ \sim & \begin{cases} x_{i1},\ldots,x_{iN} \ , & {\rm (SLIM)} \\ \GP(y_{i1},\ldots,y_{iN}|k_{\upsilon_i,x}(\cdot)) \ , & {\rm (SNIM)} \end{cases}
\end{align*}
}\hspace{-10mm}\parbox{0.5\textwidth}{
\begin{tikzpicture}[ >= latex, font = \small, node distance = 1cm and 1cm, rounded corners = 4pt ]
	\tikzstyle{obs} = [ circle, thick, draw = black!80, fill = imp2, minimum size = 3mm ]
	\tikzstyle{lat} = [ circle, thick, draw = black!100, fill = red!0, minimum size = 3mm ]
	\tikzstyle{par} = [ circle, draw, fill = black!100, minimum width = 1pt, inner sep = 0pt ]
	\tikzstyle{every label} = [ black!100 ]
	\node [obs] (x)  [ label = -135:$x_{in}$ ] {};
	\node [lat] (B) [ above of = x, node distance = 1.5cm, label = 135:$w_{ik}$ ] {}
		edge [post] (x);
	\node [lat] (y) [ below of = x, node distance = 1.5cm, label = 135:$y_{in}$ ] {}
		edge [post] (x);
	\node [lat] (z) [ left of = x, node distance = 1.5cm and 1cm, label = -135:$z_{jn}$ ] {}
		edge [post] (x);	
	\node [lat] (upsilon) [ below of = z, label = -135:$\upsilon_{jn}$ ] {}
		edge [post] (z);
	\node [lat] (r) [ right of = B, label = 45:$h_{ik}$ ] {}
		edge [post] (B);
	\node [lat] (eta) [ above of = r, label = 45:$\eta_{ik}$ ] {}
		edge [post]  (r);
	\node [lat] (nu) [ above of = eta, node distance = 1.0cm, label = 45:$\nu_k$ ] {}
		edge [post]  (eta);
	\node [lat] (tau) [ above of = B, label = 135:$\tau_{ik}$ ] {}
		edge [post] (B);
	\node [lat] (rc) [ right of = y, label = -45:$\upsilon_i$ ] {}
		edge [post] (y);
	\node [lat] (phi) [ below of = r, node distance = 0.75cm, label = -45:$\psi_i$ ] {}
		edge [post] (x)
		edge [post] (B);
	\draw ( -0.3,-2.1 ) node {\tiny{$i=1:d$}};
	\draw ( -0.1, 4.0 ) node {\tiny{$k=1:2d+m$}};
	\draw ( -2.0, 0.6 ) node {\tiny{$n=1:N$}};
	\draw ( -1.9,-2.1 ) node {\tiny{$j=1:d+m$}};
	\begin{pgfonlayer}{background}
		\filldraw[ line width = 1pt, draw = black!50, fill = black!5 ]
			(  1.8cm, 3.1cm ) rectangle ( -0.9cm,-2.3cm )
			(  1.9cm, 4.2cm ) rectangle ( -1.0cm, 1.1cm )
			(  0.6cm, 0.8cm ) rectangle ( -2.7cm,-1.9cm )
			( -1.0cm, 0.4cm ) rectangle ( -2.8cm,-2.3cm );
	\end{pgfonlayer}
\end{tikzpicture}
}
\vspace{2mm}

\noindent where we have omitted $\P$ and the hyperparameters in the graphical model. Latent variable and driving signal parameters $\upsilon$ can have one of several priors: $\Exp(\upsilon|\lambda^2)$ (Laplace), $\Ga(\upsilon\inv|\theta/2,\theta/2)$ (Student's $t$) or $\Ga(\upsilon|u_s,\kappa)$ (GP), see equations \eqref{eq:sLad}, \eqref{eq:sStd} and \eqref{eq:GPhyp}, respectively. The latent variables/driving signals $z_{jn}$ and the mixing/connectivity matrices with elements $c_{ij}$ or $b_{ij}$ are modeled independently. Each element in $\B$ and $\C$ has its own slab variance $\tau_{ij}$ and probability of being non-zero $\eta_{ij}$. Moreover, there is a shared sparsity rate per column $\nu_k$. Variables $\upsilon_{jn}$ are variances if $z_{jn}$ use a scale mixture of Gaussian's representation, or length scales in the GP prior case. Since we assume no sparsity for the driving signals, $\eta_{ik}=1$ for $d+i=k$ and $\eta_{ik}=0$ for $d+i\neq k$. In addition, we can recover the pure DAG by making $m=0$ and the standard factor model by making instead $\eta_{ik}=0$ for $k\leq 2d$. All the details for the Gibbs sampling based inference are summarized in appendix \ref{ap:inf}.
\subsection{Proposed workflow} 
We propose the workflow shown in Figure \ref{fg:slimfig} to integrate all elements of SLIM, namely factor model and DAG inference, stochastic order search and model selection using predictive densities.
\begin{enumerate}{\leftmargin=1em}
	\item Partition the data into $\{\X,\X^\star\}$.
	\item Perform inference on the factor model and stochastic order search. One Gibbs sampling update consists of computing the conditional posteriors in equations \eqref{eq:gibbsPsi}, \eqref{eq:gibbsZ}, \eqref{eq:gibbsD}, \eqref{eq:gibbstau}, \eqref{eq:gibbsq}, \eqref{eq:gibbseta} and \eqref{eq:gibbsnu} in sequence, followed by several repetitions (we use 10) of the M-H update in equation \ref{eq:rlik} for the permutation matrices $\P$ and $\PC$. 
	\item Summarize the factor model, mainly $\C$, $\{\eta_{ij}\}$ and $\mathcal{L}_\FM$ using quantiles (0.025, 0.5 and 0.975).
	\item Summarize the orderings, $\P$. Select the top $m_\topc$ candidates according to their frequency during inference in step 2.
	\item Perform inference on the DAGs for each one of the ordering candidates, $\P^{(1)},\ldots,\P^{(m_\topc)}$ using Gibbs sampling by computing equations \eqref{eq:gibbsPsi}, \eqref{eq:gibbsZ}, \eqref{eq:gibbsD}, \eqref{eq:gibbstau}, \eqref{eq:gibbsq}, \eqref{eq:gibbseta} and \eqref{eq:gibbsnu} in sequence, up to minor changes described in Appendix \ref{ap:inf}.
	\item Summarize the DAGs, $\B$, $\C_L$, $\{\eta_{ik}\}$ and $\mathcal{L}_\DAG^{(1)},\ldots,\mathcal{L}_\DAG^{(m_\topc)}$ using quantiles (0.025, 0.5 and 0.975). Note that $\{\eta_{ik}\}$ contains non-zero probabilities for $\R$ and $\Q$ corresponding to $\B$ and $\C_L$, respectively.
\end{enumerate}
We use medians to summarize all quantities in our model because $\D$, $\B$ and $\{\eta_{ik}\}$ are bimodal while the remaining variables are in general skewed posterior distributions. Inference with GP priors for time series data (CSLIM) or non-linear DAGs (SNIM) is fairly similar to the \iid case, see Appendix \ref{ap:inf} for details. Source code for SLIM and all its variants proposed so far has been made available at \url{http://cogsys.imm.dtu.dk/slim/} as Matlab scripts.
\subsection{Computational cost} 
The cost of running the linear DAG with latent variables or the factor model is roughly the same, \ie $\mathcal{O}(N_sd^2N)$ where $N_s$ is the total number of samples including the burn-in period. The memory requirements on the other hand are approximately $\mathcal{O}(N_pd^2)$ if all the samples after the burn-in period $N_p$ are stored. This means that the inference procedures scale reasonably well if $N_s$ is kept in the lower ten thousands. The non-linear version of the DAG is considerably more expensive due to the GP priors, hence the computational cost rises up to $\mathcal{O}(N_s(d-1)N^3)$.

The computational cost of LiNGAM, being the closest to our linear models, is mainly dependent on the statistic used to prune/select the model. Using bootstrapping results in $\mathcal{O}(N_b^3)$, where $N_b$ is the number of bootstrap samples. The Wald statistic leads to $\mathcal{O}(d^6)$, while Wald with $\chi^2$ second order model fit test amounts to $\mathcal{O}(d^7)$. As for the memory requirements, bootstrapping is very economic whereas Wald-based statistics require $\mathcal{O}(d^6)$.

\begin{wrapfigure}{r}{0.35\textwidth}
	\vspace{-6pt}
	\centering
	\begin{psfrags}
		 \psfrag{time}[c][c][0.65]{Time}\psfrag{d}[c][c][0.65]{$d$}\psfrag{bootstrap}[l][l][0.55]{Bootstrap}\psfrag{wald}[l][l][0.55]{Wald}\psfrag{slim}[l][l][0.55]{SLIM}
		\includegraphics[width=0.33\textwidth]{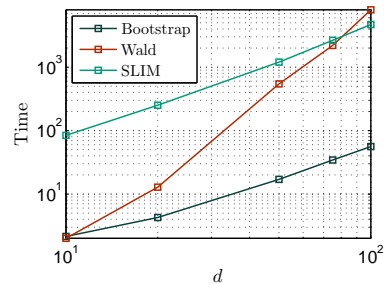}
	\end{psfrags}
	\vspace{-10pt}
	\caption{Runtime comparison.}
	\label{fg:rt_all}
	\vspace{0pt}
\end{wrapfigure}
The method for non-linear DAGs described in \citet{hoyer08} is defined for a pair of variables, and it uses GP-based regression and kernelized independence tests. The computational cost is $\mathcal{O}(N_gN^3)$ where $N_g$ is the number of gradient iterations used to maximize the marginal likelihood of the GP. This is the same order of complexity as our non-linear DAG sampler.

Figure \ref{fg:rt_all} shows average running times in a standard desktop machine (two cores, 2.6GHz and 4Gb RAM) over 10 different models with $N=1000$ and $d=\{10,20,50,100\}$. As expected, LiNGAM with bootstrap is very fast compared to the others while our model approaches LiNGAM with Wald statistic as the number of observations increases. We did not include LiNGAM with second order model fit because for $d=50$ it is already prohibitive. For this small test we used a C implementation of our model with $N_s=19000$. We are aware that the performance of a C and a Matlab implementation can be different, however we still do the comparison because the most expensive operations in the Matlab code for LiNGAM are computed through BLAS routines not involving large loops, thus a C implementation of LiNGAM should not be noticeably faster than its Matlab counterpart.
\section{Simulation results} \label{sc:res}
We consider six sets of experiments to illustrate the features of SLIM. In our comparison with other methods we focus on the DAG structure learning part because it is somewhat easier to benchmark a DAG than a factor model. However, we should stress that DAG learning is just one component of SLIM. Both types of model and their comparison are important, as will be illustrated through the experiments. For the reanalysis of flow cytometry data using our models, quantitative model comparison favors the DAG with latent variables rather than the standard factor model or the pure DAG which was the paradigm used in the structure learning approach of \citet{sachs05}.

The first two experiments consist of extensive tests using artificial data in a setup originally from LiNGAM and network structures taken from the Bayesian net repository. We test the features of SLIM and compare with LiNGAM and some other methods in settings where they have proved to work well. The third set of experiments addresses model comparison, the fourth and fifth present results for our DAG with latent variables and the non-linear DAG (SNIM) on both artificial and real data. The sixth uses real data previously published by \citet{sachs05} and the last one provides simple results for a factor model using Gaussian process priors for temporal smoothness (CSLIM), tested on a time series gene expression data set \citep{kao04}. In all cases we ran 10000 samples after a burn-in period of 5000 for the factor model, and a single chain with 3000 samples and 1000 as burn-in iterations for the DAG, \ie $N_s=19000$ used in the computational cost comparison. As a summary statistic we use median values everywhere, and Laplace distributions for the latent factors if not stated otherwise.
\subsection{Artificial data}
We evaluate the performance of our model against LiNGAM\footnote{Matlab package (v.1.42) available at \url{http://www.cs.helsinki.fi/group/neuroinf/lingam/}.}, using the artificial model generator presented and fully explained in \citet{shimizu06}. Concisely, the generator produces both dense and sparse networks with different degrees of sparsity, $\Z$ is generated from a heavy-tailed non-Gaussian distribution through a generalized Gaussian distribution with zero mean, unit variance and random shape, $\X$ is generated recursively using equation \eqref{eq:PBxCz} with $m=0$ and then randomly permuted to hide the correct order, $\P$. Approximately, half of the networks are fully connected while the remaining portion comprises sparsity levels between $10\%$ and $80\%$. Having dense networks ($0\%$ sparsity) in the benchmark is crucial because in such cases the correct order of the variables is unique, thus more difficult to find. This setup is particularly challenging because the model needs to identify both dense and sparse models. For the experiment we have generated $1000$ different dataset/models using $d=\{5,10\}$, $N=\{200,500,1000,2000\}$ and the DAG was selected using the median of the training likelihood, $p(\X|\PL^{(k)},\R^{(k)},\B^{(k)},\C_D^{(k)},\Z,\bPsi,\cdot)$, for $k=1,\ldots,m_\topc$.
\begin{figure}[t]
	\centering
	\begin{psfrags}
		 \psfrag{N}[c][c][0.6]{$N$}\psfrag{LINGAM}[l][l][0.45]{LiNGAM}\psfrag{LSLIM}[l][l][0.45]{DS}\psfrag{DSLIM}[l][l][0.45]{DENSE}\psfrag{SLIM}[l][l][0.45]{SLIM}\psfrag{cand}[c][c][0.6]{Candidate}\psfrag{prop}[c][c][0.6]{Correct ordering rate}\psfrag{200}[c][c][0.5]{200}\psfrag{500}[c][c][0.5]{500}\psfrag{1000}[c][c][0.5]{1000}\psfrag{2000}[c][c][0.5]{2000}\psfrag{5000}[l][l][0.5]{5000}\psfrag{d=5}[l][l][0.5]{$d=5$}\psfrag{d=10}[l][l][0.5]{$d=10$}
		\subfigure[]{\includegraphics[scale = 0.37]{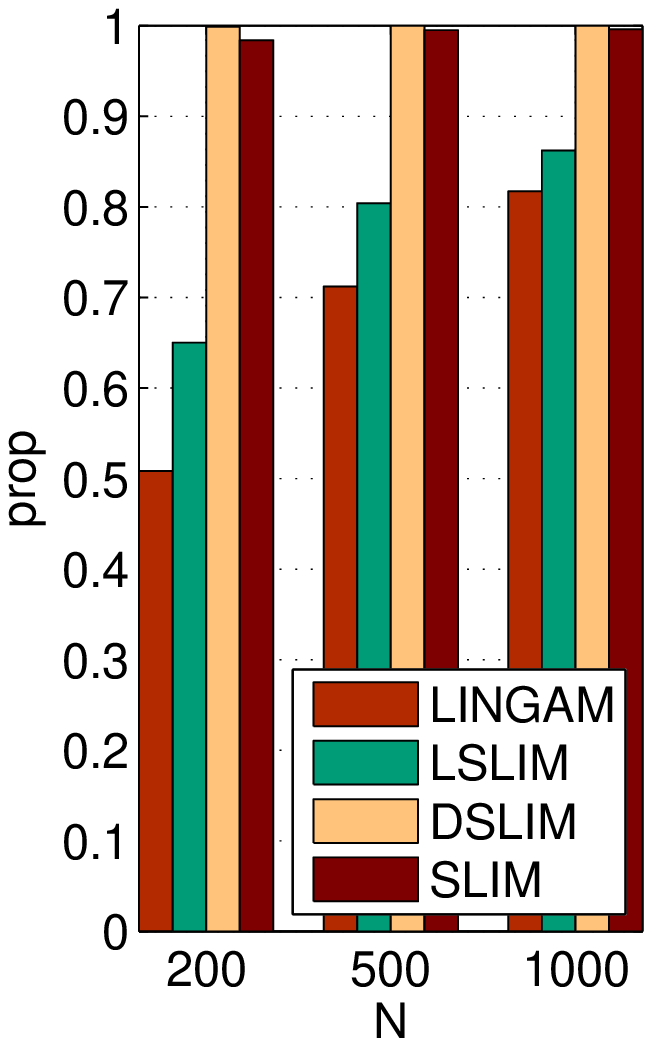}\label{fg:oerr5}}
		\subfigure[]{\includegraphics[scale = 0.37]{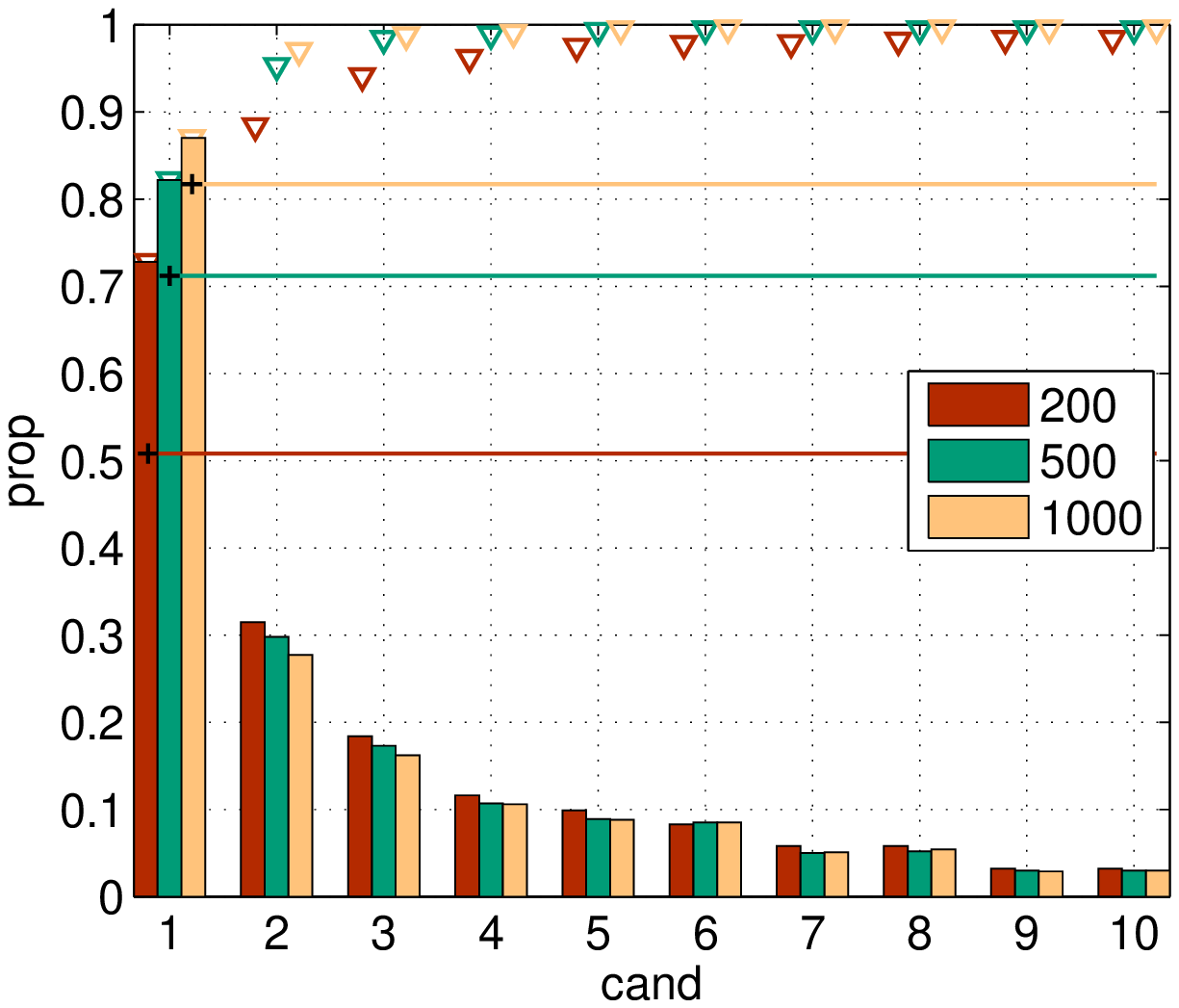}\label{fg:oerr5c}}
		\subfigure[]{\includegraphics[scale = 0.37]{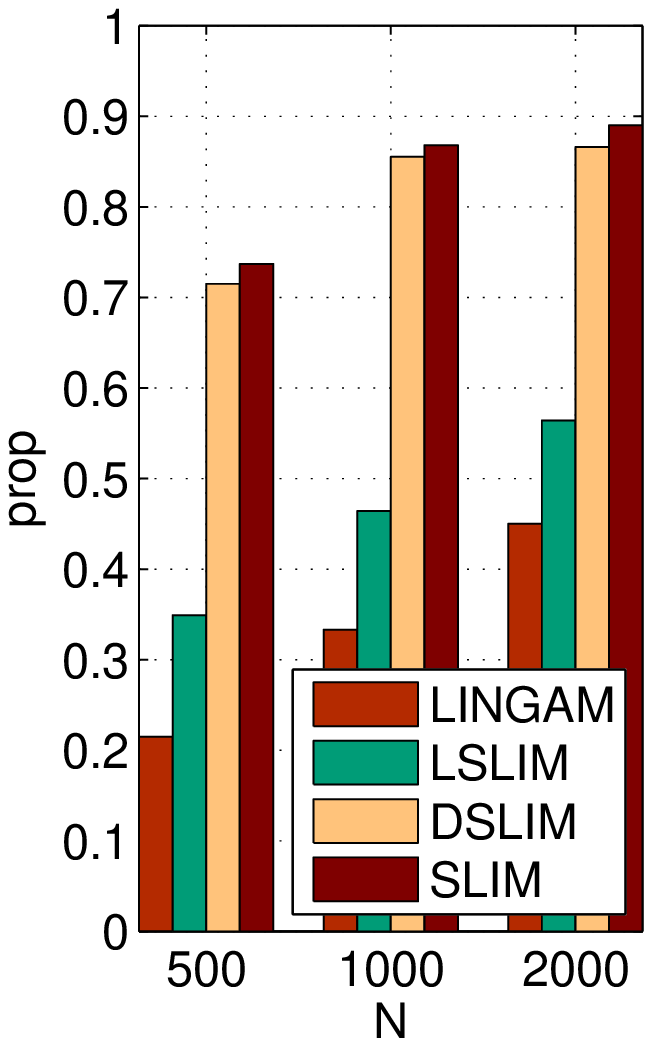}\label{fg:oerr10}}
		\subfigure[]{\includegraphics[scale = 0.37]{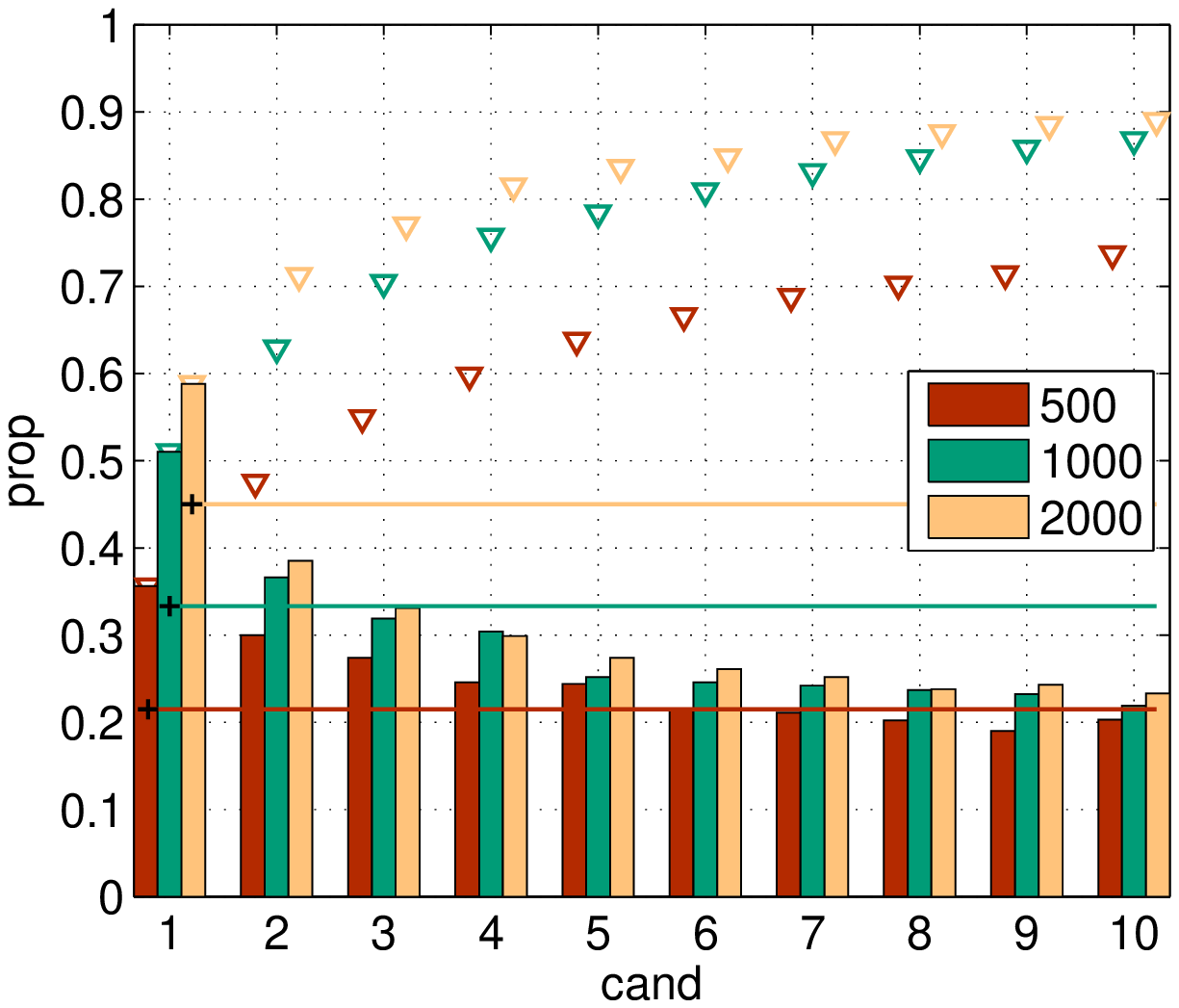}\label{fg:oerr10c}}
	\end{psfrags}
	\caption{Ordering accuracies for LiNGAM suite using $d=5$ in (a,b) and $d=10$ in (c,d). (a,c) Total correct ordering rates where DENSE is our factor model without sparsity prior and DS corresponds to DENSE but using the deterministic ordering search used in LiNGAM. (b,c) Correct ordering rate vs. candidates from SLIM. The crosses and horizontal lines correspond to LiNGAM while the triangles are accumulated correct orderings across candidates used by SLIM.} \label{fg:lingam_oerrs}
\end{figure}
\paragraph{Order search.} With this experiment we want to quantify the impact of using sparsity, stochastic ordering search and more than one ordering candidate, \ie $m_\topc=10$ in total. Figure \ref{fg:lingam_oerrs} evaluates the proportion of correct orderings for different settings. We have the following abbreviations for this experiment, DENSE is our factor model without sparsity prior, \ie assuming that $p(r_{ij}=1)=1$ a priori. DS (deterministic search) assumes no sparsity as in DENSE but replaces our stochastic search for permutations with the deterministic approach used by LiNGAM, \ie we replace the M-H update from equation \eqref{eq:rlik} by the procedure described next: after inference we compute $\D\inv$ followed by a column permutation search using the Hungarian algorithm and a row permutation search by iterative pruning until getting a version of $\D$ as triangular as possible \citep{shimizu06}. Several comments can be made from the results, (i) For $d=5$ there is no significant gain for increasing $N$, mainly because the size of the permutation space is small, \ie $5!$. (ii) The difference in performance between SLIM and DENSE is not significative because we look for triangular matrices in a probabilistic sense, hence there is no real need for exact zeros but just very small values, this does not mean that the sparsity in the factor model is unnecessary, on the contrary we still need it if we want to have readily interpretable mixing matrices. (iii) Using more than one ordering candidate considerably improves the total correct ordering rate, \eg by almost $30\%$ for $d=5, \ N=200$ and $35\%$ for $d=10, \ N=500$. (iv) The number of accumulated correct orderings found saturates as the number of candidates used increases, suggesting that further increasing $m_\topc$ will not considerably change the overall results. (v) The number of correct orderings tends to accumulate on the first candidate when $N$ increases since the uncertainty of the estimation of the parameters in the factor model decreases accordingly. (vi) When the network is not dense, it could happen that more than one candidate has a correct ordering, hence the total rates (triangles) are not just the sum of the bar heights in Figures \ref{fg:oerr5c} and \ref{fg:oerr10c}. (vii) It seems that except for $d=10, \ N=5000$ it is enough to consider just the first candidate in SLIM to obtain as many correct orderings as LiNGAM does. (viii) From Figures \ref{fg:oerr5} and \ref{fg:oerr10}, the three variants of SLIM considered perform better than LiNGAM, even when using the same single candidate ordering search proposed by \citet{shimizu06}. (ix) In some cases the difference between SLIM and LiNGAM is very large, for example, for $d=10$ using two candidates and $N=1000$ is enough to obtain as many correct orderings as LiNGAM with $N=5000$.
\begin{figure}[t]
	\centering
	\begin{psfrags}
		\psfrag{TP}[c][c][0.6]{True positive rate (sensitivity)}\psfrag{FP}[c][c][0.6]{False positive rate (1 - specificity)}\psfrag{olsboot}[l][l][0.5]{Bootstrap}\psfrag{wald}[l][l][0.5]{Wald}\psfrag{bonferroni}[l][l][0.5]{Bonferroni}\psfrag{modelfit}[l][l][0.5]{Wald + $\chi^2$}\psfrag{slim}[l][l][0.5]{SLIM}\psfrag{bslim}[l][l][0.5]{ORACLE}
		\subfigure[$d=5, \ N=200$]{\includegraphics[scale = 0.43]{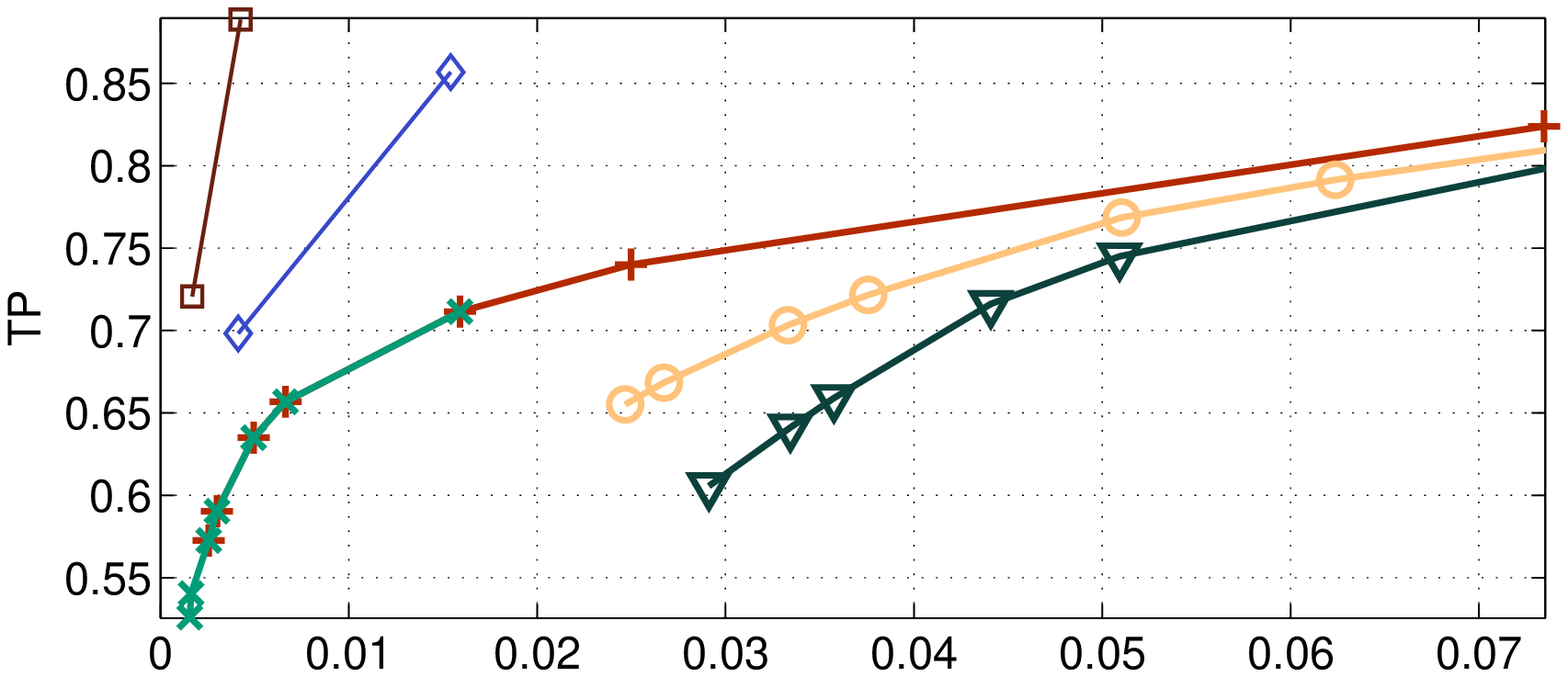}}
		\subfigure[$d=10, \ N=500$]{\includegraphics[scale = 0.43]{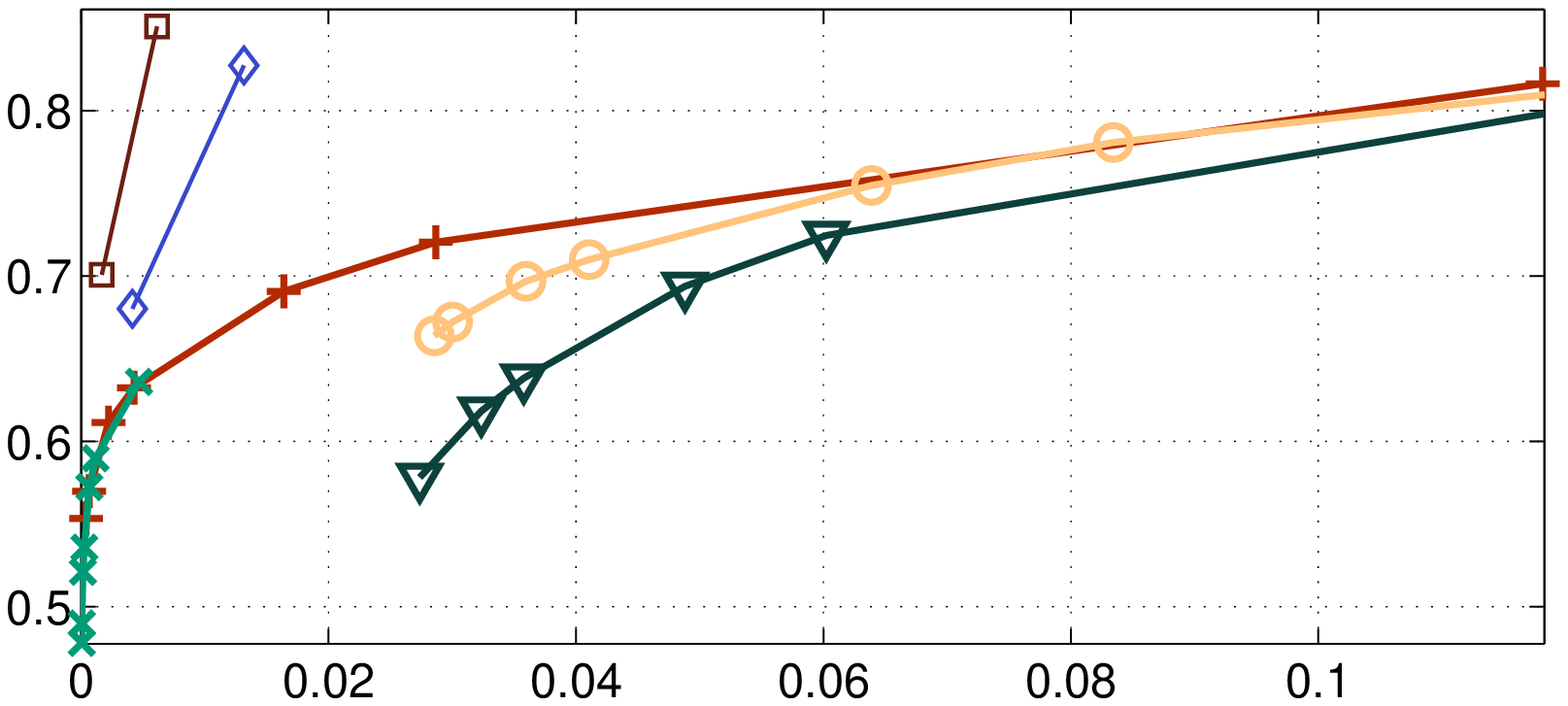}}
		\subfigure[$d=5, \ N=500$]{\includegraphics[scale = 0.43]{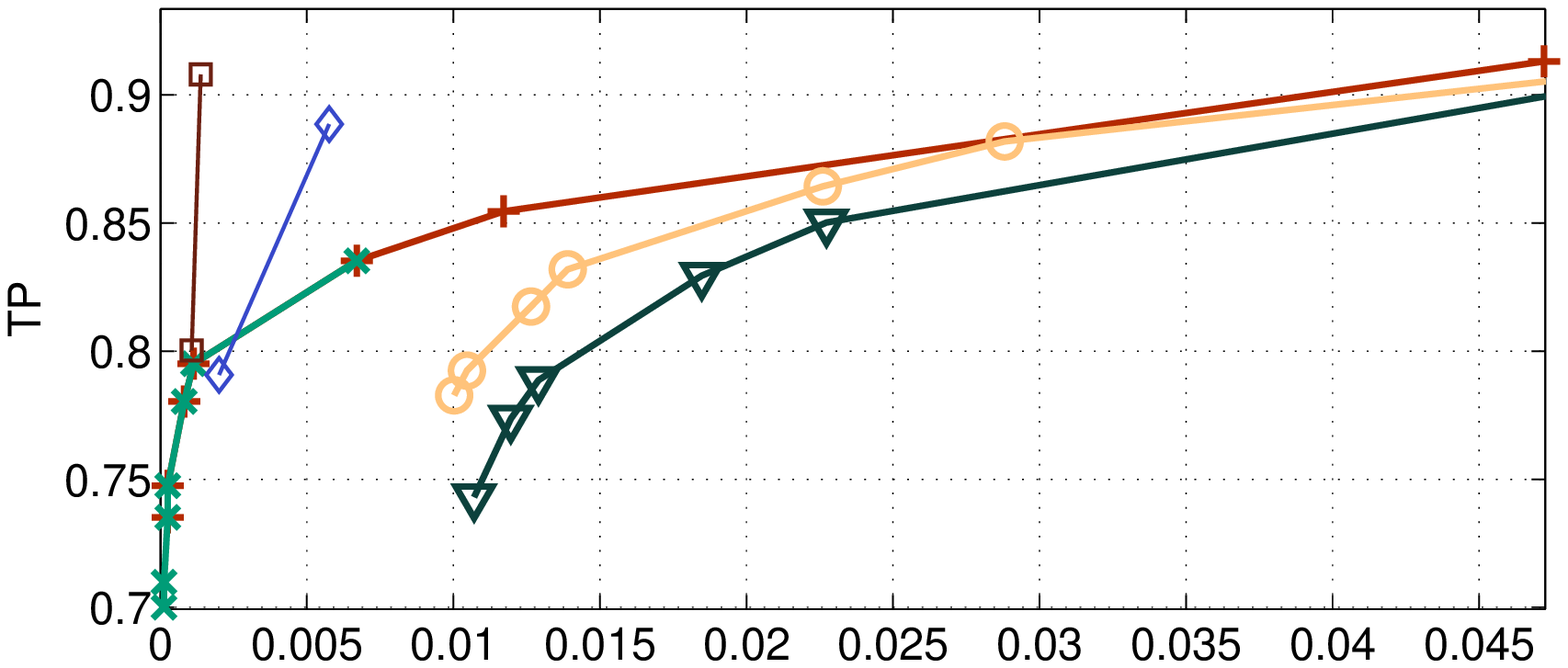}}
		\subfigure[$d=10, \ N=1000$]{\includegraphics[scale = 0.43]{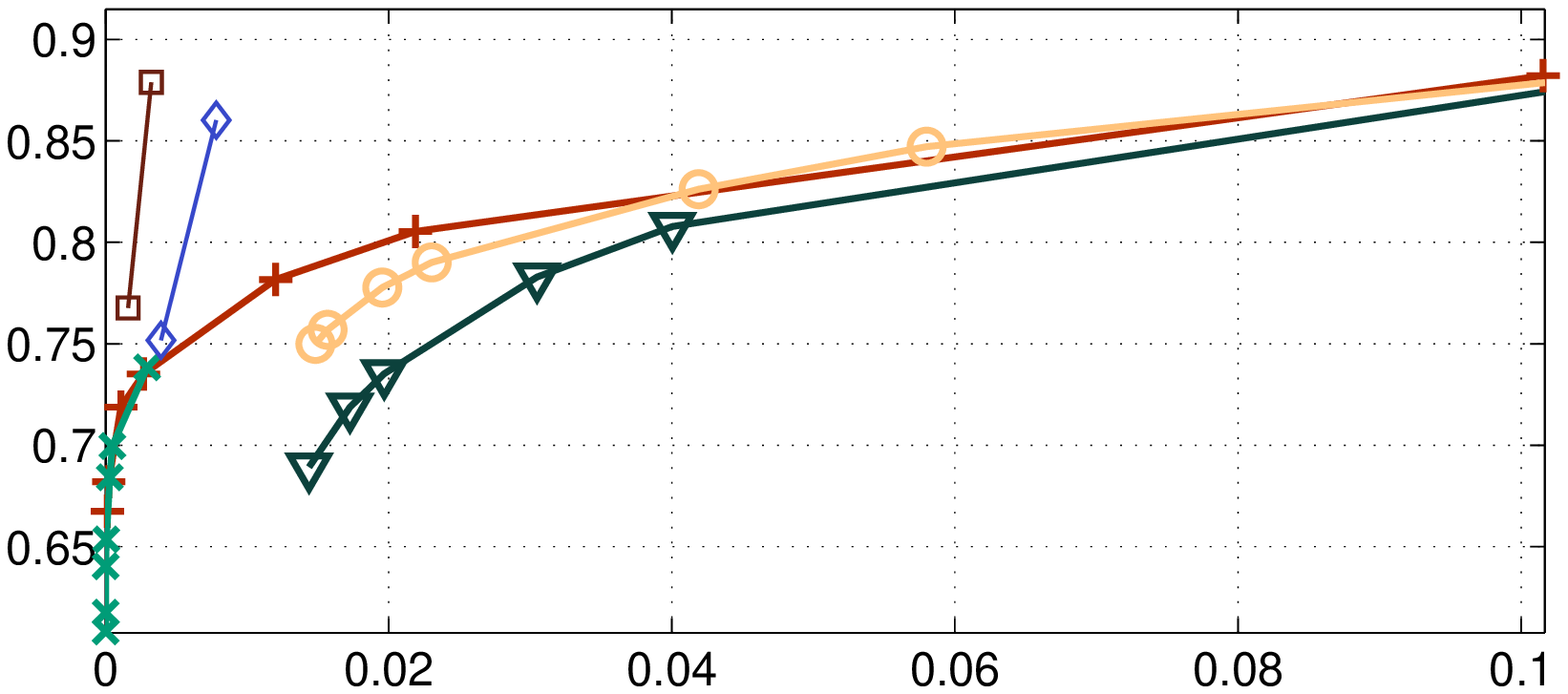}}
		\subfigure[$d=5, \ N=1000$]{\includegraphics[scale = 0.43]{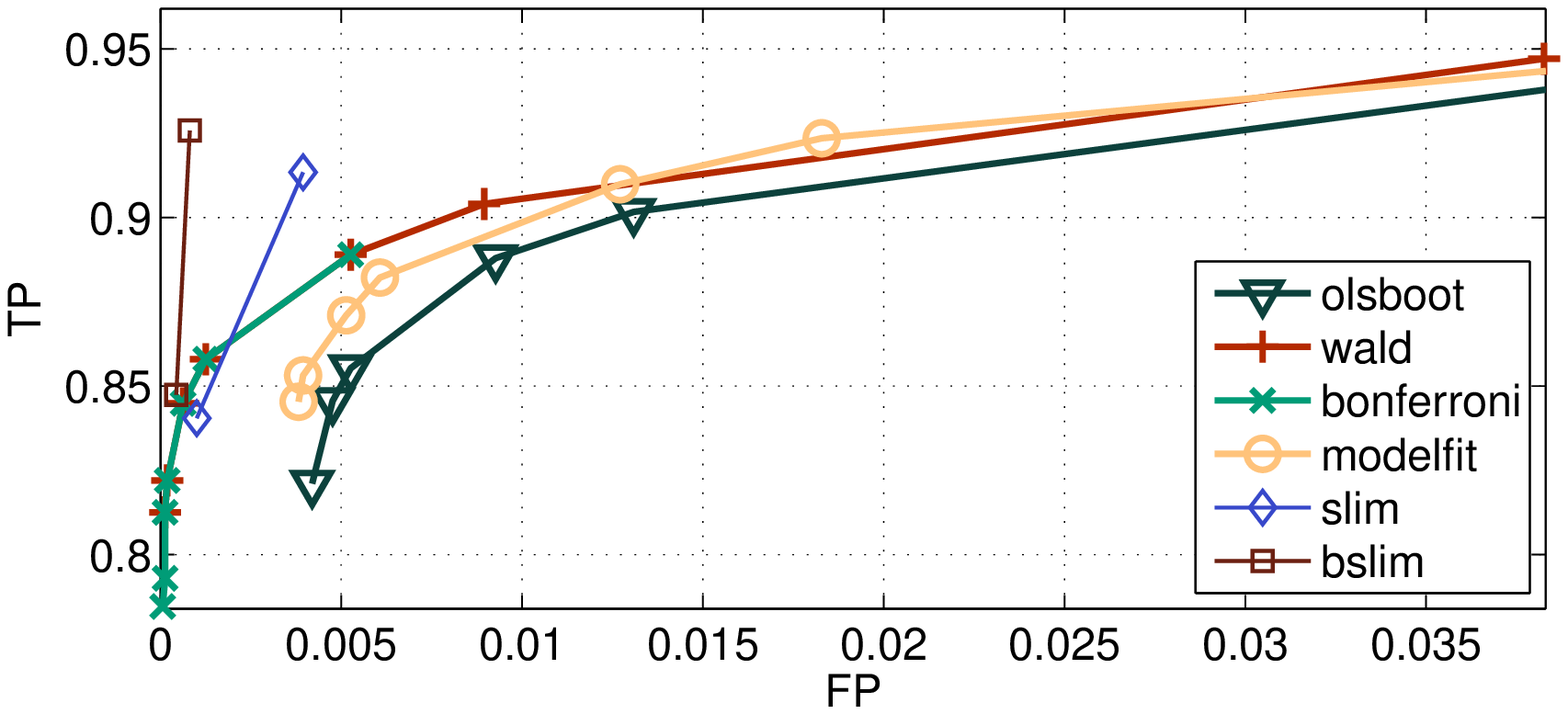}}
		\subfigure[$d=10, \ N=2000$]{\includegraphics[scale = 0.43]{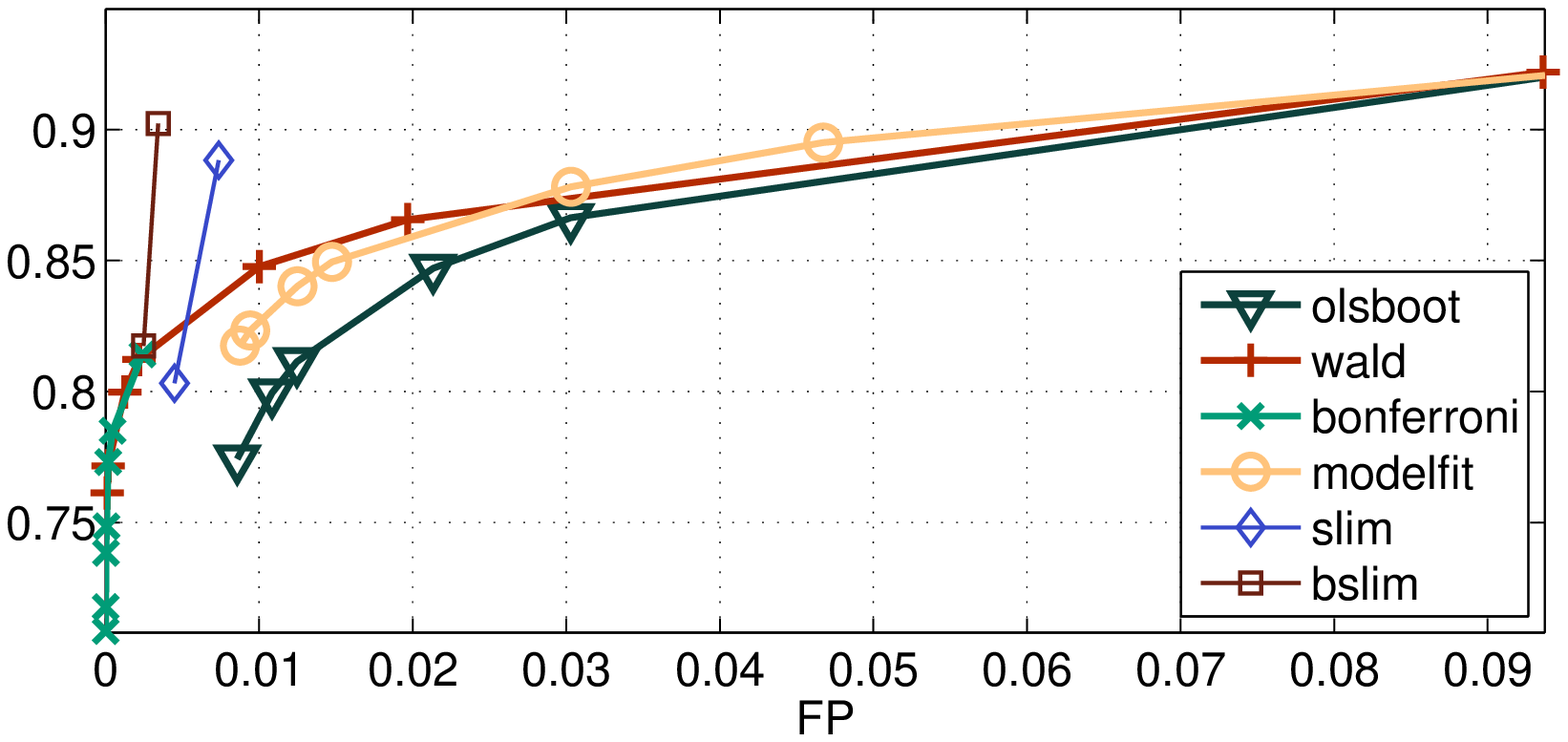}}
	\end{psfrags}
	\caption{Performance measures for LiNGAM suite. Results include the settings: $d=\{5,10\}$, $N=\{200,500,1000,2000\}$, four model selectors for LiNGAM (bootstrap, Wald, Bonferroni and Wald + $\chi^2$ statistics) and seven $p$-value cutoffs for the statistics used in LiNGAM (0.0005, 0.001, 0.005, 0.01, 0.05, 0.1, 0.5). ORACLE corresponds to oracle results for SLIM, both computed for two settings: diffuse $\beta_m=0.99$ and sparse $\beta_m=0.1$ priors. Markers close to the top-left corner denote better results in average.} \label{fg:lingam_suite}
\end{figure}
\paragraph{DAG learning.} Now we evaluate the ability of our model to capture the DAG structure in the data, provided the permutation matrices obtained in the previous stage as a result of our stochastic order search. Results are summarized in Figure \ref{fg:lingam_suite} using receiving operating characteristic (ROC) curves. The true and false positive rates are averaged over the number of trials (1000) for each setting to make the scaling in the plots more meaningful given the various levels of sparsity considered. The rates are computed in the usual way, however it must be noted that the true number of absent links in a network can be as large as $d(d-1)$, \ie twice the number of links in a DAG, because in the case of an estimated DAG based in a wrong ordering the number of false positives can sum up to $d(d-1)/2$ even if the true network is not empty. For LiNGAM we use four different statistics to prune the DAG after the ordering has been found, namely bootstrapping, Wald, Bonferroni and Wald with second order $\chi^2$ model fit test. In every case we run LiNGAM for 7 different $p$-value cutoffs, namely, 0.0005, 0.001, 0.005, 0.01, 0.05, 0.1 and 0.5 to build the ROC curve. For SLIM we consider the two settings for $\beta_m$ discussed in Section \ref{sc:sse}, \ie a diffuse prior supporting the existence of dense graphs, $\beta_m=0.99$ and $\beta_m=0.1$. In order to test how good SLIM is at selecting one DAG out of the $m_\topc$ candidates, we also report the oracle results under the name of ORACLE, where in every case we select the candidate with less error instead of $\argmax_k \  p(\X|\PL^{(k)},\R^{(k)},\B^{(k)},\C_D^{(k)},\Z,\bPsi,\cdot)$. Using $\beta_m=0.99$ is not very useful in practice because in a real situation we expect that the underlying DAG is sparse, however the LiNGAM suite has as many dense graphs as sparse ones making $\beta_m=0.1$ a poor choice. From Figure \ref{fg:lingam_suite}, it is clear that for $\beta_m=0.99$, SLIM is clearly superior, providing the best true positive rate (TPR) - false positive rate (FPR) tradeoff. For $\beta_m=0.1$ there is no real difference between SLIM and some settings of LiNGAM (Wald and Bonferroni). Concerning SLIM's model selection procedure, it can be seen that the difference between SLIM and ORACLE nicely decreases as the number of observations increases. We also tested the DAG learning procedure in SLIM when the true ordering is known (results not shown) and we found only a very small difference compared to ORACLE. It is important to mention that further increasing or reducing $\beta_m$ does not significantly change the results shown; this is because $\beta_m$ does not fully control the sparsity of the model, thus even for $\beta_m=1$ the model will be still sparse due to element-wise link confidence, $\alpha_m$. As for LiNGAM, it seems that Wald performs better than Wald $+ \ \chi^2$, however just by looking at Figure \ref{fg:lingam_suite}, it is to be expected that for larger $N$ the latter perform better because the Wald statistic alone will tend to select more dense models.
\begin{figure}[p]
	\centering
	\begin{psfrags}
	 \psfrag{x1}[c][c][0.6]{$x_1$}\psfrag{x2}[c][c][0.6]{$x_2$}\psfrag{x3}[c][c][0.6]{$x_3$}\psfrag{x4}[c][c][0.6]{$x_4$}\psfrag{x5}[c][c][0.6]{$x_5$}\psfrag{x6}[c][c][0.6]{$x_6$}\psfrag{x7}[c][c][0.6]{$x_7$}\psfrag{x8}[c][c][0.6]{$x_8$}\psfrag{x9}[c][c][0.6]{$x_9$}\psfrag{x10}[c][c][0.6]{$x_{10}$}\psfrag{fac}[c][c][0.6]{Factors}\psfrag{1.00}[c][c][0.6]{}\psfrag{cand}[t][c][0.6]{Orderings}\psfrag{freq}[b][c][0.6]{Frequency ($\%$)}\psfrag{var}[t][c][0.6]{Variables}\psfrag{bij}[c][c][0.7]{$b_{ij}$}\psfrag{etaij}[c][c][0.7]{$\eta_{ij}$}\psfrag{Candidates}[c][c][0.45]{Candidates}\psfrag{prob}[c][c][0.6]{$p(r_{ij}=1|\X,\cdot)$}\psfrag{mag}[b][c][0.6]{Magnitude}
		\subfigure[]{\includegraphics[scale=0.42]{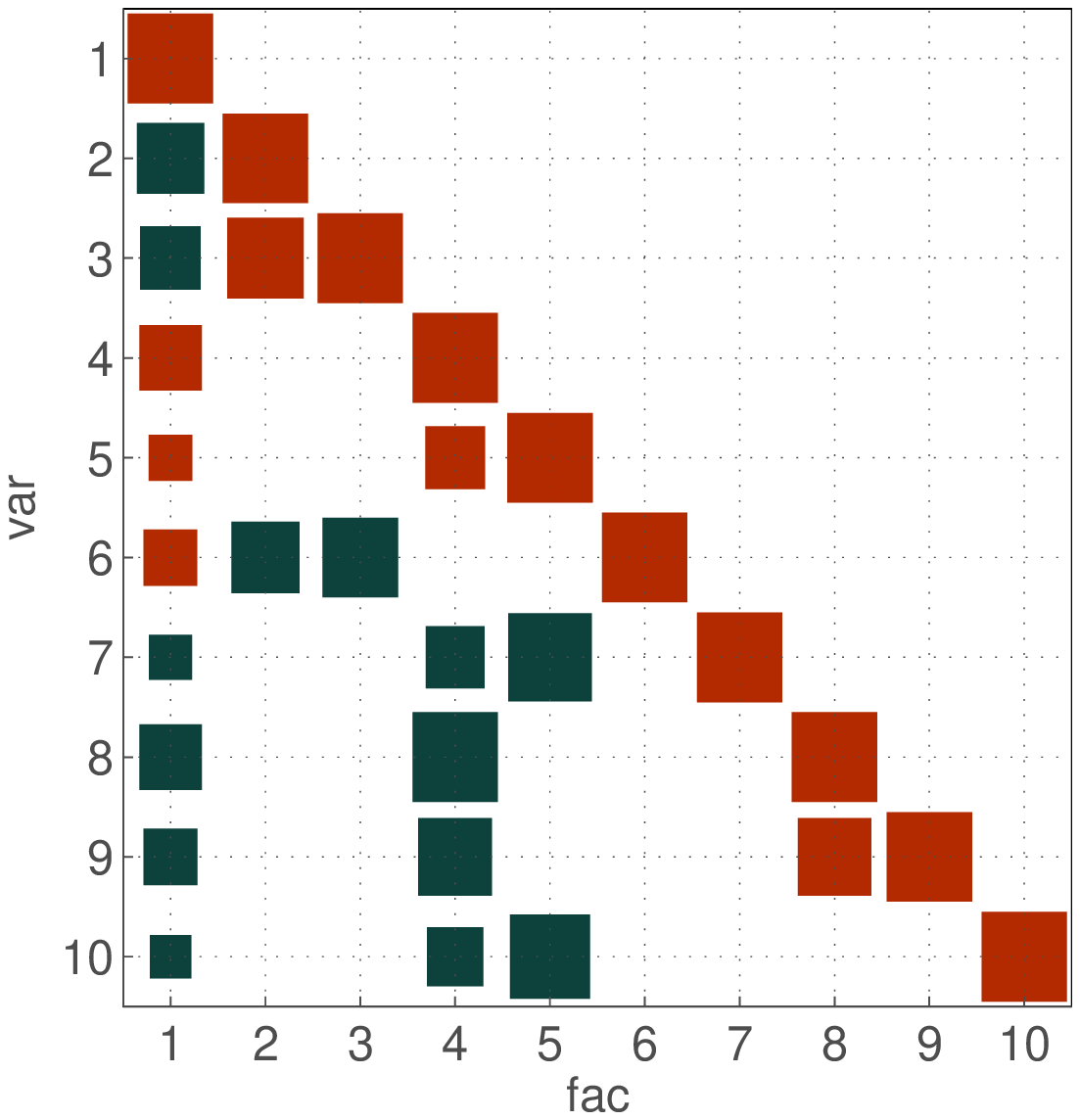}\label{fg:singletrue_A}}
		\subfigure[]{\includegraphics[scale=0.42]{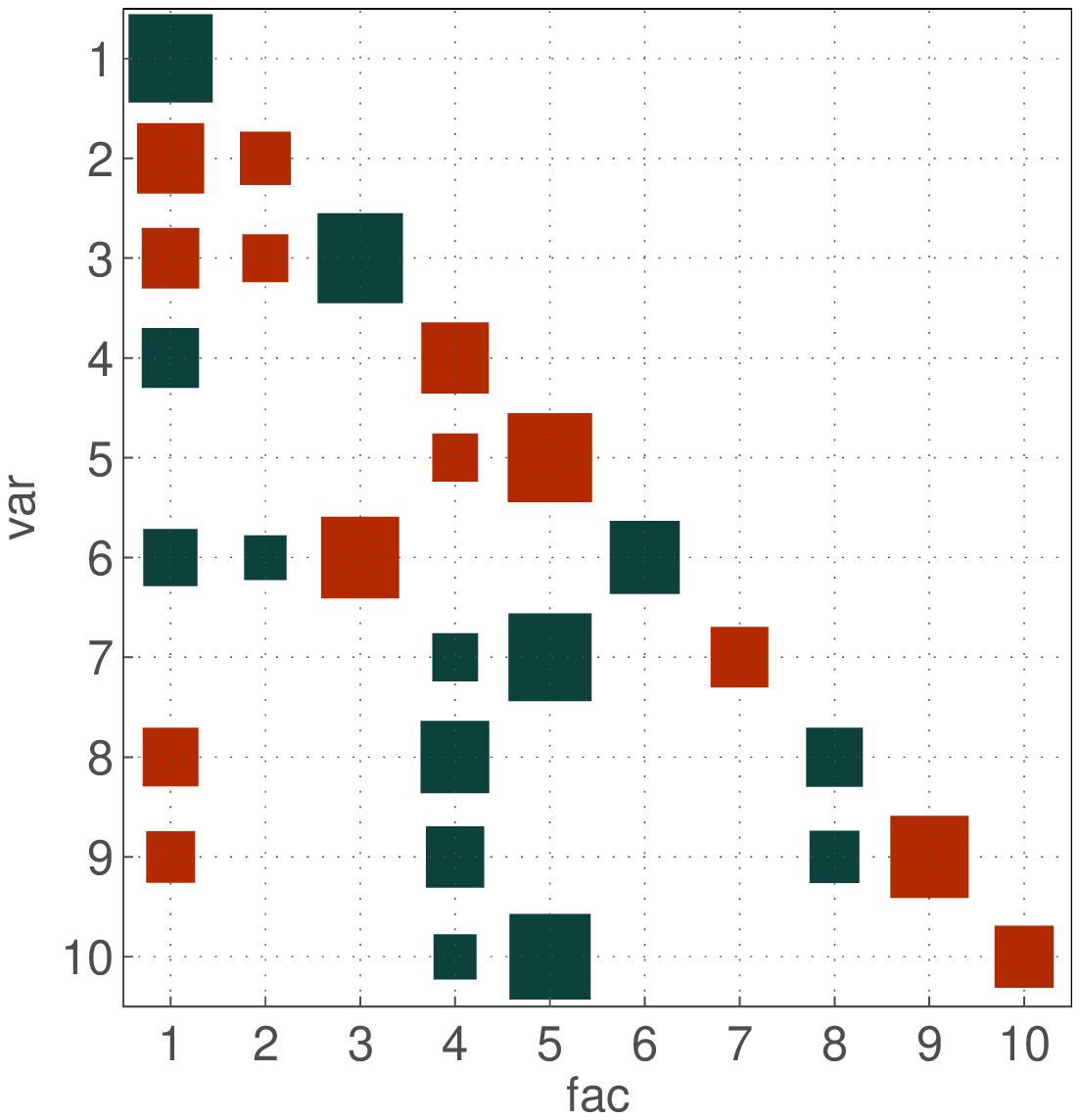}\label{fg:singlesFA_A}}
		\subfigure[]{\includegraphics[scale=0.379]{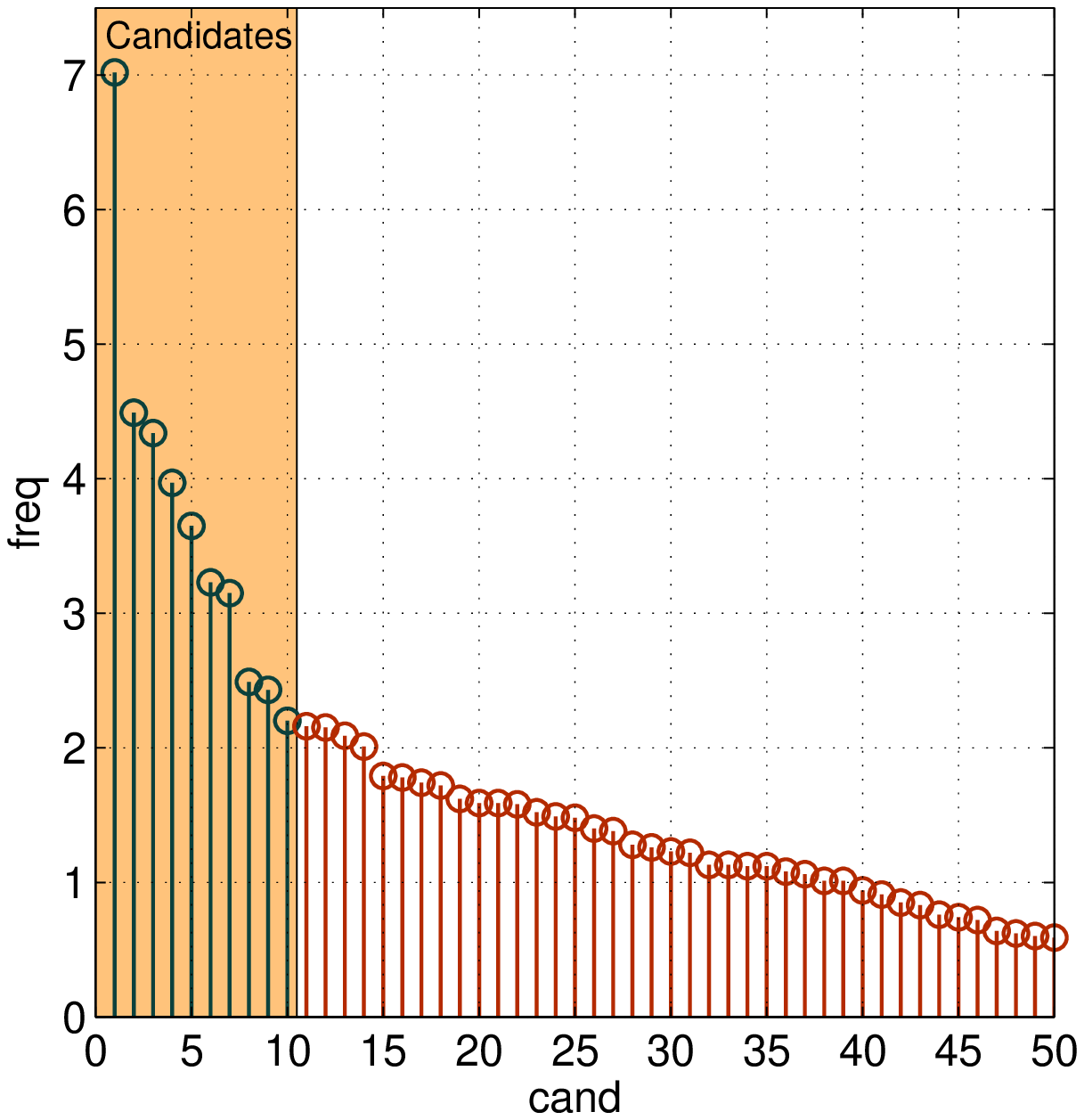}\label{fg:singlecand}}
		\subfigure[]{\includegraphics[scale=0.4]{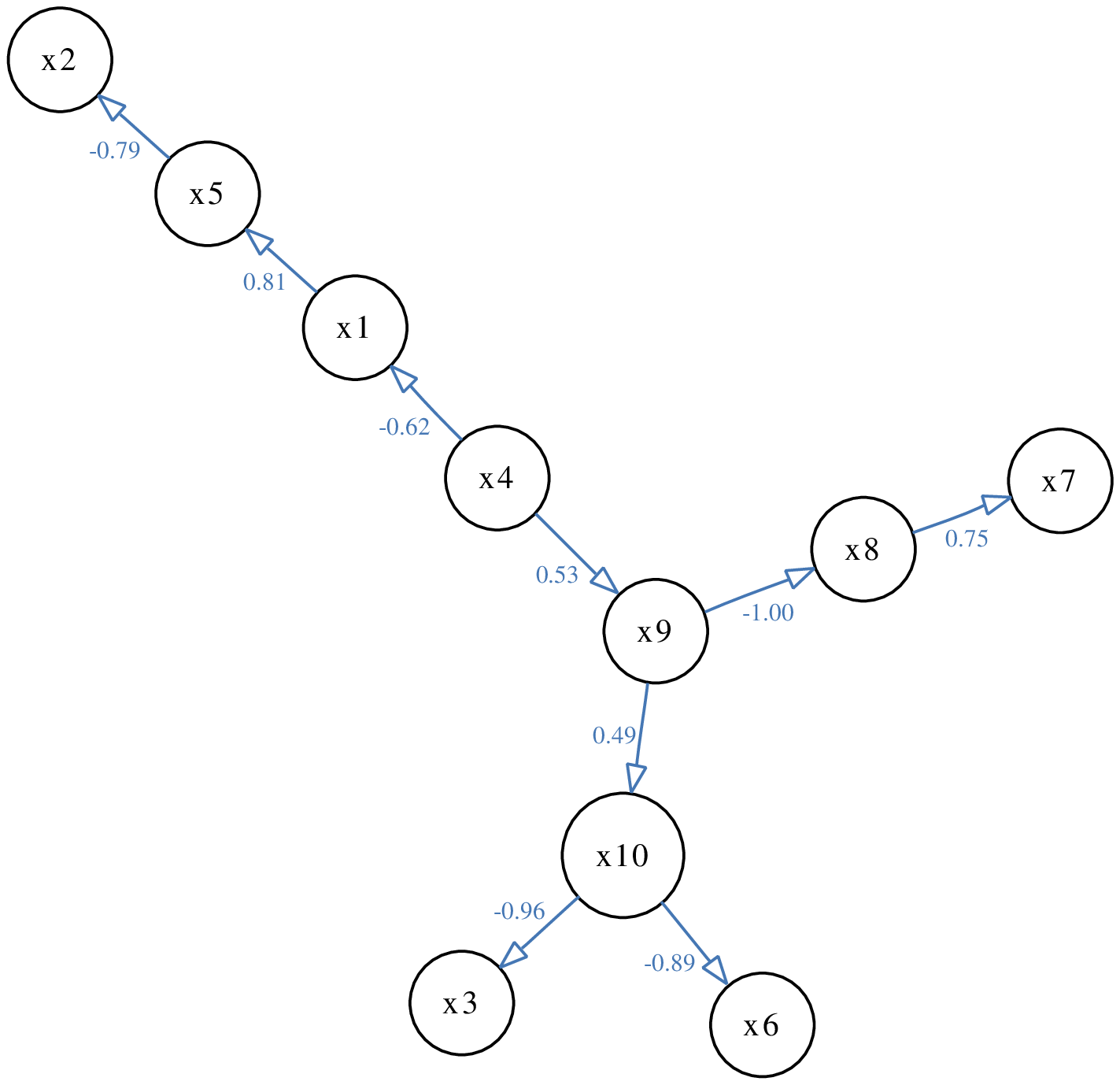}\label{fg:singletrue_B}}
		\subfigure[]{\includegraphics[scale=0.4]{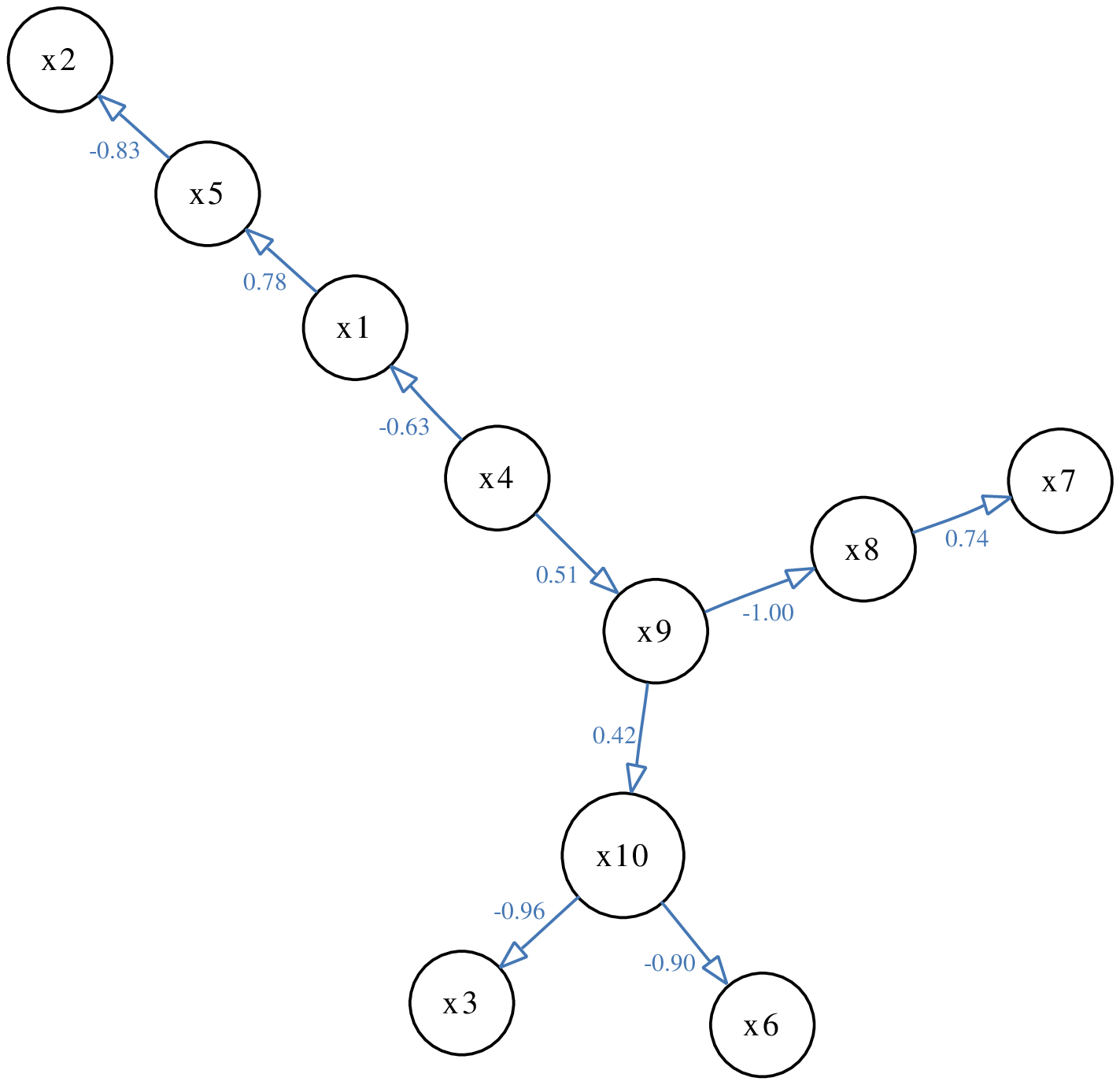}\label{fg:singlesFA_B}}
		\subfigure[]{\includegraphics[scale=0.46]{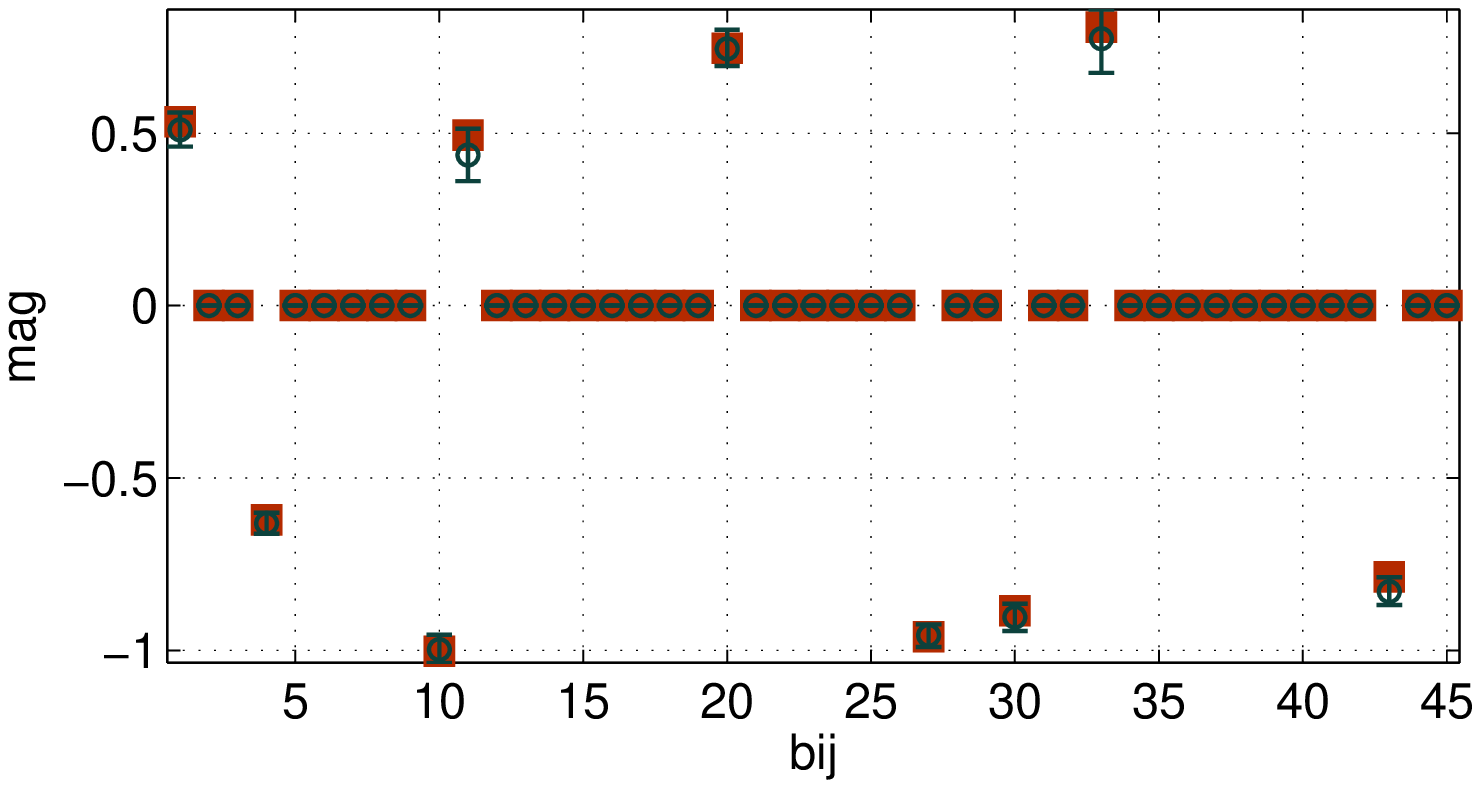}\label{fg:singleebars}}
		\subfigure[]{\includegraphics[scale=0.46]{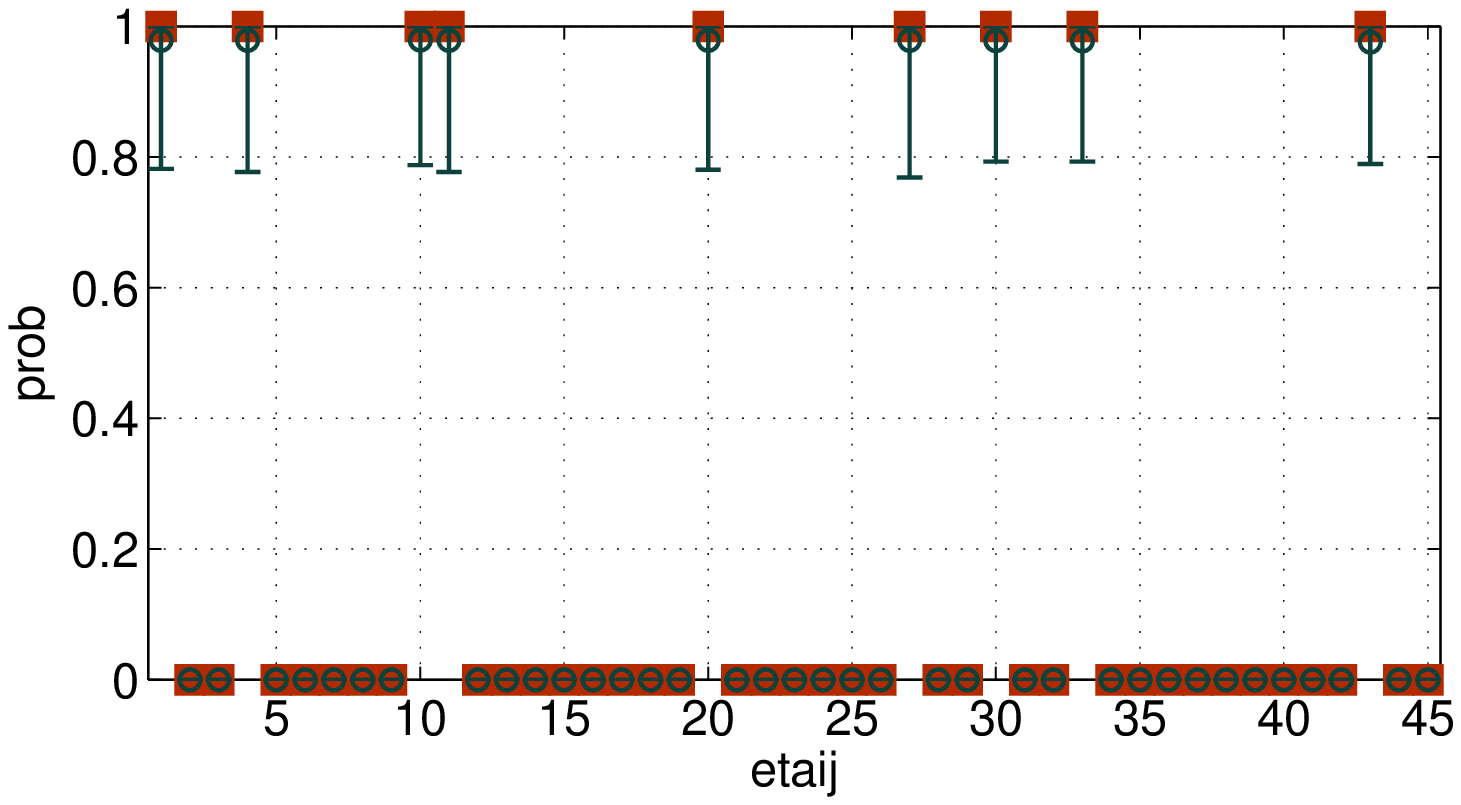}\label{fg:singleetabars}}
	\end{psfrags}
	\caption{Ground truth and estimated structures. (a) Ground truth mixing matrix. (b) Estimated mixing matrix using our sparse factor model. Note the sign ambiguity in some of the columns. (c) First 50 (out of 92) ordering candidates produced by our method during inference and their frequency, the first $m_\topc$ candidates were used for to learn DAGs. (d) Ground truth DAG. (e) Top candidate estimated using SLIM. (f) Estimated median weights for the DAG including $95\%$ credible intervals and ground truth (squares). (g) Summary of link probabilities measured as $\eta_{ij}=p(r_{ij}=1|\X,\cdot)$.} \label{fg:singlenets}
\end{figure}

\paragraph{Illustrative example.} Finally we want to show some of the most important elements of SLIM taking one successfully estimated example from the LiNGAM suite. Figure \ref{fg:singlenets} shows results for a particular DAG with $10$ variables obtained using $500$ observations, see Figures \ref{fg:singletrue_B} and \ref{fg:singlesFA_B} for the ground truth and the estimated DAG, respectively. True and estimated mixing matrices $\D$ for the equivalent factor model are also shown in Figures \ref{fg:singletrue_A} and \ref{fg:singlesFA_A}, respectively. In total our algorithm produced 92 orderings out of $3.6\times10^6$ possible, from which all $m_\topc=10$ candidates were correct. Figure \ref{fg:singlecand} shows the first 50 candidates and their frequency during sampling, the shaded area encloses the $m_\topc=10$ candidates. From Figure \ref{fg:singleebars} we see that the elements of $\B$ are correctly estimated and their credible intervals are small, mainly due to the lack of model mismatch. Figure \ref{fg:singleetabars} shows a good separation between zero and non-zero elements of $\B$ as summarized by $p(r_{ij}=1|\X,\cdot)$. It is worthwhile mentioning that using $\beta_m=0.99$ instead of $\beta_m=0.1$ in this example, still produces the right DAG, although the separation between zero and non-zero elements in Figure \ref{fg:singleetabars} will be smaller and with higher uncertainty, \ie larger credible intervals.
\subsection{Bayesian networks repository}
\begin{figure}[t]
	\begin{psfrags}
		\psfrag{tpr}[c][c][0.6]{True positive rate}\psfrag{fpr}[c][c][0.6]{False positive rate}\psfrag{rev}[c][c][0.6]{Reversed links rate}\psfrag{fnr}[c][c][0.6]{False negative rate}\psfrag{water (32)}[l][l][0.5]{water (32)}\psfrag{mildew (35)}[l][l][0.5]{mildew (35)}\psfrag{insurance (27)}[l][l][0.5]{insurance (27)}\psfrag{hailfinder (56)}[l][l][0.48]{hailfinder (56)}\psfrag{carpo (61)}[l][l][0.5]{carpo (61)}\psfrag{barley (48)}[c][c][0.5]{barley (48)}\psfrag{alarm (37)}[l][l][0.5]{alarm (37)}\psfrag{water}[l][l][0.5]{water}\psfrag{mildew}[l][l][0.5]{mildew}\psfrag{insurance}[l][l][0.5]{insurance}\psfrag{hailfinder}[l][l][0.48]{hailfinder}\psfrag{carpo}[l][l][0.5]{carpo}\psfrag{barley}[c][c][0.5]{barley}\psfrag{alarm}[l][l][0.5]{alarm}\psfrag{lingamxxxx}[l][l][0.45]{LiNGAM}\psfrag{slim}[l][l][0.45]{SLIM}
		\subfigure[]{\includegraphics[scale = 0.38]{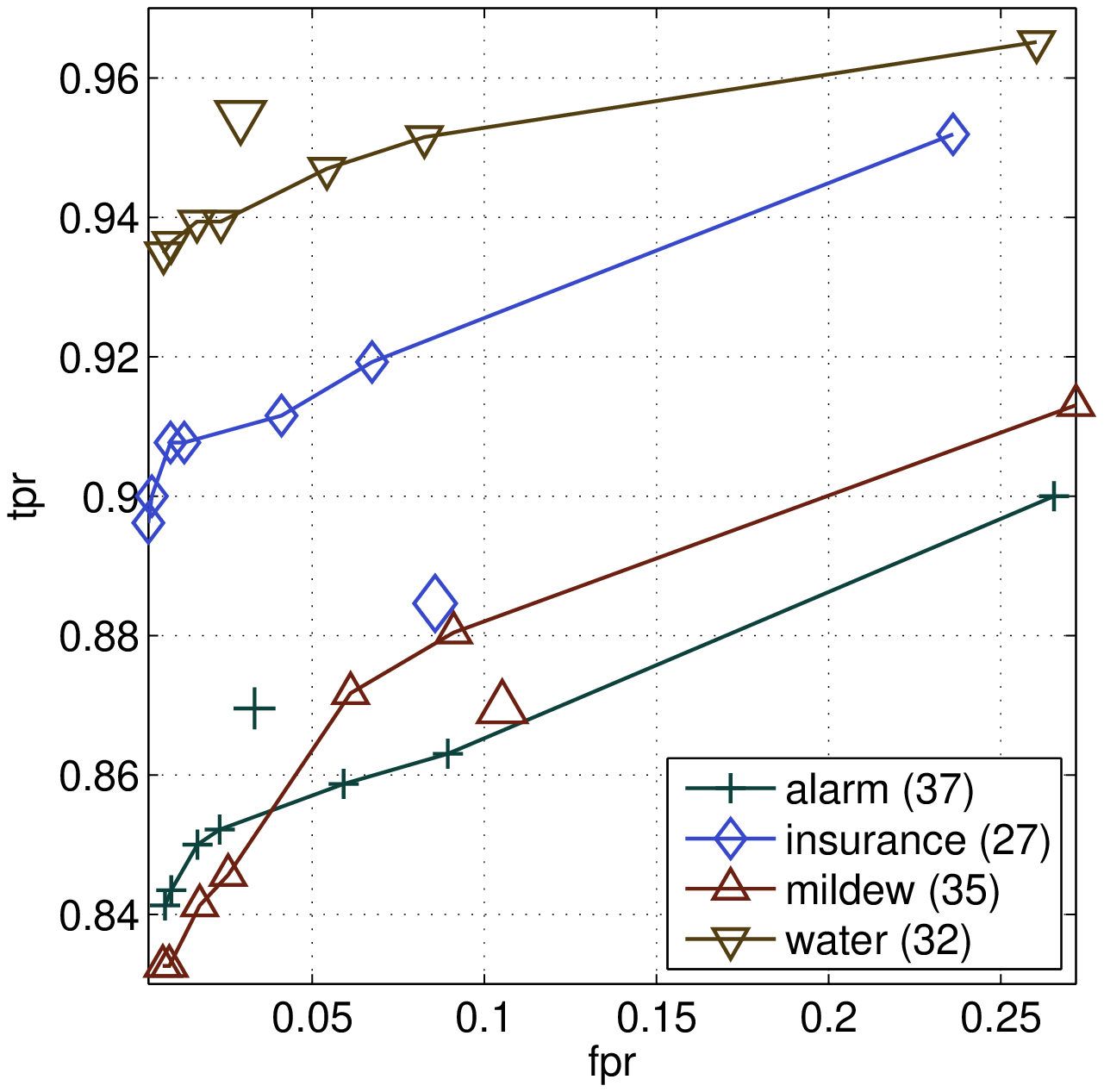}\label{fg:bnrepo_roc1}}
		\subfigure[]{\includegraphics[scale = 0.38]{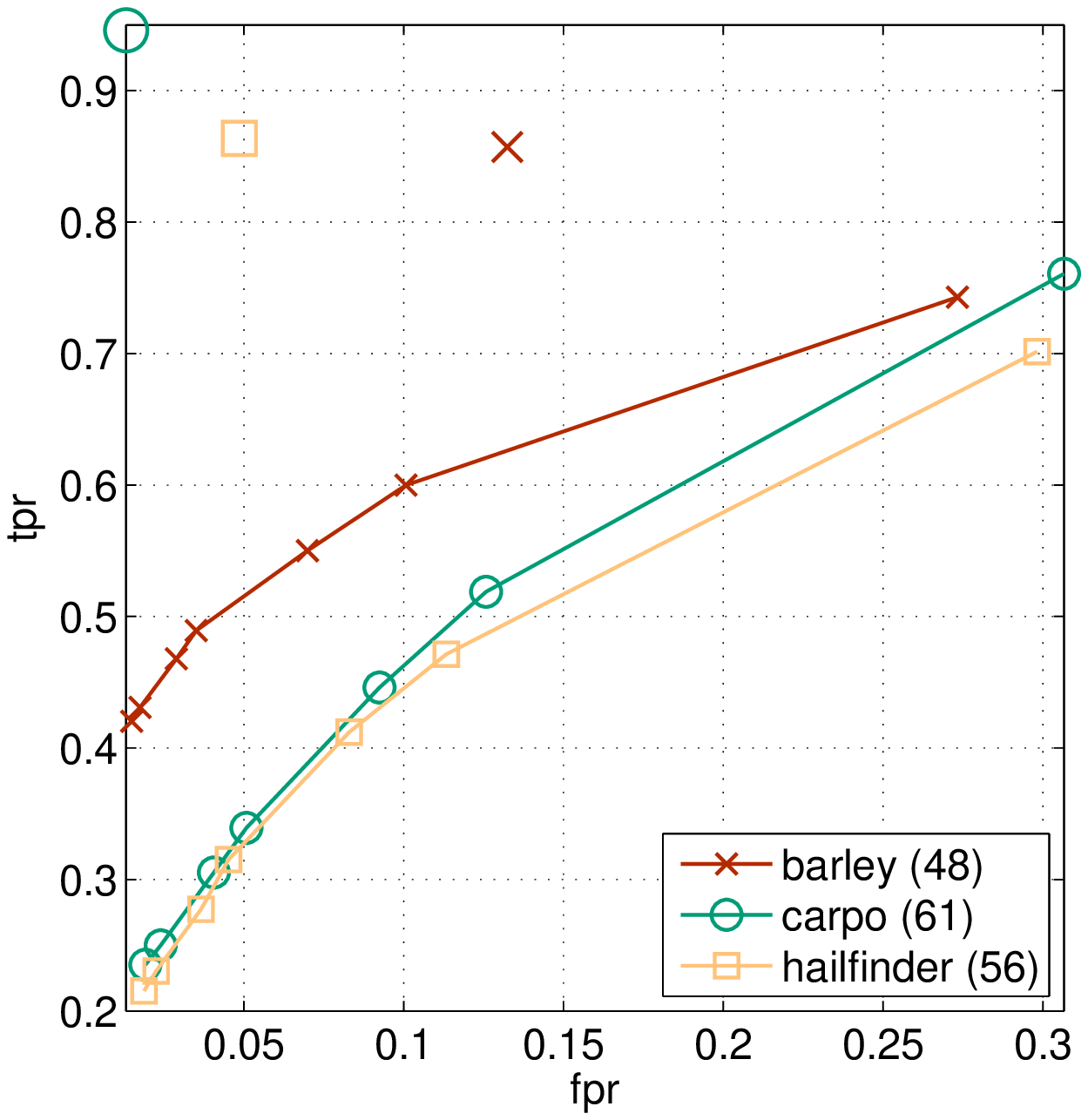}\label{fg:bnrepo_roc2}}
		\subfigure[]{\includegraphics[scale = 0.38]{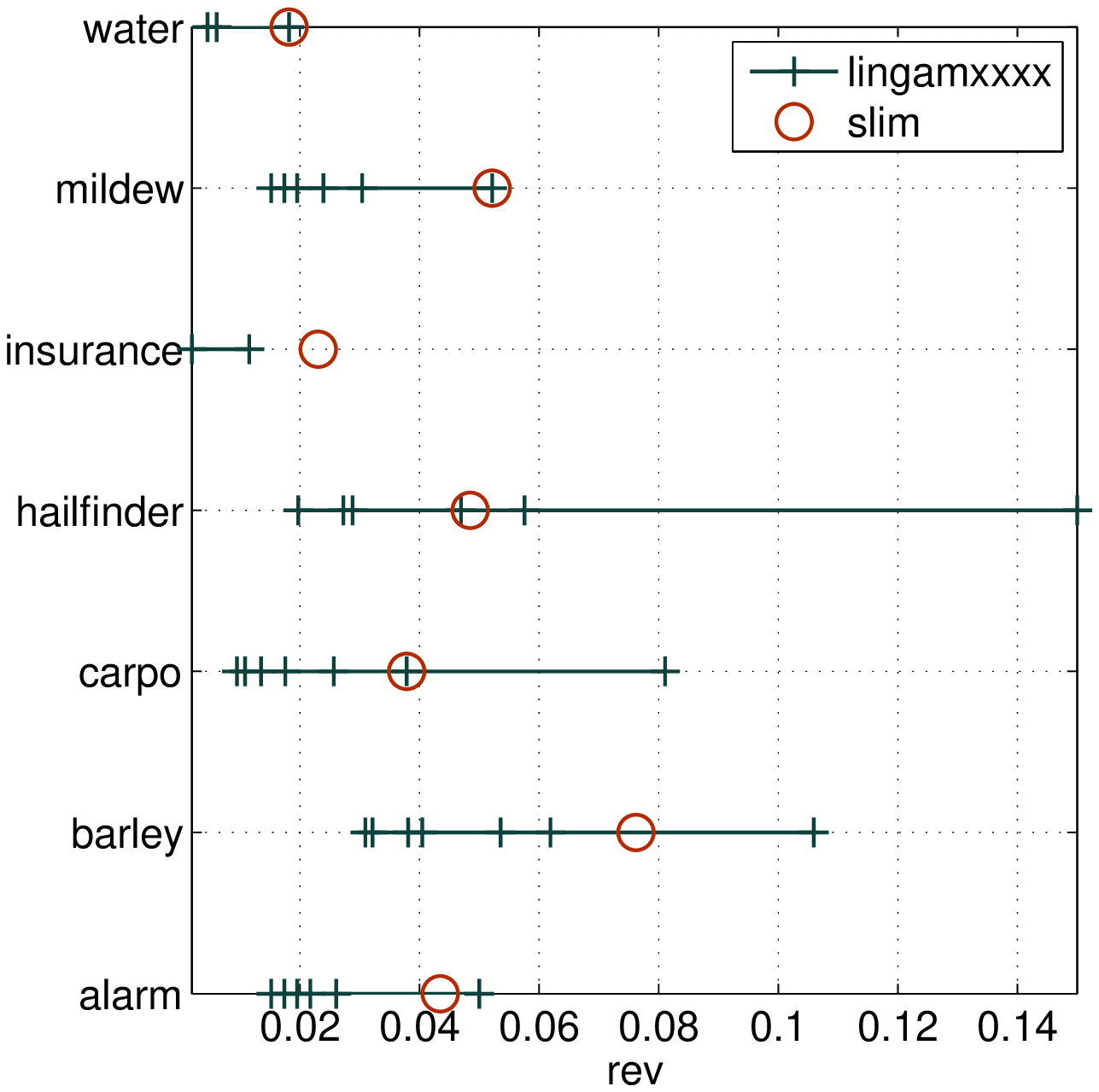}\label{fg:bnrepo_rev}}
	\end{psfrags}
	\caption{Performance measures for the Bayesian networks repository experiments. Each connected marker correspond to a different $p$-value in LiNGAM, starting left to right from 0.005. Disconnected markers denote SLIM results. Numbers in parentheses indicate number of variables.} 
\end{figure}

Next we want to compare our method against LiNGAM on some realistic structures. We consider 7 well known benchmark structures from the Bayesian network repository\footnote{Network structures available at \url{http://compbio.cs.huji.ac.il/Repository/}.}, namely alarm, barley, carpo, hailfinder, insurance, mildew and water ($d=$ 37, 48, 61, 56, 27, 35, 32 respectively). Since we do not have continuous data for any of the structures, we generated 10 datasets of size $N=500$ for each of them using heavy-tailed distributions with different parameters and equation \eqref{eq:PBxCz} with $m=0$, in a similar way as we did for the previous set of experiments, with $\R$ set to the ground truth and $\B$ from $\sign(\DN(0,1))+\DN(0,0.2)$. For LiNGAM, we only use Wald statistics because as seen in the previous experiment, it performs significantly better that bootstrapping. Again, we estimate models for different $p$-value cutoffs (0.0005, 0.001, 0.005, 0.01, 0.05, 0.1 and 0.5). For SLIM, we set $\beta_m=0.1$ since all the networks in the repository are sparse. Figures \ref{fg:bnrepo_roc1}, \ref{fg:bnrepo_roc2} and \ref{fg:bnrepo_rev} show averaged performance measures respectively as ROC curves and the proportion of links reversed in the estimated model due to ordering errors.

In this case, the results are mixed when looking at the performances obtained. Figure \ref{fg:bnrepo_roc2} shows that SLIM is better than LiNGAM in the larger datasets with a significant difference. Figure \ref{fg:bnrepo_roc1} shows for the remaining four datasets, that LiNGAM is better in two cases corresponding to the insurance and mildew networks. In general, both methods perform reasonably well given the size of the problems and the amount of data used to fit the models. However, SLIM tends to be more stable, when looking at the range of the true positive rates. It is important to note that the best and worst case for SLIM correspond to the largest and smallest network, respectively. We do not have a sensible explanation about why SLIM is performing that poorly on the insurance network. Figure \ref{fg:bnrepo_rev} implicitly reveals that both methods are unable to find the right ordering of the variables.

We also tried the following methods with encoded Gaussian assumptions: standard DAG search, order search, sparse candidate pruning then DAG search \citep{friedman99}, L1MB then DAG search \citep{schmidt07}, and sparse candidate pruning then order search \citep{teyssier05}. We observed (results not shown) that these methods produce similar results to those obtained by either LiNGAM or SLIM when only looking at the resulting undirected graph, \ie removing the directionality of the links. Evaluation of directionality in Gaussian models is out of the question because such methods can only find DAGs up to Markov equivalence classes, thus evaluation must be made using partially directed acyclic graphs (PDAGs). It is still possible to modify some of the methods mentioned above to handle non-Gaussian data by for instance using some other appropriate conditional independence tests, however this is out of the scope of this paper.
\subsection{Model comparison}
\begin{figure}[t]
	\centering
	\begin{psfrags}
		\psfrag{freq}[c][c][0.6]{Frequency}\psfrag{lr}[c][c][0.6]{Log-likelihood ratio}\psfrag{Factor models}[c][c][0.55]{Factor models}\psfrag{DAGs}[c][c][0.55]{DAGs}
		\subfigure[]{\includegraphics[scale = 0.49]{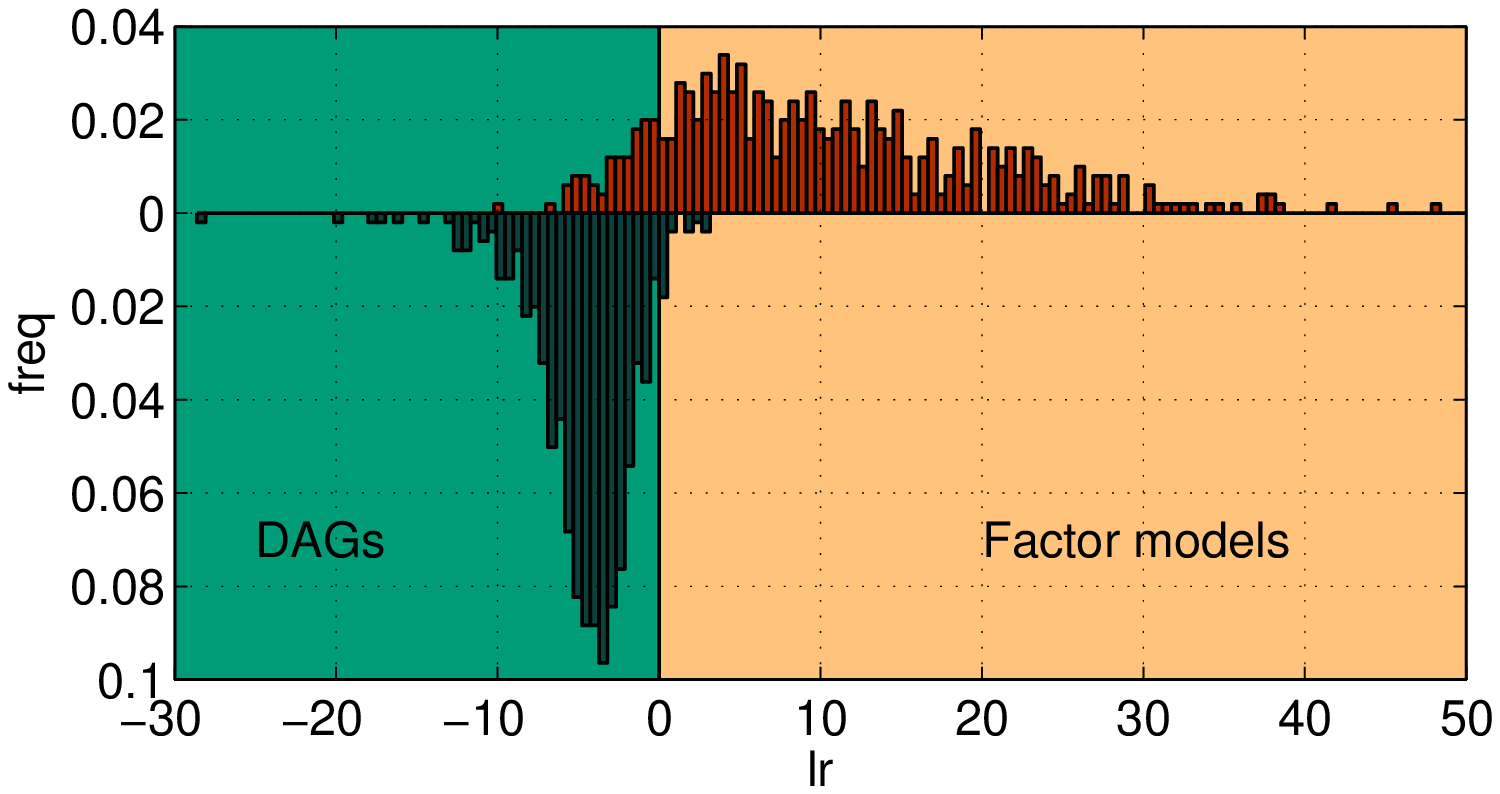}\label{fg:mcomp_500}}
		\subfigure[]{\includegraphics[scale = 0.49]{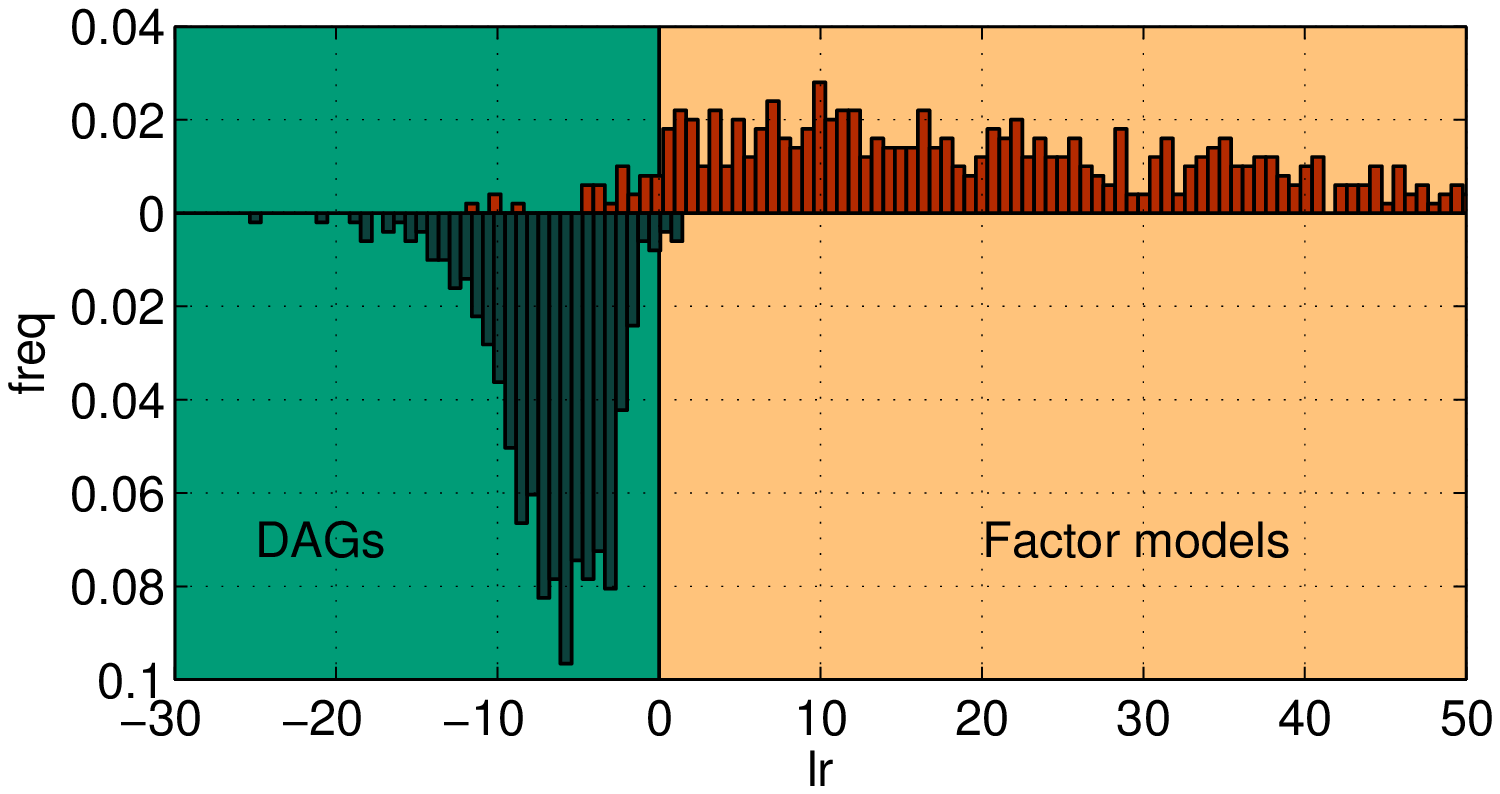}\label{fg:mcomp_1000}}
	\end{psfrags}
	\caption{Log-likelihood ratio empirical distributions for, (a) $N=500$ and (b) $N=1000$. Top bars correspond to true factor models, bottom bars to true DAGs and the ratio is computed as described in Section \ref{sc:ms}. Top bars lying below zero are true factor models predicted to be better explained by DAGs, thus model comparison errors.} \label{fg:mcomp}
\end{figure}

In this experiment we want to evaluate the model selection procedure described in Section \ref{sc:ms}. For this purpose we have generated 1000 different datasets/models with $d=5$ and $N=\{500,1000\}$ following the same procedure described in the first experiment, but this time we selected the true model to be either a factor model or a DAG with equal probability. In order to generate a factor model, we basically just need to ensure that $\D$ cannot be permuted to a triangular form, so the data generated from it does not admit a DAG representation. We kept $20\%$ of the data to compute the predictive densities to then select between all estimated DAG candidates and the factor model. We found that for $N=500$ our approach was able to select true DAGs $96.78\%$ of the times and true factor models $87.05\%$, corresponding to an overall accuracy of $91.9\%$. Increasing the number of observations, \ie for $N=1000$, the true DAG, true factor model rates and overall error increased to $98.99\%$, $95.0\%$ and $96.99\%$, respectively. Figure \ref{fg:mcomp} shows separately the empirical log-likelihood ratio distributions obtained from the 1000 datasets for DAGs and factor models. The shaded areas correspond to the true DAG/factor model regions, with zero as their boundary. Note that when the wrong model is selected the likelihood ratio is nicely close to the boundary and the overlap of the two distributions decreases with the number of observations used, since the quality of the predictive density increases accordingly. The true DAG rates tend to be larger than for factor models because it is more likely that the latter is confused with a DAG due to estimation errors or closeness to a DAG representation, than a DAG being confused with a factor model which is naturally more general. This is precisely why the likelihood ratios tend to be larger on the factor model side of he plots. All in all, these results demonstrate that our approach is very effective at selecting the true underlying structure when the data is generated by one of the two hypotheses.
\subsection{DAGs with latent variables}
\begin{figure}[tb]
	\centering
	\begin{psfrags}
		 \psfrag{chain}[c][c][0.6]{Chain}\psfrag{var}[c][c][0.6]{Variable}\psfrag{b11}[c][c][0.6]{$b_{11}$}\psfrag{b12}[c][c][0.6]{$b_{12}$}\psfrag{b21}[c][c][0.6]{$b_{21}$}\psfrag{b22}[c][c][0.6]{$b_{22}$}\psfrag{c1L}[c][c][0.6]{$c_{1L}$}\psfrag{c2L}[c][c][0.6]{$c_{2L}$}\psfrag{d11}[c][c][0.6]{$d_{11}$}\psfrag{d12}[c][c][0.6]{$d_{12}$}\psfrag{d21}[c][c][0.6]{$d_{21}$}\psfrag{d22}[c][c][0.6]{$d_{22}$}\psfrag{d31}[c][c][0.6]{$d_{31}$}\psfrag{d32}[c][c][0.6]{$d_{32}$}\psfrag{lik}[c][c][0.6]{Test likelihood}\psfrag{ 0}[c][c][0.45]{T}
		\subfigure[]{\includegraphics[scale = 0.4]{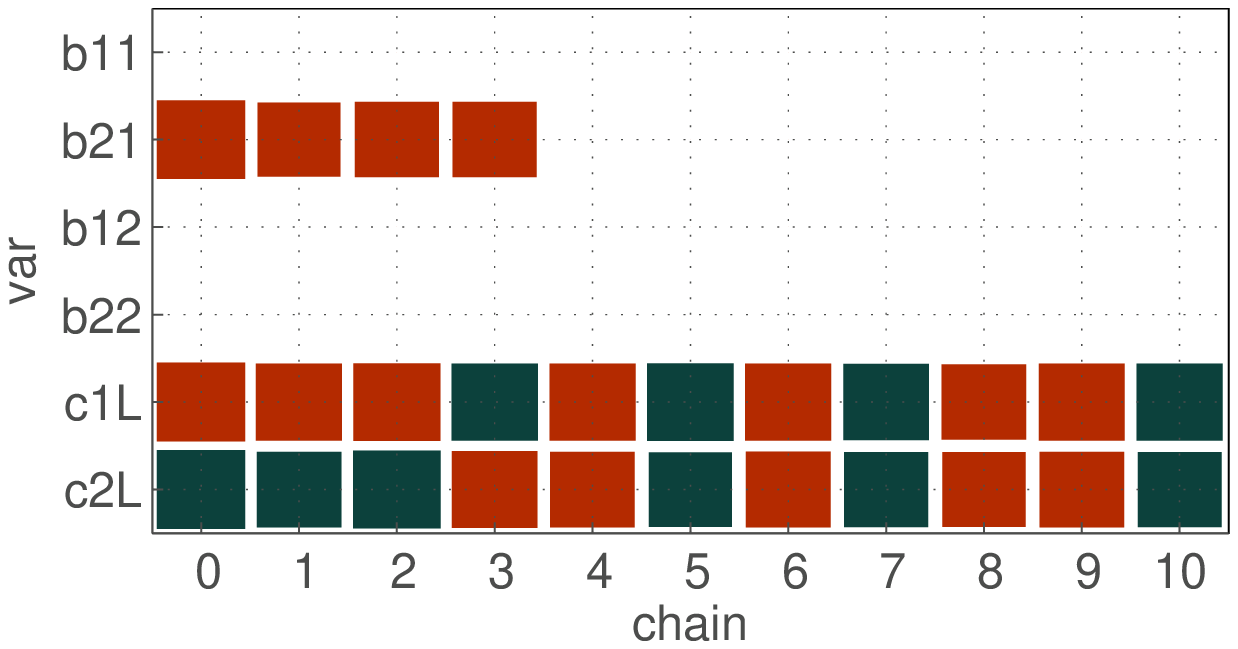}\label{fg:mixDAG_B_ica1}}\hspace{3mm}
		\subfigure[]{\includegraphics[scale = 0.4]{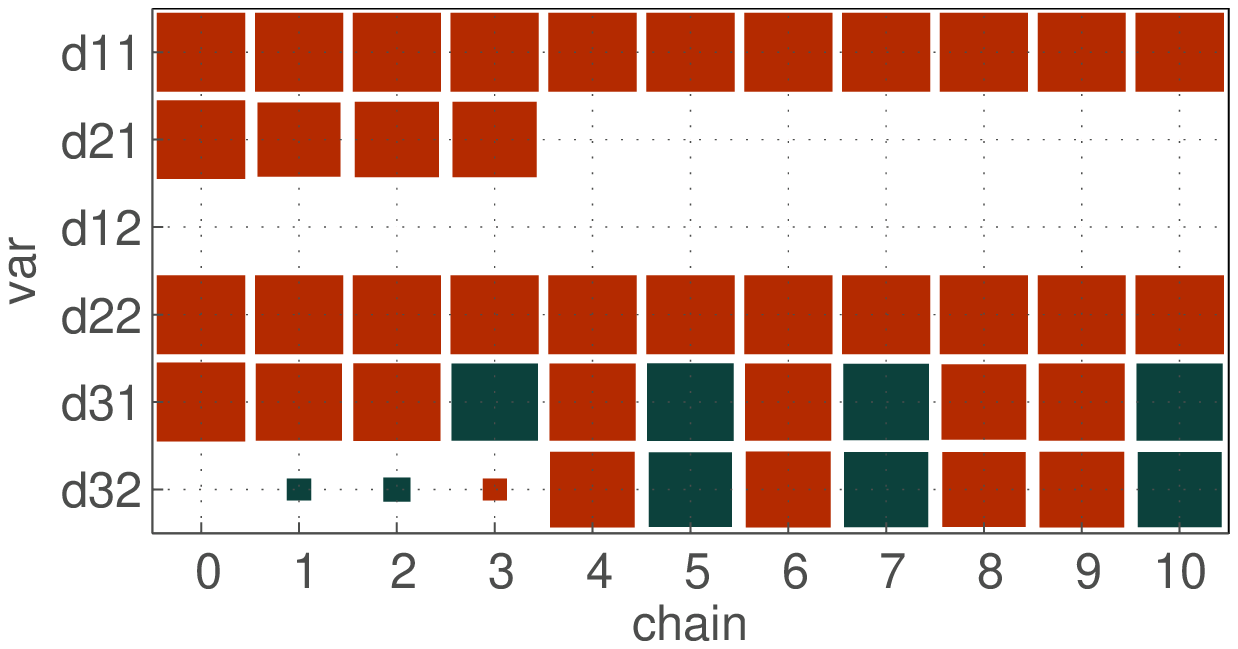}\label{fg:mixDAG_D_ica1}}\hspace{3mm}
		\subfigure[]{\includegraphics[scale = 0.4]{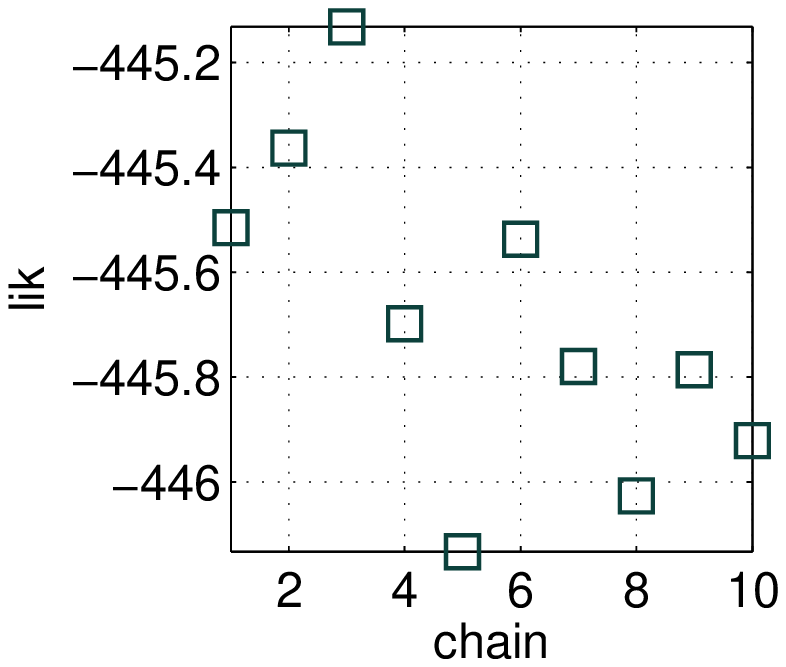}\label{fg:mixDAG_lik_ica1}}
		\subfigure[]{\includegraphics[scale = 0.4]{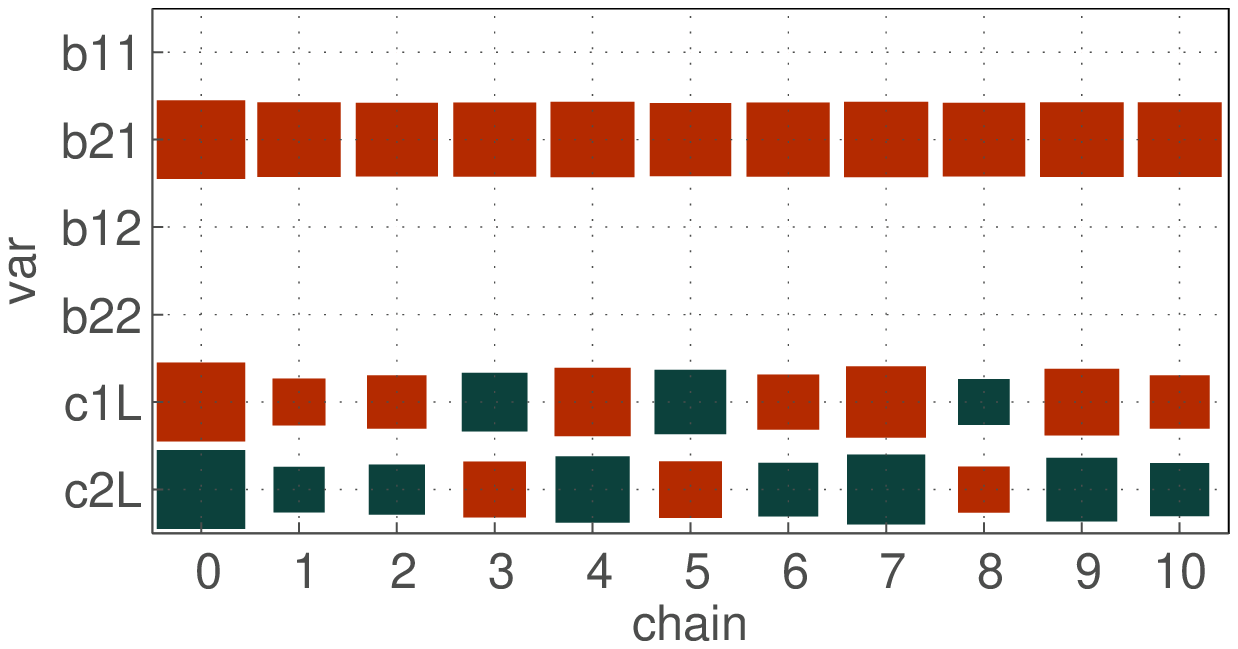}\label{fg:mixDAG_B_ica2}}\hspace{3mm}
		\subfigure[]{\includegraphics[scale = 0.4]{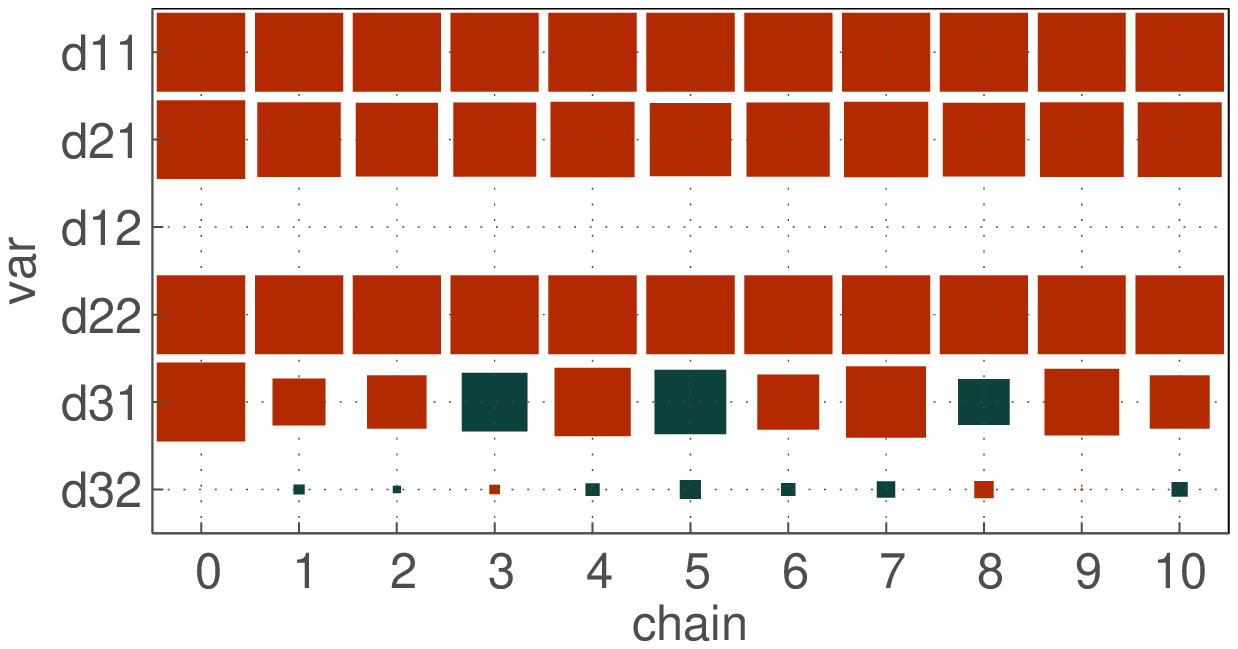}\label{fg:mixDAG_D_ica2}}\hspace{3mm}
		\subfigure[]{\includegraphics[scale = 0.4]{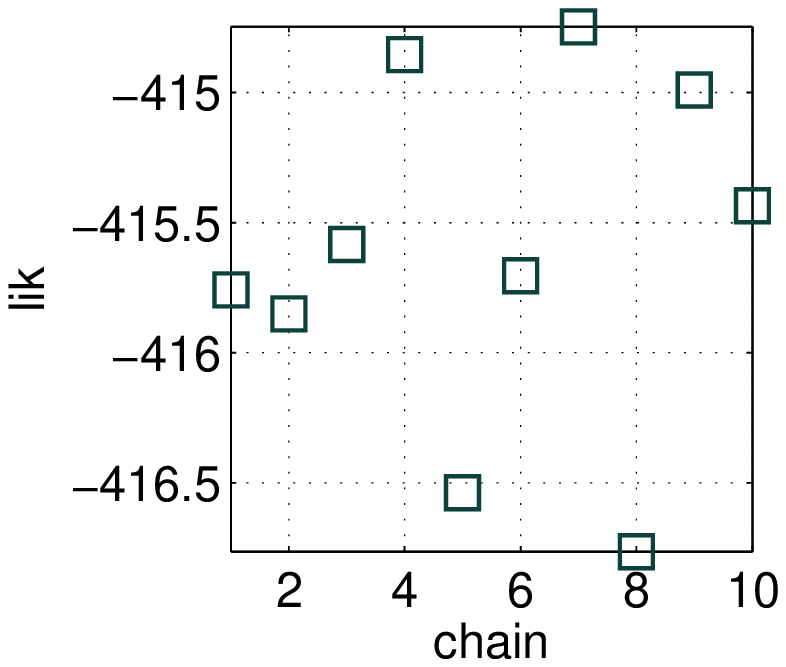}\label{fg:mixDAG_lik_ica2}}
	\end{psfrags}
	\caption{Identifiability experiment for the DAG with latent variables. Connectivities $\B$ and $\C_L$ are shown for (u) in (a) and (i) in (d). Equivalent mixing matrix $\D$ for (u) in (b) and for (i) in (d). Test likelihoods for (u) and (i) are shown in (c) and (f) respectively. The first column in (a,b,d,e) denoted as T is the ground truth. Dark and light boxes are negative and positive numbers, accordingly.} \label{fg:mixDAGtoy}
\end{figure}

We will start by illustrating the identifiability issues of the model in equation \eqref{eq:PBxCz} discussed in Section \ref{sc:idf} with a very simple example. We generated $N=500$ observations from the graph in Figure \ref{fg:toyDAGL2} and kept $20\%$ of the data to compute test likelihoods. Now, we perform inference on two slightly different models, namely, (u) where $\z'=[z'_1 \ z'_2 \ z'_L]$ is provided with Laplace distributions with unit variance, \ie $\lambda=2$, and (i) where $z_1,z_2$ have Laplace distributions with unit variance and $z_L$ is Cauchy distributed. We want to show that even if both models match the true generating process, (u) is non-identifiable whereas (i) can be successfully estimated. In order to keep the experiment controlled as much as possible, we set $\beta_m=0.99$ to reflect that the ground truth is dense and we did not infer $\C_D$ and set it to the true values, \ie the identity. Then, we ran 10 independent chains for each one of the models and summarized $\B$, $\C_L$, $\D$ and the test likelihoods in Figure \ref{fg:mixDAGtoy}.

Figure \ref{fg:mixDAG_B_ica1} shows that model (u) finds the DAG in Figure \ref{fg:toyDAGL2} (the ground truth) in 3 cases, and in the remaining 7 cases it finds the DAG in Figure \ref{fg:toyDAGL1}. Note also that the test likelihoods in Figure \ref{fg:mixDAG_lik_ica1} are almost identical, as must be expected due to the lack of identifiability of the model, so they cannot be used to select among the two alternatives. Model (i) finds the right structure all the times as shown in Figure \ref{fg:mixDAG_B_ica2}. The mixing matrix of the equivalent factor model, $\D$ is shown in Figures \ref{fg:mixDAG_D_ica1} and \ref{fg:mixDAG_D_ica2} for (u) and (i), respectively. In Figure \ref{fg:mixDAG_D_ica1}, the first and third column of $\D$ exchange positions because all the components of $\z$ have the same distribution, which is not the case of Figure \ref{fg:mixDAG_D_ica2}. The small quantities in $\D$ are due to estimation errors when computing $b_{21}c_{1L}+c_{2L}$, and this cancels out in the true model. The sign changes in Figures \ref{fg:mixDAG_B_ica1} and \ref{fg:mixDAG_B_ica2} are caused by the sign ambiguity of $\z_L$ in the product $\C_L\z_L$. We also tested the alternative model in Figure \ref{fg:toyDAGL2} obtaining equivalent results, \ie 4 successes for model (u) and 10 for model (i). This small example shows how non-identifiability may lead to two very different DAG solutions with distinct interpretations of the data. 

\citet{hoyer08a} recently presented an approach to DAGs with latent variables based on LiNGAM \citep{shimizu06}. Their procedure uses probabilistic ICA and bootstrapping to infer the equivalent factor model distribution $p(\D|\X)$, then greedily selects $m$ columns of $\D$ to be latent variables until the remaining ones can be permuted to triangular and the resulting DAG is compatible with the faithfulness assumption \citep[see,][]{pearl00}. If we assume that their procedure is able to find the exact $\D$ for the graphs in Figures \ref{fg:toyDAGL1} and \ref{fg:toyDAGL2}, due to the faithfulness assumption, the DAG in Figure \ref{fg:toyDAGL1} will be always selected regardless of the ground truth\footnote{See \citet{robins03} for a very interesting explanation of faithfulness using the same example presented here.}. In practice, the solution obtained for $\D$ is dense and needs to be pruned, hence we rely on $p(\X,\D)$ being larger for the ground truth in Figure \ref{fg:toyDAGL2} than for the graph in Figure \ref{fg:toyDAGL1}, however since both models differ only by a permutation of the columns of $\D$, they have exactly the same joint density $p(\X,\D)$ --- they are non-identifiable, thus the algorithm will select one of the options by chance. Since the source of non-identifiability of their algorithm is permutations of columns of $\D$, it does not matter if probabilistic ICA match or not the distribution of the underlying process as in our model. Anyway, we decided to try models (u) and (i) described above using the algorithm just described \footnote{Matlab package (v.1.1) freely available at \url{http://www.cs.helsinki.fi/group/neuroinf/lingam/}.}. Regardless of the ground truth, Figures \ref{fg:toyDAGL1} or \ref{fg:toyDAGL2}, the algorithm always selected the DAG in Figure \ref{fg:toyDAGL2}, which in this particular case is due to $p(\X,\D)$ being slightly larger for the denser model.

Now we test the model in a more general setting. We generate 100 models and datasets of size $N=500$ using a similar procedure to the one in the artificial data experiment. The models have $d=5$ and $m=1$, no dense structures are generated and the distributions for $\z$ are heavy-tailed, drawn from a generalized Gaussian distribution with random shape. For SLIM, we use the following settings, $\beta_m=0.1$, $\z_D$ is Laplace with unit variances and $\z_L$ is Cauchy. Furthermore, we have doubled the number of iterations of the DAG sampler, \ie 6000 samples and a burn-in period of 2000, so as to compensate for the additional parameters that need to be inferred due to inclusion of latent variables. Our ordering search procedure was able to find the right ordering 78 out of 100 times. The true positive rates, true negative rates and median AUC are 88.28$\%$, 96.40$\%$ and 0.929, respectively, corresponding to approximately 1.5 structure errors per network. Using \citet{hoyer08a} we obtained 1 true ordering out of 100, 91.63$\%$ true positive rate, 65.18$\%$ true negative rate and 0.800 median AUC, showing again the preference of the algorithm for denser models. We regard these results as very satisfactory for both methods considering the difficulty of the task and the lack of identifiability of the model by \citet{hoyer08a}.
\subsection{Non-linear DAGs}
\begin{figure}[tb]
	\subfigure[]{\begin{tikzpicture}[ bend angle = 45, >=latex, font = \footnotesize ]
		\tikzstyle{obs} = [ circle, thick, draw = black!80, fill = imp2, minimum size = 1mm, inner sep = 2pt ]
		\tikzstyle{lat} = [ circle, thick, draw = black!80, fill = black!0, minimum size = 1mm, inner sep = 2pt ]
		\tikzstyle{cellf} = [ rectangle, draw = black!100, fill = black!0, minimum width = 4mm, minimum height = 4mm, inner sep = 2pt ]
		\tikzstyle{celle} = [ rectangle, draw = black!100, fill = black!10, minimum width = 4mm, minimum height = 4mm, inner sep = 2pt ]
		\tikzstyle{lab} = [ rectangle, draw = black!0, fill = black!0, minimum width = 4mm, minimum height = 4mm, inner sep = 2pt ]
		\tikzstyle{dmy} = [ circle, thick, draw = black!0, fill = black!0, minimum size = 1mm, inner sep = 1pt]
		\tikzstyle{every label} = [black!100]
		\begin{scope}[ node distance = 1.8cm and 1.8cm, rounded corners = 4pt ]
			\node [obs] (x1)  [] {$x_1$};
			\node [obs] (x2) [ below left of = x1 ] {$x_2$}
				edge [pre] node[ above left ] {$x_1^2$} (x1);
			\node [obs] (x3) [ below right of = x1 ] {$x_3$}
				edge [pre] node[ above right ] {$4\sqrt{|x_1|}$} (x1);
			\node [obs] (x4) [ below right of = x2] {$x_4$}
				edge [pre] node[ below left ] {$2\sin(x_2)$} (x2)
				edge [pre] node[ below right ] {$2\sin(x_3)$} (x3);
			\node [dmy] (x5) [ below of = x4, node distance = 0.68cm ] {}; 
		\end{scope}
	\end{tikzpicture}\label{fg:SNIMtoy_gt}}
	\begin{psfrags}
		 \psfrag{ord}[c][c][0.6]{Orderings}\psfrag{err}[c][c][0.6]{Error}\psfrag{lik}[c][c][0.6]{Likelihood}\psfrag{telik}[c][c][0.6]{Test likelihood}
		\subfigure[]{\includegraphics[scale = 0.44]{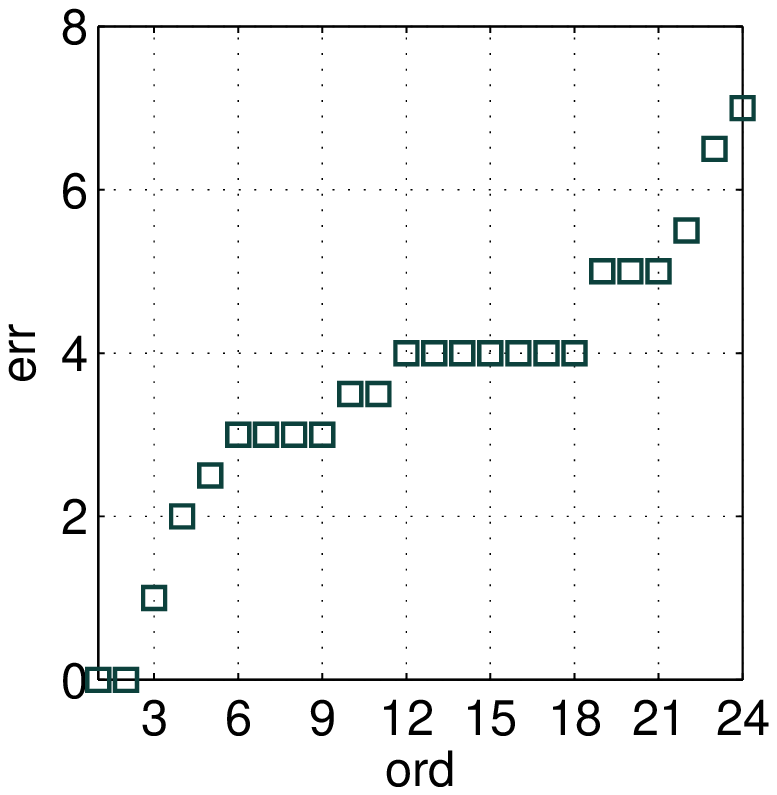}\label{fg:SNIMtoy_err}}
		\subfigure[]{\includegraphics[scale = 0.44]{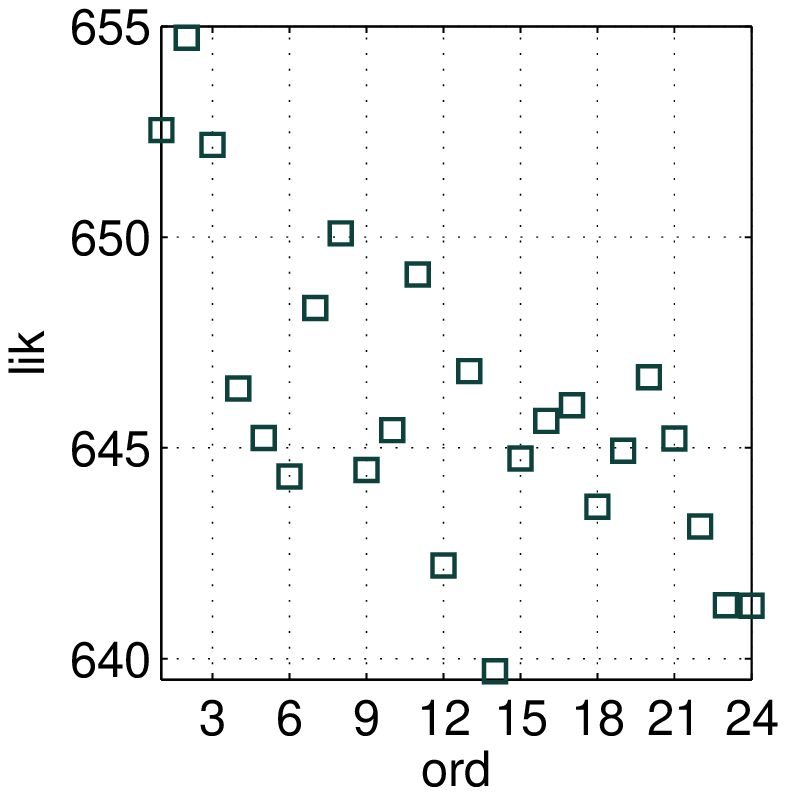}\label{fg:SNIMtoy_lik}}
		\subfigure[]{\includegraphics[scale = 0.44]{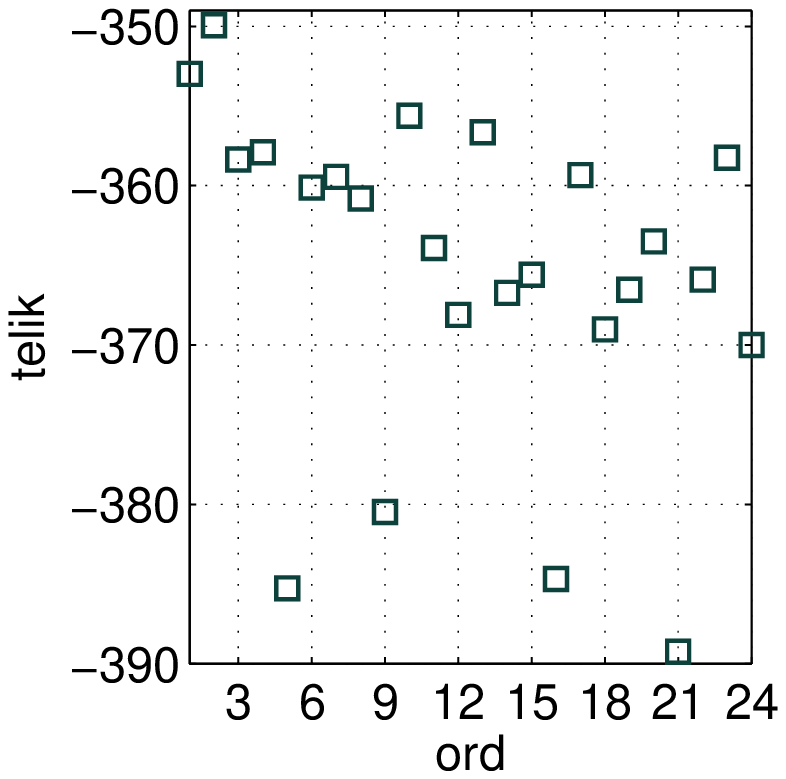}\label{fg:SNIMtoy_telik}}
	\end{psfrags}
	\caption{Non-linear DAG artificial example. (a) Network with non-linear interactions between observed nodes used as ground truth. (b,c,d) Median error, likelihood and test likelihood for all possible orderings and 10 independent repetitions. The plots are sorted according to number of errors and only the first two are valid according to the ground truth in (a), \ie $(1,2,3,4)$ and $(1,3,2,4)$. Note that when the error is zero in (b) the likelihoods are larger with respect to the remaining orderings in (c) and (d).} 
\end{figure}

For Sparse Non-linear Identifiable Modeling (SNIM) described in Section \ref{sc:snim}, first we want to show that our method can find and select from DAGs with non-linear interactions. We used the artificial network from \citet{hoyer08} shown here in Figure \ref{fg:SNIMtoy_gt} and generated 10 different datasets corresponding to $N=100$ observations, each time using driving signals sampled from different heavy-tailed distributions. Since we do not yet have an ordering search procedure for non-linear DAGs, we perform DAG inference for all possible orderings and datasets. The results obtained are evaluated in two ways, first we check if we can find the true connectivity matrix when the ordering is correct. Second, we need to validate that the likelihood is able to select the model with less error and correct ordering among all possible candidates so we can use it in practice. Figures \ref{fg:SNIMtoy_err}, \ref{fg:SNIMtoy_lik} and \ref{fg:SNIMtoy_telik} show the median errors, training and test likelihoods (using 20\% of the data) for each one of the orderings, respectively. In this particular case we only have two correct orderings, namely, $(1,2,3,4)$ and $(1,3,2,4)$, corresponding to the first and second candidates in the plots. Figure \ref{fg:SNIMtoy_err} shows that the error is zero only for the two correct orderings, then our model is able to infer the structure once the right ordering is given as desired. As a result of the identifiability, data and test likelihoods shown in Figures \ref{fg:SNIMtoy_lik} and \ref{fg:SNIMtoy_telik} correlate nicely with the structural error in Figure \ref{fg:SNIMtoy_err}. This means that we can use use the likelihoods as a proxy for the structural error just as in the linear case.

We also tested the network in Figure \ref{fg:SNIMtoy_gt} using three non-linear structure learning procedures namely greedy standard hill-climbing DAG search, the \quotes{ideal parent} algorithm \citep{elidan07} and kernel PC \citep{tillman09}. The first two methods use a scaled sigmoid function to capture the non-linearities in the data. In particular, they assume that a variable $x$ can be explained as scaled sigmoid transformation of a linear combination of its parents. The best median result we could obtain after tuning the parameters of the algorithms was 2 errors and 2 reversed links\footnote{Maximum number of iterations, random restarts to avoid local minima, regularization of the non-linear regression and the number of ranking candidates in ideal parent algorithm.}. Both methods perform similarly in this particular example, the only significant difference being their computational cost, which is considerably smaller for the \quotes{ideal parent} algorithm, as it was also pointed out by \citet{elidan07}. The reason why we consider these algorithms do not perform well here is that the sigmoid function can be very limited at capturing certain non-linearities due to its parametric form whereas the nonparametric GP gives flexible non-linear functions. The third method uses non-linear independence tests together with non-linear regression (relevance vector machines) and the PC algorithm to produce mixed DAGs. The best median result we could get in this case was 2 errors, 0 reversed links and 1 bidirectional links. These three non-linear DAG search algorithms have the great advantage of not requiring exhaustive enumeration of the orderings as our method and others available in the literature. \citet{zhang09} provides theoretical evidence of the possibility for flexible non-linear modeling without exhaustive order search but not a way to do it in practice. Yet another possibility not tried here will be to take the best parts of both strategies by taking the outcome of the non-linear DAG search algorithm and refine it using a nonparametric method like SNIM. However, it is not entirely clear how the non-linearities can affect the ordering of the variables. In the remaining part of this section we only focus on tasks for pairs of variables where the ordering search is not an issue.

\begin{figure}[tb]
	\centering
	\begin{psfrags}
		\psfrag{hyp}[c][c][0.6]{Hypothesis}\psfrag{tlik}[c][c][0.6]{Test likelihood}\psfrag{d-i}[c][c][0.3]{Duration $\rightarrow$ Interval}\psfrag{i-d}[c][c][0.3]{Interval $\rightarrow$ Duration}\psfrag{duration---interval}[c][c][0.4]{Duration $\rightarrow$ Interval}\psfrag{interval---duration}[c][c][0.4]{Interval $\rightarrow$ Duration}\psfrag{den}[c][c][0.6]{Density}\psfrag{lik}[c][c][0.6]{Likelihood}
		\subfigure[]{\includegraphics[scale = 0.35]{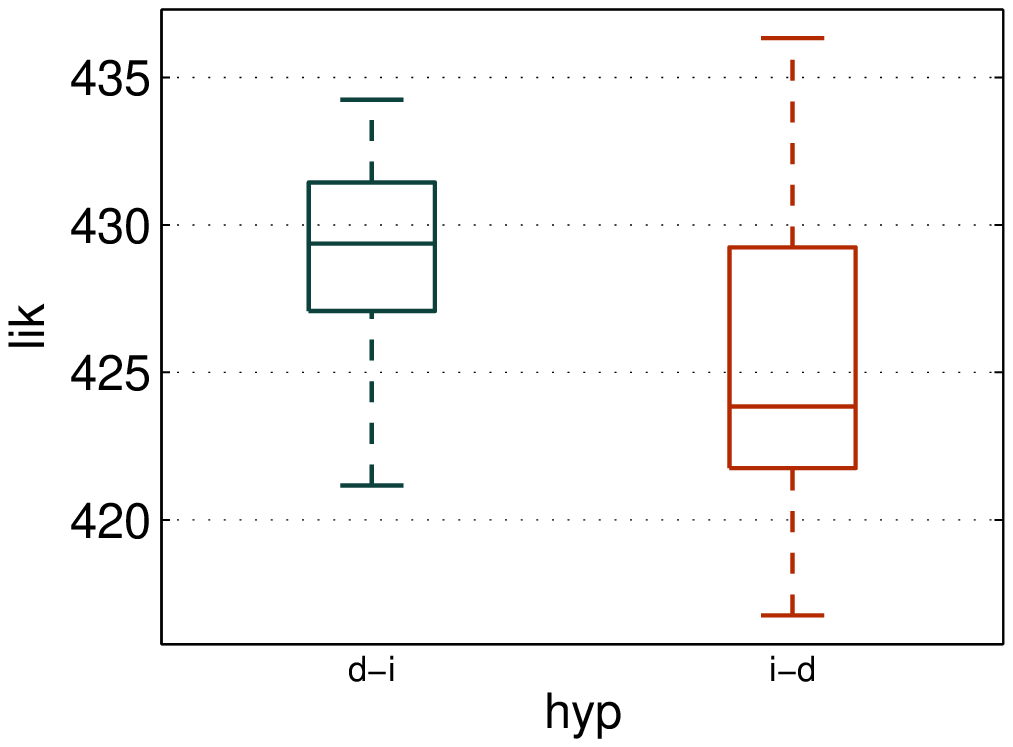}\label{fg:faithful_blik}}
		\subfigure[]{\includegraphics[scale = 0.35]{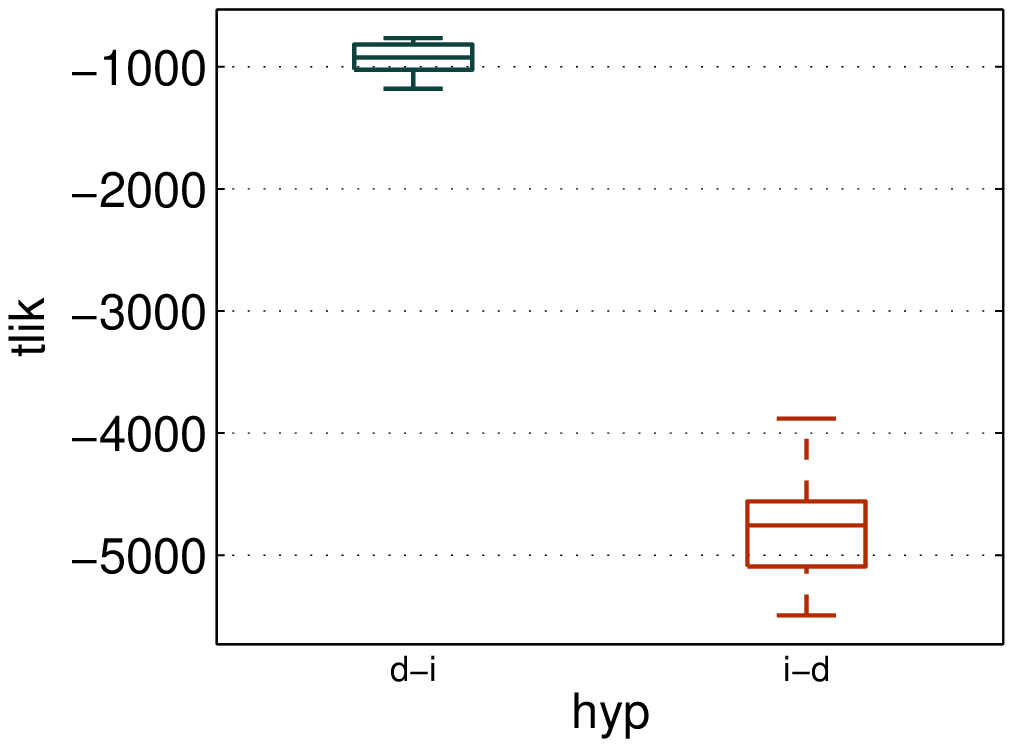}\label{fg:faithful_btlik}}
		\subfigure[]{\includegraphics[scale = 0.35]{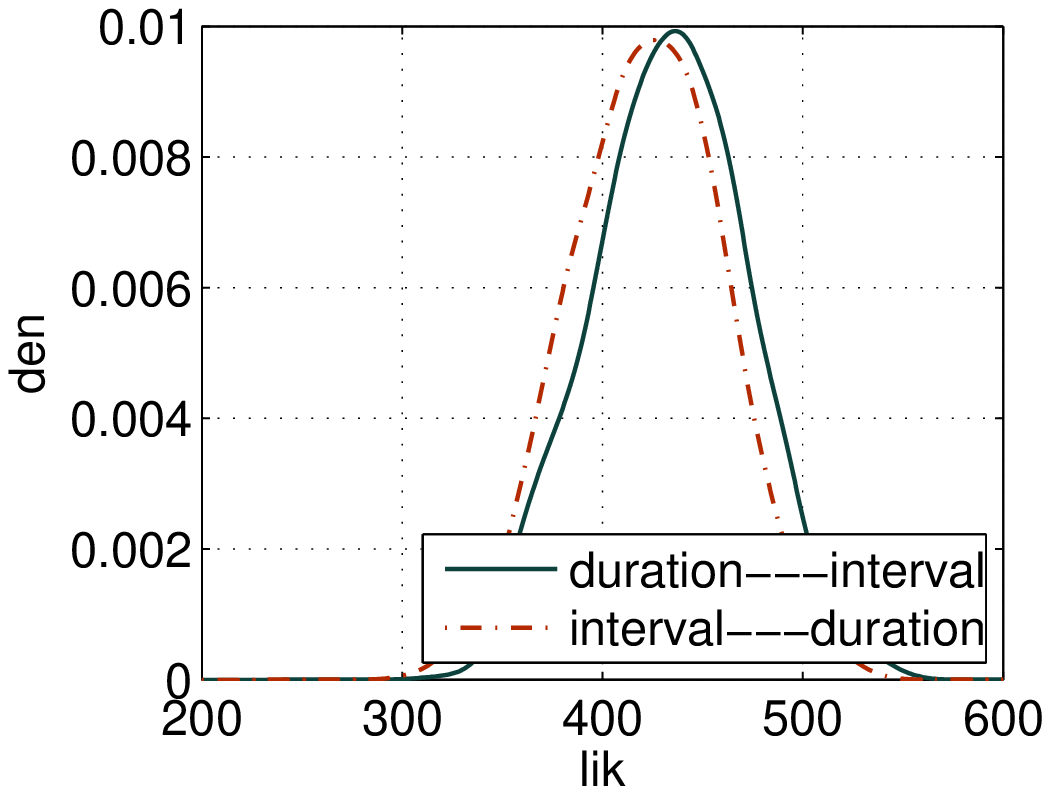}\label{fg:faithful_dlik}}
		\subfigure[]{\includegraphics[scale = 0.35]{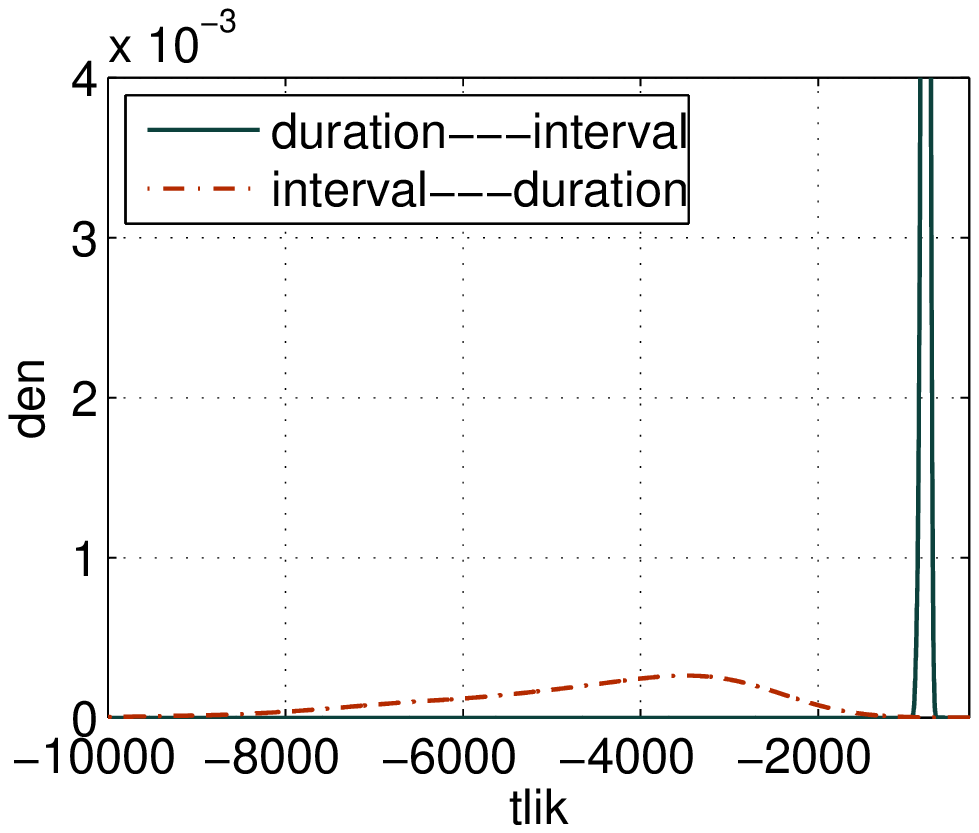}\label{fg:faithful_dtlik}}
	\end{psfrags}
	\caption{Testing $\{$duration, interval$\}$ in Old Faithful dataset. (a,b) Data and test likelihood boxplots for 10 independent repetitions. (c,d) Training and test likelihood densities for one of the repetitions. The test likelihood separates consistently the two tested hypotheses.} 
\end{figure}

The dataset known as Old Faithful \citep{asuncion07} contains 272 observations of two variables measuring waiting time between eruptions and duration of eruptions for the Old Faithful geyser in Yellowstone National Park, USA. We want to test the two possible orderings, duration $\rightarrow$ interval and interval $\rightarrow$ duration. Figures \ref{fg:faithful_blik} and \ref{fg:faithful_btlik} show training and test likelihood boxplots for 10 independent randomizations of the dataset with $20\%$ of the observations used to compute test likelihoods. Our model was able to find the right ordering, \ie duration $\rightarrow$ interval in all cases when the test likelihood was used but only 7 times with the training likelihood due to the proximity of the densities, see Figure \ref{fg:faithful_dlik}. On the other hand, the predictive density is very discriminative, as shown for instance in Figure \ref{fg:faithful_dtlik}. This is not a very surprising result since making the duration a function of the interval results in a very non-linear function, whereas the alternative function is almost linear (data not shown).

\begin{figure}[tb]
	\centering
	\begin{psfrags}
		\psfrag{hyp}[c][c][0.6]{Hypothesis}\psfrag{tlik}[c][c][0.6]{Test likelihood}\psfrag{l-a}[c][c][0.3]{Length $\rightarrow$ Age}\psfrag{a-l}[c][c][0.3]{Age $\rightarrow$ Length}\psfrag{length---age}[c][c][0.4]{Length $\rightarrow$ Age}\psfrag{age---length}[c][c][0.4]{Age$ \rightarrow$ Length}\psfrag{den}[c][c][0.6]{Density}\psfrag{lik}[c][c][0.6]{Likelihood}
		\subfigure[]{\includegraphics[scale = 0.35]{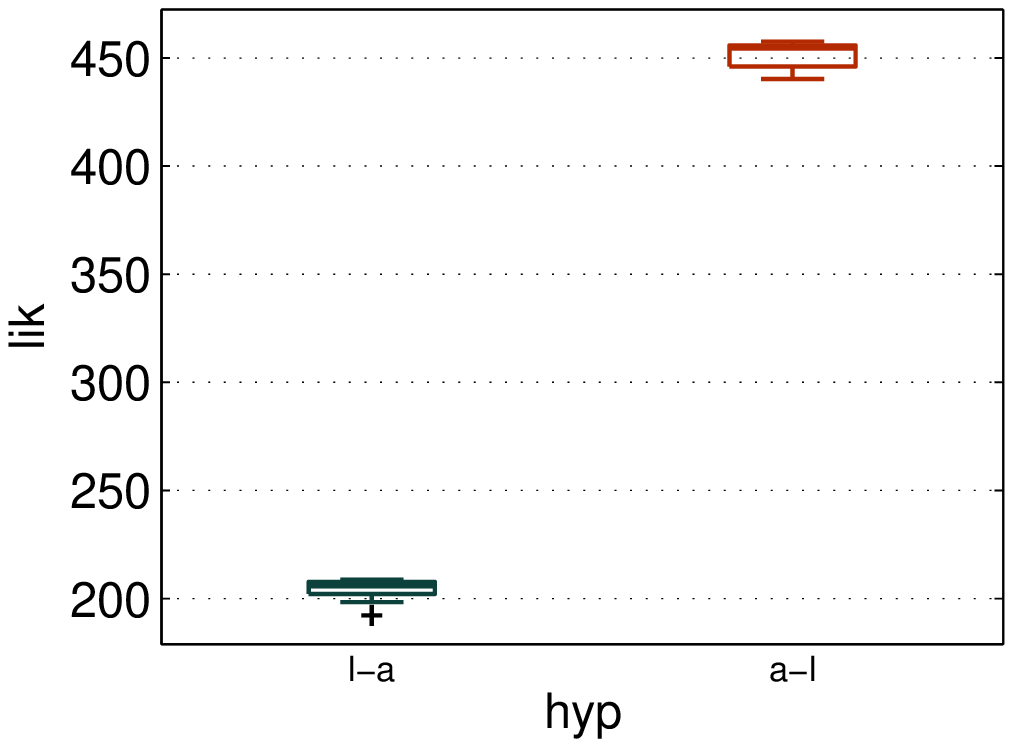}\label{fg:abalone_blik}}
		\subfigure[]{\includegraphics[scale = 0.35]{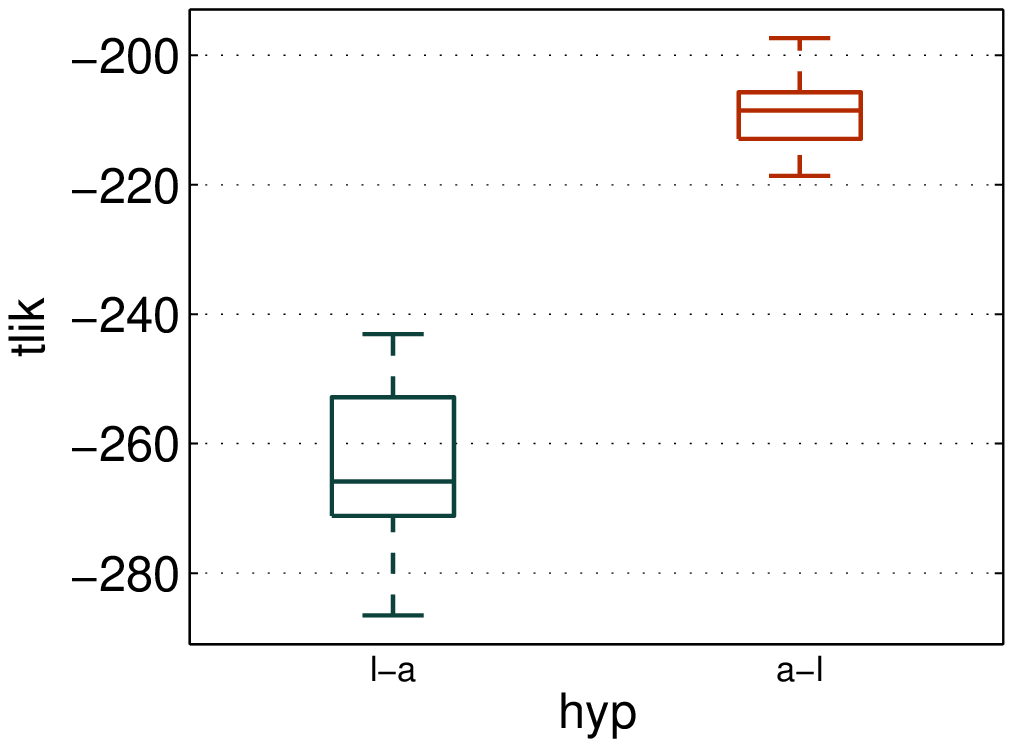}\label{fg:abalone_btlik}}
		\subfigure[]{\includegraphics[scale = 0.35]{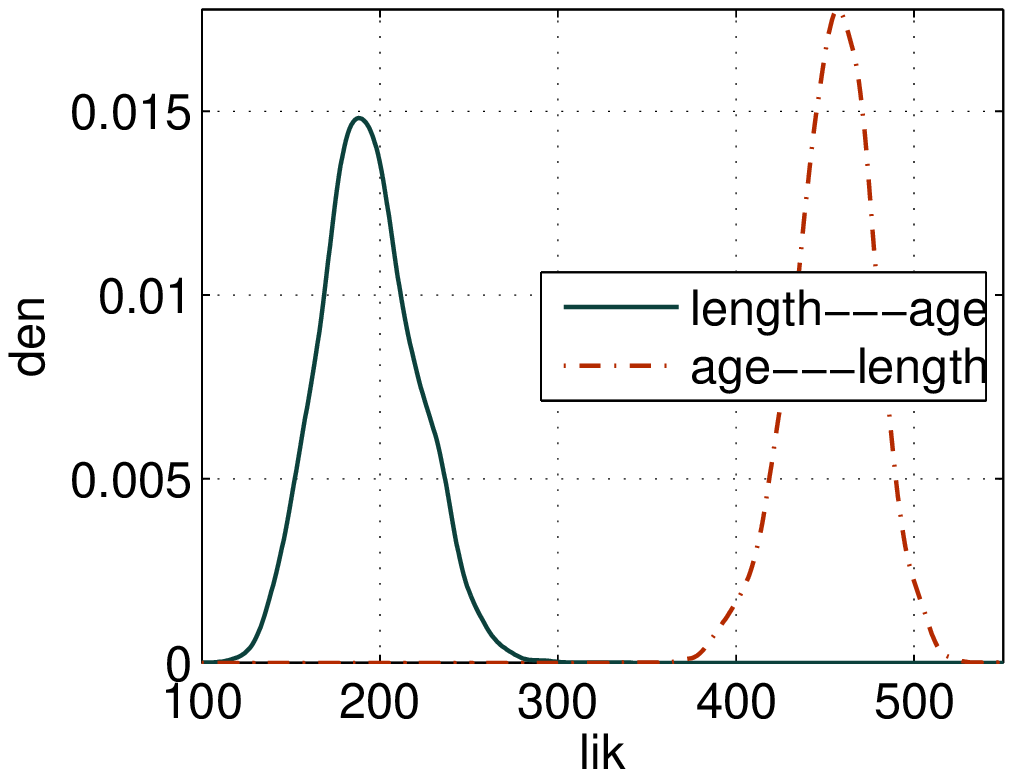}\label{fg:abalone_dlik}}
		\subfigure[]{\includegraphics[scale = 0.35]{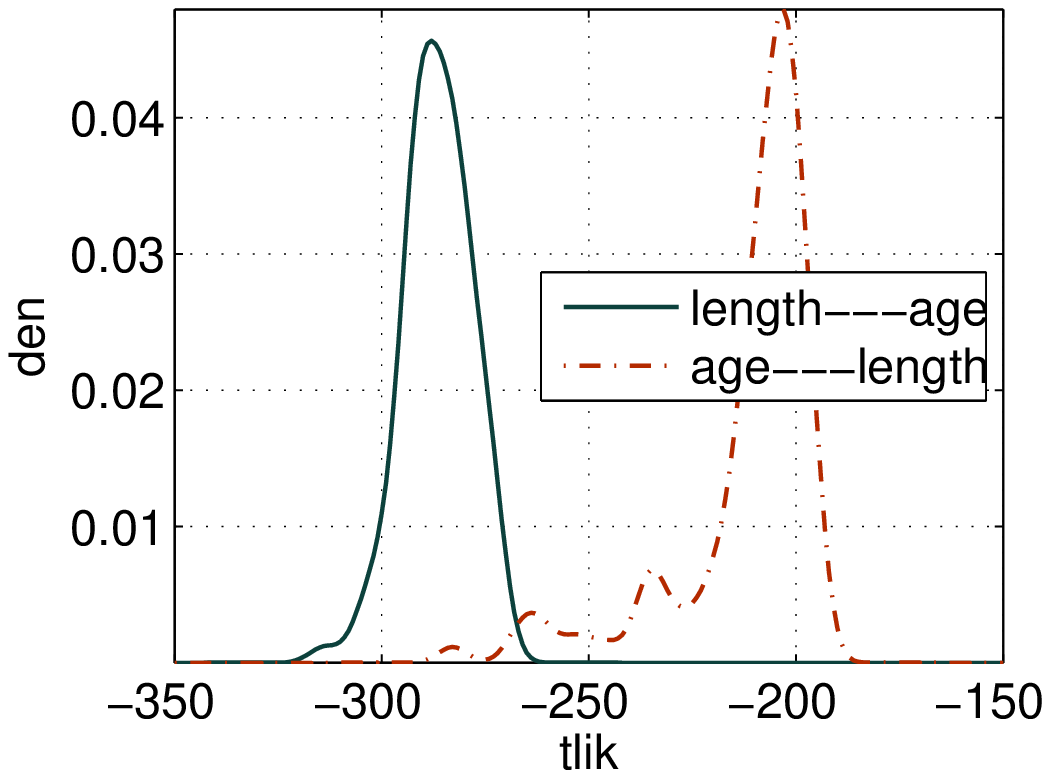}\label{fg:abalone_dtlik}}
	\end{psfrags}
	\caption{Testing $\{$length, age$\}$ in Abalone dataset. (a,b) Data and test likelihood boxplots for 10 independent repetitions. (c,d) Training and test likelihood densities for one of the repetitions. The likelihoods largely separate the two tested hypotheses.} 
\end{figure}

Abalone is one of the datasets from the UCI ML repository \citep{azzalini90}. It is targeted to predict the age of abalones from a set of physical measurements. The dataset contains 9 variables and 4177 observations. First we want to test the pair $\{$age, length$\}$. For this purpose, we use 10 subsets of $N=200$ observations to build the models and compute likelihoods just as before. Figures \ref{fg:abalone_blik} and \ref{fg:abalone_btlik} show training and test likelihoods respectively as boxplots. Both training and test likelihoods pointed to the right ordering in all 10 repetitions. In this experiment, the separation of the densities for the two hypotheses considered is very large, making age $\rightarrow$ length significantly better supported by the data. Figures \ref{fg:abalone_dlik} and \ref{fg:abalone_dtlik} show predictive densities for one of the trials indicating again that age $\rightarrow$ length is consistently preferred. We also decided to try another three sets of hypotheses: $\{$age, diameter$\}$, $\{$age, weight$\}$ and $\{$age, length, weight$\}$ for which we found the right orderings $\{10,10\}$, $\{10,10\}$ and $\{10,6\}$ out of 10 by looking at the training and the test likelihoods, respectively. In the model with three variables, increasing the number of observations used to fit the model from $N=200$ to $N=400$, increased the number of cases in which the test likelihood selected the true hypothesis from 6 to 8 times, which is more than enough to make a decision about the leading hypothesis.

To conclude this set of experiments we test SNIM against another three recently proposed methods\footnote{Matlab packages available at \url{http://webdav.tuebingen.mpg.de/causality/}.}, namely Non-linear Additive Noise (NAN) model \citep{hoyer08}, Post-Non-Linear (PNL) model \citep{zhang09} and Informational Geometric Causal Inference (IGCI) \citep{daniusis10}, using an extended version of \quotes{cause-effect pairs} task for the NIPS 2008 causality competition\footnote{Data available at \url{http://webdav.tuebingen.mpg.de/cause-effect/}.} \citep{mooji10}. The task consists on distinguishing the cause from the effect of 51 different pairs of observed variables. NAN and PNL rely on an independence test \citep[HSIC, Hilbert-Schmidt Independence Criterion,][]{gretton08} to decide which of the two variable is the cause. NAN was able to take 10 decisions all being accurate. PNL was accurate 40 times out of 42 decisions made. IGCI and SNIM obtained an accuracy of 40 and 39 pairs, respectively\footnote{Results for NAN, PNL and IGCI were taken from \citet{daniusis10} because we were unable to entirely reproduce their results with the software provided by the authors.}. The results indicate (i) that NAN and PNL are very accurate when the independence test used is able to reach a decision and (ii) in terms of accuracy, the results obtained by PNL, IGCI and SNIM are comparable. For SNIM we decide based upon the test likelihood and for IGCI we used a uniform reference measure (rescaling the data between 0 and 1). From the four tested methods we can identify two main trends. One is to explicitly model the data and decide the cause-effect direction using independence tests or test likelihoods like in NAN, PNL and SNIM. The second is to directly define a measure for directionality as in IGCI. The first option has the advantage of being able to convey more information about the data at hand whereas the second option is orders of magnitude faster than the other three because it only tests for directionality.
\subsection{Protein-signaling network}
\begin{figure}[tb]
	\centering
	\subfigure[]{\begin{tikzpicture}[ bend angle = 45, >=latex, font = \tiny ]
		\tikzstyle{obs} = [ circle, thick, draw = black!80, fill = imp2, minimum size = 1mm, inner sep = 2pt ]
		\tikzstyle{lat} = [ circle, thick, draw = black!80, fill = black!0, minimum size = 1mm, inner sep = 2pt ]
		\tikzstyle{dmy} = [ circle, thick, draw = black!0, fill = black!0, minimum size = 1mm, inner sep = 1pt]
		\tikzstyle{every label} = [black!100]
		\draw[ draw = black!40, rounded corners = 4pt ] (-2.45,2.25) rectangle (2.45,-3.45);
		\begin{scope}[ node distance = 1.0cm and 1.0cm, rounded corners = 4pt ]
			\node [obs] (pkc)  [] {pkc};
			\node [lat] (mm3) [ below of = pkc ] {m$_3$}
				edge [pre] (pkc);
			\node [lat] (mm4) [ left of = mm3 ] {m$_4$}
				edge [pre] (pkc);
			\node [obs] (pka) [ below of = mm3 ] {pka}
				edge [post] (mm4)
				edge [post] (mm3)
				edge [pre, bend left = 30] (pkc);
			\node [obs] (p38) [ right of = mm3] {p38}
				edge [pre] (mm3);
			\node [obs] (jnk) [ below of = mm4] {jnk}
				edge [pre] (mm4);
			\node [obs] (raf) [ right of = p38] {raf}
				edge [pre] (pka)
				edge [pre] (pkc);
			\node [obs] (mek) [ below of = raf] {mek}
				edge [pre] (raf)
				edge [pre] (pka);
			\node [obs] (erk) [ below of = pka] {erk}
				edge [pre] (mek)
				edge [pre] (pka);
			\node [obs] (plc) [ above of = mm4 ] {plc}
				edge [post] (pkc);
			\node [obs] (pip2) [ above of = plc] {pip$_2$}
				edge [pre] (plc)
				edge [post] (pkc);
			\node [obs] (pip3) [ left of = plc] {pip$_3$}
				edge [post] (plc)
				edge [pre and post] (pip2);
			\node [lat] (pik3) [ above of = pip3, node distance = 1.8cm ] {pik3}
				edge [post] (pip2)
				edge [post] (pip3);
			\node [obs] (akt) [ below of = pip3, node distance = 3.0cm ] {akt}
				edge [pre] (pip3)
				edge [pre] (pka)
				edge [pre] (erk);
			\node [lat] (ras) [ above of = raf, node distance = 2.8cm ] {ras}
				edge [pre] (pkc)
				edge [post] (raf)
				edge [pre] (pik3);
		\end{scope}
	\end{tikzpicture} \label{fg:tsachs}} \hspace{-2mm}
	\subfigure[$\log\langle\Lik_\DAG\rangle=-4.30e3$]{\begin{tikzpicture}[ bend angle = 45, >=latex, font = \tiny ]
		\tikzstyle{obs} = [ circle, thick, draw = black!80, fill = imp2, minimum size = 1mm, inner sep = 2pt ]
		\tikzstyle{lat} = [ circle, thick, draw = black!80, fill = black!0, minimum size = 1mm, inner sep = 2pt ]
		\tikzstyle{dmy} = [ circle, thick, draw = black!0, fill = black!0, minimum size = 1mm, inner sep = 1pt]
		\tikzstyle{every label} = [black!100]
		\draw[ draw = black!40, rounded corners = 4pt ] (-2.45,2.25) rectangle (2.45,-3.45);
		\begin{scope}[ node distance = 1.0cm and 1.0cm, rounded corners = 4pt ]
			\node [obs] (pkc)  [] {pkc};
			\node [dmy] (t_) [ below of = pkc ] {};
			\node [obs] (pka) [ below of = t_ ] {pka};
			\node [obs] (p38) [ right of = t_ ] {p38}
				edge [pre] (pkc);
			\node [obs] (jnk) [ left of = t_ ] {jnk}
				edge [pre] (pkc)
				edge [imp4, pre, densely dashed] (p38);
			\node [obs] (raf) [ right of = p38 ] {raf};
			\node [obs] (mek) [ below of = raf ] {mek}
				edge [pre] (raf);
			\node [obs] (erk) [ below of = pka ] {erk}
				edge [pre] (pka);
			\node [obs] (plc) [ above of = jnk ] {plc};
			\node [obs] (pip2) [ above of = plc ] {pip$_2$}
				edge [pre] (plc);
			\node [obs] (pip3) [ left of = plc ] {pip$_3$}
				edge [post] (plc)
				edge [post] (pip2);
			\node [obs] (akt) [ below of = pip3, node distance = 3.0cm ] {akt}
				edge [pre] (pka)
				edge [pre] (erk);
		\end{scope}
	\end{tikzpicture} \label{fg:gsachs}} \hspace{-2mm}
	\subfigure[$\log\langle\Lik_\DAG\rangle=-4.10e3$]{\begin{tikzpicture}[ bend angle = 45, >=latex, font = \tiny ]
		\tikzstyle{obs} = [ circle, thick, draw = black!80, fill = imp2, minimum size = 1mm, inner sep = 2pt ]
		\tikzstyle{lat} = [ circle, thick, draw = black!80, fill = black!0, minimum size = 1mm, inner sep = 2pt ]
		\tikzstyle{dmy} = [ circle, thick, draw = black!0, fill = black!0, minimum size = 1mm, inner sep = 1pt]
		\tikzstyle{every label} = [black!100]
		\draw[ draw = black!40, rounded corners = 4pt ] (-2.45,2.25) rectangle (2.45,-3.45);
		\begin{scope}[ node distance = 1.0cm and 1.0cm, rounded corners = 4pt ]
			\node [obs] (pkc)  [] {pkc};
			\node [dmy] (t_) [ below of = pkc ] {};
			\node [obs] (pka) [ below of = t_ ] {pka};
			\node [obs] (p38) [ right of = t_ ] {p38}
				edge [pre] (pkc);
			\node [obs] (jnk) [ left of = t_ ] {jnk}
				edge [pre] (pkc)
				edge [imp4, post, densely dashed] (p38);
			\node [obs] (raf) [ right of = p38 ] {raf};
			\node [obs] (mek) [ below of = raf ] {mek}
				edge [pre] (raf);
			\node [obs] (erk) [ below of = pka ] {erk}
				edge [pre] (pka);
			\node [obs] (plc) [ above of = jnk ] {plc};
			\node [obs] (pip2) [ above of = plc ] {pip$_2$}
				edge [pre] (plc);
			\node [obs] (pip3) [ left of = plc ] {pip$_3$}
				edge [imp3, pre, densely dotted] (plc)
				edge [pre] (pip2);
			\node [obs] (akt) [ below of = pip3, node distance = 3.0cm ] {akt}
				edge [pre] (pka)
				edge [pre] (erk);
		\end{scope}
	\end{tikzpicture} \label{fg:osachs}}
	\subfigure[$\log\langle\Lik_\DAG\rangle=-3.4e3$]{\begin{tikzpicture}[ bend angle = 45, >=latex, font = \tiny ]
		\tikzstyle{obs} = [ circle, thick, draw = black!80, fill = imp2, minimum size = 1mm, inner sep = 2pt ]
		\tikzstyle{lat} = [ circle, thick, draw = black!80, fill = black!0, minimum size = 1mm, inner sep = 2pt ]
		\tikzstyle{dmy} = [ circle, thick, draw = black!0, fill = black!0, minimum size = 1mm, inner sep = 1pt]
		\tikzstyle{every label} = [black!100]
		\draw[ draw = black!40, rounded corners = 4pt ] (-2.45,2.25) rectangle (2.45,-3.45);
		\begin{scope}[ node distance = 1.0cm and 1.0cm, rounded corners = 4pt ]
			\node [obs] (pkc)  [] {pkc};
			\node [dmy] (t_) [ below of = pkc ] {};
			\node [obs] (pka) [ below of = t_ ] {pka};
			\node [obs] (p38) [ right of = t_ ] {p38}
				edge [imp3, post, densely dotted] (pkc);
			\node [obs] (jnk) [ left of = t_ ] {jnk}
				edge [pre] (pkc)
				edge [imp4, pre, densely dashed] (p38);
			\node [obs] (raf) [ right of = p38 ] {raf};
			\node [obs] (mek) [ below of = raf ] {mek}
				edge [pre] (raf);
			\node [obs] (erk) [ below of = pka ] {erk}
				edge [pre] (pka);
			\node [obs] (plc) [ above of = jnk ] {plc};
			\node [obs] (pip3) [ above of = plc ] {pip$_3$}
				edge [imp3, pre, densely dotted] (plc);
			\node [obs] (pip2) [ left of = plc ] {pip$_2$}
				edge [pre] (plc)
				edge [post] (pip3);
			\node [obs] (akt) [ below of = pip2, node distance = 3.0cm ] {akt}
				edge [pre] (pka)
				edge [pre] (erk);
			\node [lat] (L2) [ above of = pip2, node distance = 1.8cm ] {l$_1$}
				edge [post] (pip3);
			\node [lat] (L1) [ above of = raf, node distance = 2.8cm ] {l$_2$}
				edge [post] (plc)
				edge [post] (pip3)
				edge [post] (raf)
				edge [post, bend right = 30] (mek);
		\end{scope}
	\end{tikzpicture} \label{fg:lsachs1}} \hspace{-2mm}
	\subfigure[$\log\langle\Lik_\DAG\rangle=-3.70e3$]{\begin{tikzpicture}[ bend angle = 45, >=latex, font = \tiny ]
		\tikzstyle{obs} = [ circle, thick, draw = black!80, fill = imp2, minimum size = 1mm, inner sep = 2pt ]
		\tikzstyle{lat} = [ circle, thick, draw = black!80, fill = black!0, minimum size = 1mm, inner sep = 2pt ]
		\tikzstyle{dmy} = [ circle, thick, draw = black!0, fill = black!0, minimum size = 1mm, inner sep = 1pt]
		\tikzstyle{every label} = [black!100]
		\draw[ draw = black!40, rounded corners = 4pt ] (-2.45,2.25) rectangle (2.45,-3.45);
		\begin{scope}[ node distance = 1.0cm and 1.0cm, rounded corners = 4pt ]
			\node [obs] (pkc)  [] {pkc};
			\node [dmy] (t_) [ below of = pkc ] {};
			\node [obs] (pka) [ below of = t_ ] {pka};
			\node [obs] (p38) [ right of = t_ ] {p38}
				edge [pre] (pkc);
			\node [obs] (jnk) [ left of = t_ ] {jnk}
				edge [pre] (pkc)
				edge [imp4, post, densely dashed] (p38);
			\node [obs] (raf) [ right of = p38 ] {raf};
			\node [obs] (mek) [ below of = raf ] {mek}
				edge [pre] (raf)
				edge [imp3, post, densely dotted] (pka);
			\node [obs] (erk) [ below of = pka ] {erk}
				edge [pre] (pka);
			\node [obs] (plc) [ above of = jnk ] {plc};
			\node [obs] (pip3) [ above of = plc ] {pip$_3$};
			\node [obs] (pip2) [ left of = plc ] {pip$_2$}
				edge [pre] (plc)
				edge [post] (pip3);
			\node [obs] (akt) [ below of = pip2, node distance = 3.0cm ] {akt}
				edge [pre] (pka)
				edge [pre] (erk);
			\node [lat] (L2) [ above of = pip2, node distance = 1.8cm ] {l$_1$}
				edge [post] (pip3)
				edge [post] (pip2)
				edge [post, bend right = 20] (plc);
			\node [lat] (L1) [ above of = raf, node distance = 2.8cm ] {l$_2$}
				edge [post] (pip3)
				edge [post] (raf)
				edge [post, bend right = 30] (mek);
		\end{scope}
	\end{tikzpicture} \label{fg:lsachs2}}
	\caption{Result for protein-signaling network data. (a) Textbook signaling network as reported in \citet{sachs05}. Estimated structure using SLIM: (b) using the true ordering, (c) obtaining the ordering from the stochastic search, (d) top DAG with 2 latent variables and (e) the runner-up (in test likelihood). False positives are shown in red dashed lines and reversed links in green dotted lines. Below each structure we also report the median test likelihood (larger is better).} 
\end{figure}
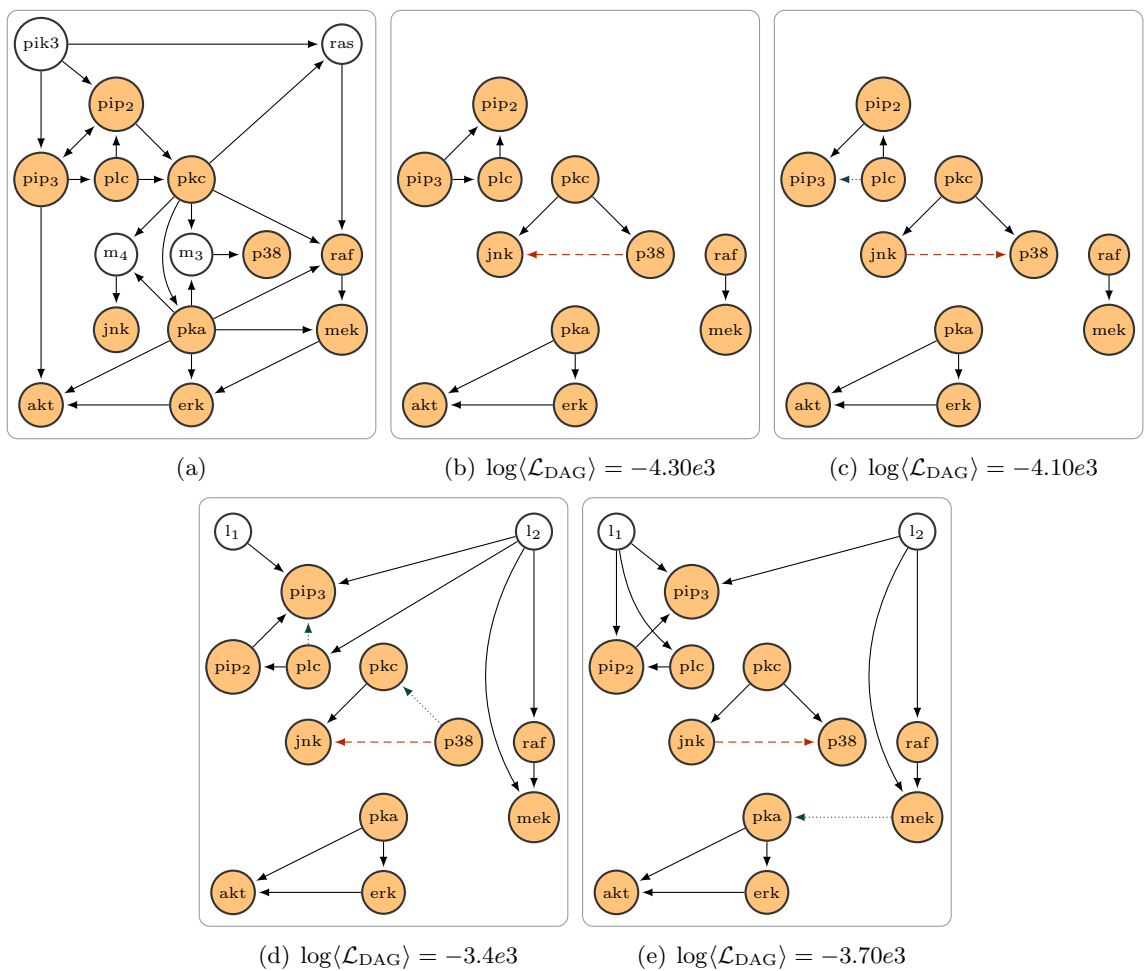

\begin{figure}[tb]
	\begin{psfrags}
	 \psfrag{raf}[r][c][0.5][0]{raf}\psfrag{erk}[r][c][0.5][0]{erk}\psfrag{p38}[r][c][0.5][0]{p38}\psfrag{jnk}[r][c][0.5][0]{jnk}\psfrag{akt}[r][c][0.5][0]{akt}\psfrag{mek}[r][c][0.5][0]{mek}\psfrag{pka}[r][c][0.5][0]{pka}\psfrag{pkc}[r][c][0.5][0]{pkc}\psfrag{pip2}[r][c][0.5][0]{pip$_2$}\psfrag{pip3}[r][c][0.5][0]{pip$_3$}\psfrag{plcy}[r][c][0.5][0]{plc}\psfrag{1.00}[r][c][1][0]{}	 
	 \psfrag{freq}[b][c][0.6][0]{Frequency ($\%$)}\psfrag{cand}[t][c][0.6][0]{Candidates}\psfrag{lik}[t][c][0.6][0]{Test log-likelihood}\psfrag{den}[b][c][0.6][0]{Density}\psfrag{lr}[b][l][0.6][0]{Test log-likelihood}\psfrag{acc}[t][l][0.6][0]{Errors}\psfrag{xx}[t][l][0.6][0]{Magnitude}\psfrag{fac}[t][l][0.6][0]{ $\log\langle\Lik_\FM\rangle=-3.46e3$}\psfrag{FM}[l][l][0.33][0]{FM}\psfrag{DAG}[l][l][0.33][0]{DAG}\psfrag{Candidates}[c][c][0.24][0]{Candidates}\psfrag{var}[t][l][0.6][0]{}\psfrag{14.5}[t][l][0.6][0]{}\psfrag{15.5}[t][l][0.6][0]{}\psfrag{16.5}[t][l][0.6][0]{}\psfrag{x}[c][c][0.6][0]{$x$}\psfrag{y}[c][c][0.6][0]{$y$}\psfrag{pip3_}[l][l][0.45][0]{pip3}\psfrag{plc_}[l][l][0.45][0]{pkc}
	\subfigure[]{\includegraphics[scale=0.27]{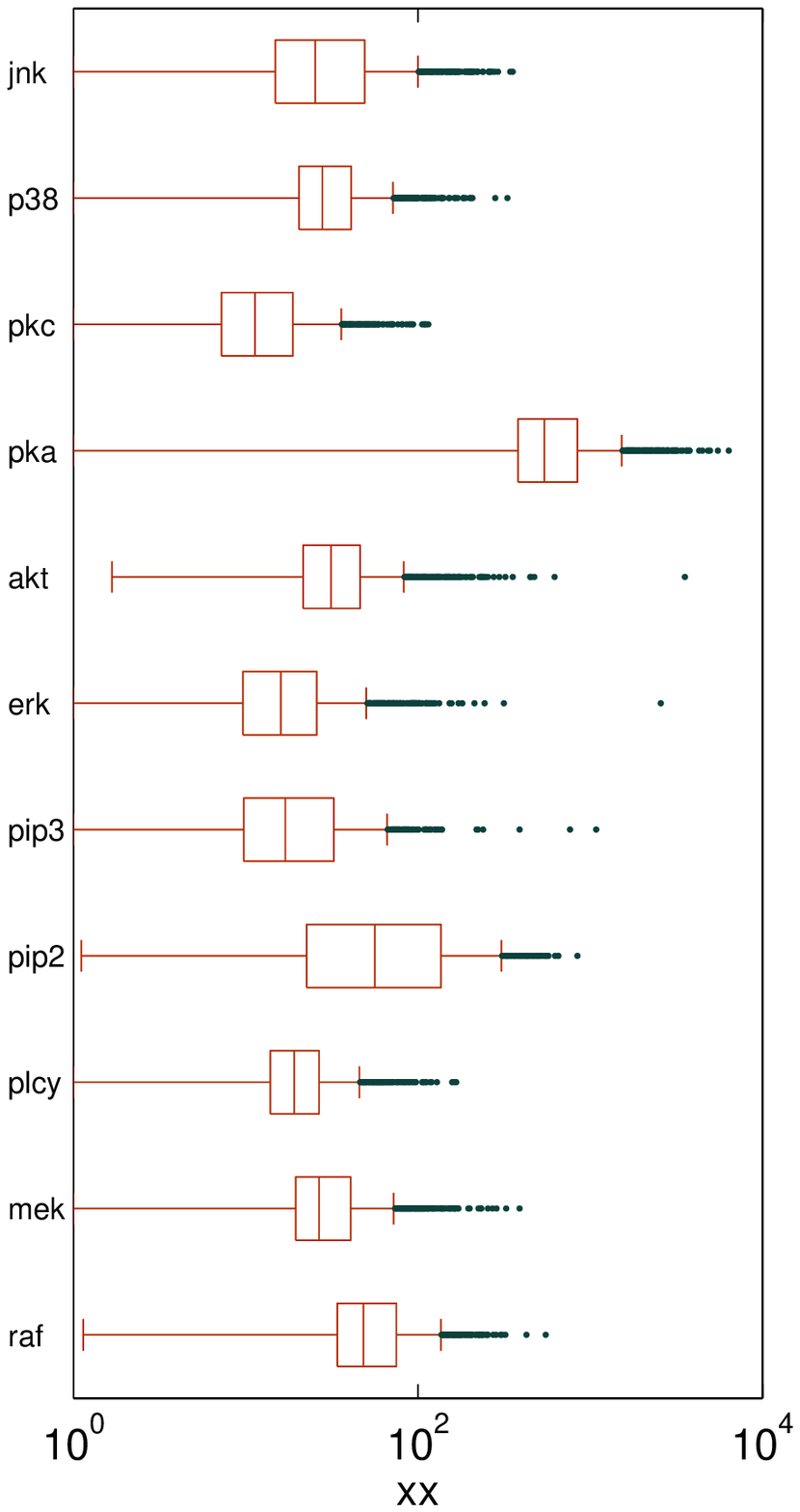}\label{fg:sachsboxplot}}
	\subfigure[]{\includegraphics[scale=0.27]{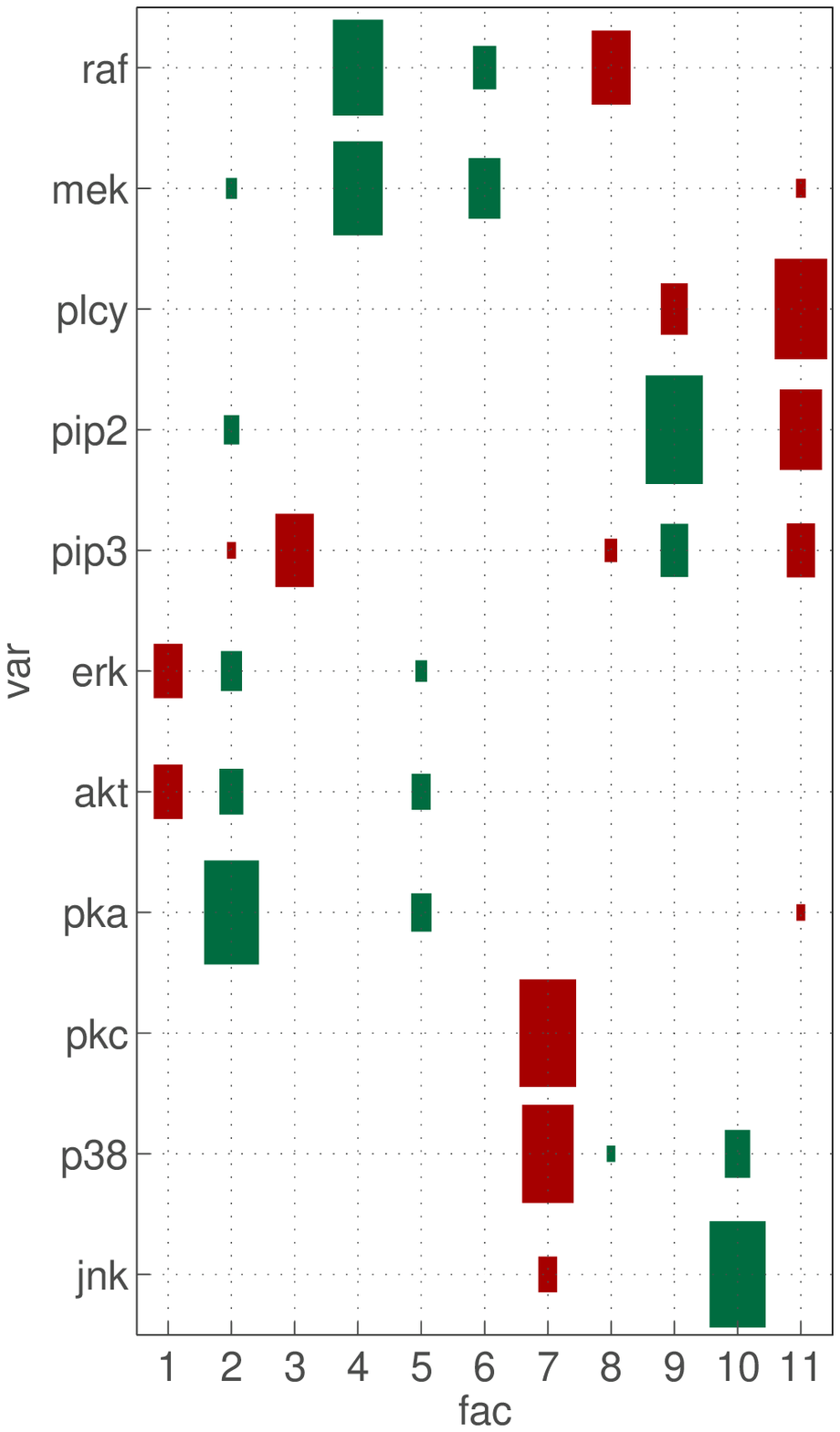}\label{fg:sachsfacs}}\hspace{1mm}
	\subfigure[]{\includegraphics[scale=0.27]{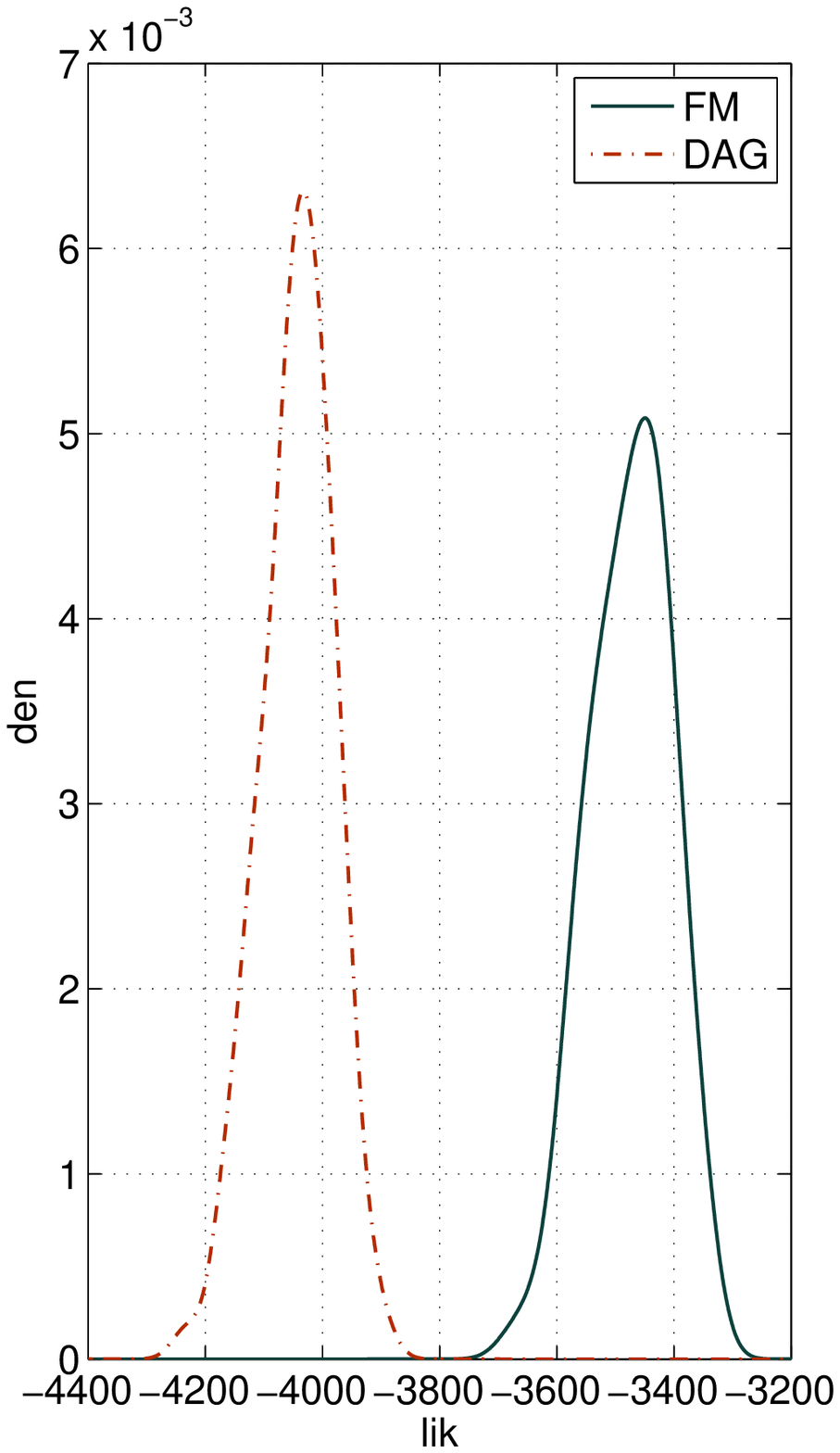}\label{fg:lik_sachs}}\hspace{2mm}
	\subfigure[]{\includegraphics[scale=0.27]{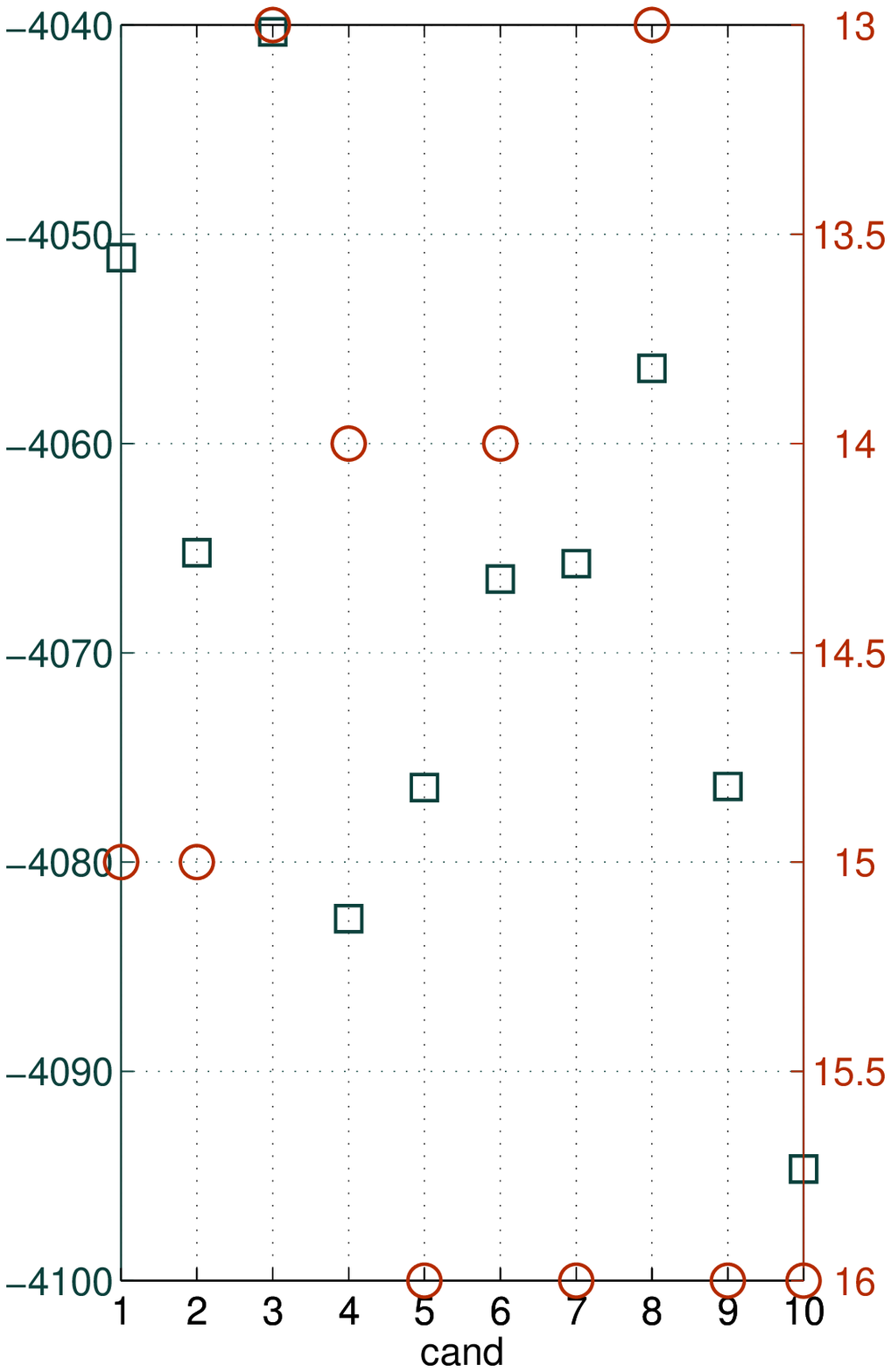}\label{fg:lr_sachs}}\hspace{2mm}
	\subfigure[]{\includegraphics[scale=0.27]{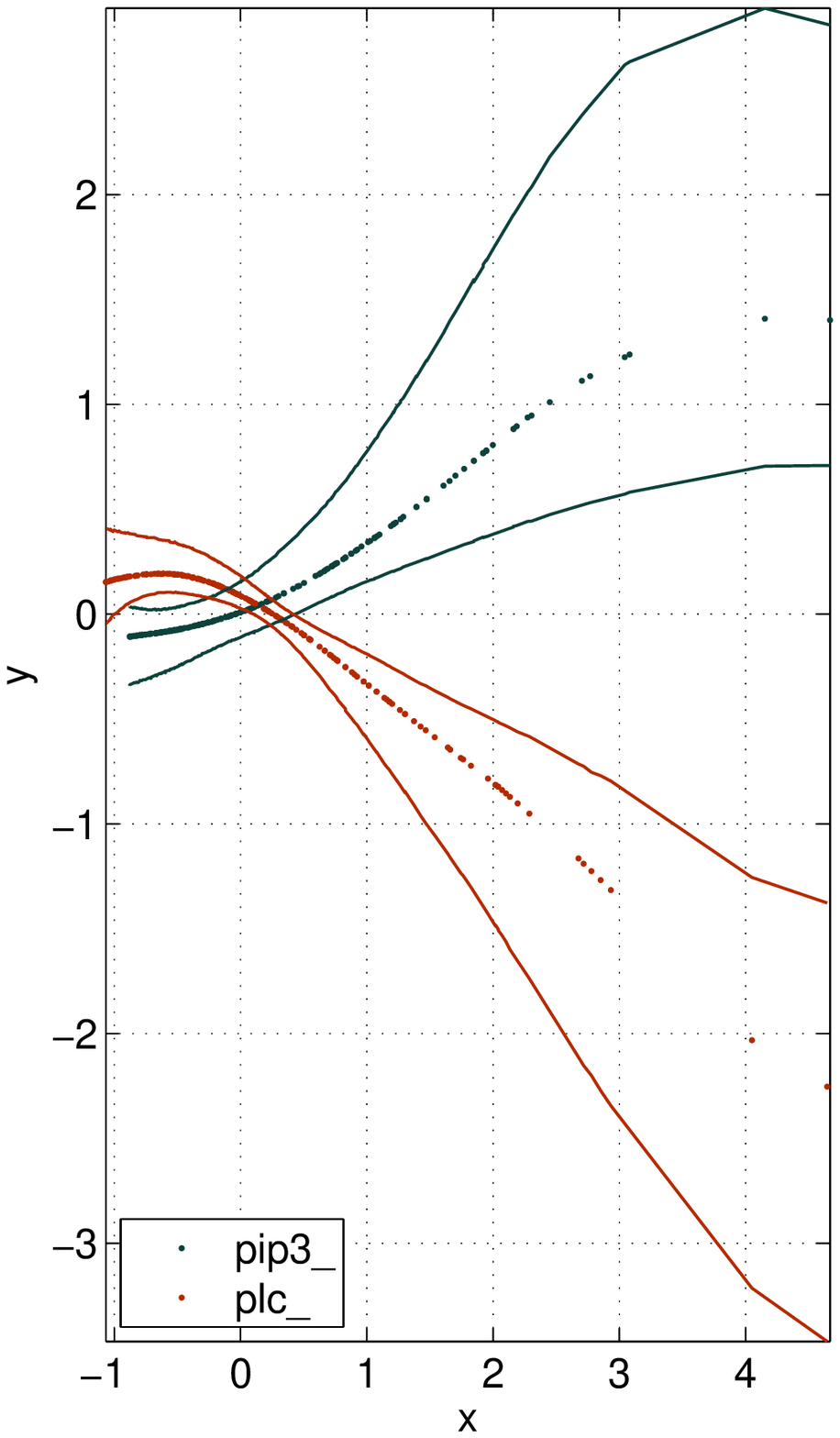}\label{fg:sachsnlsig}}
	\end{psfrags}
	\caption{Results for protein-signaling network data. (a) Boxplot for each one of the 11 variables in the dataset. (b) Estimated factor model. (c) Test likelihoods for the best DAG (dashed) and the factor model (solid). (d) Test likelihoods (squares) and structure errors (circles) included reversed links for all candidates. (e) Non-linear variables $y$ obtained as a function of the observed variables $x$ for pip3 and pkc. Each dot in the plot is an observation and the solid lines are 95$\%$ credible intervals.} 
\end{figure}

This experiment demonstrates a typical application of SLIM in a realistic biological large $N$, small $d$ setting. The dataset introduced by \citet{sachs05} consists of flow cytometry measurements of 11 phosphorylated proteins and phospholipids (raf, erk, p38, jnk, akt, mek, pka, pkc, pip$_2$, pip$_3$, plc). Each observation is a vector of quantitative amounts measured from single cells. Data was generated from a series of stimulatory cues and inhibitory interventions. Hence the data is composed of three kinds of perturbations: general activators, specific activators and specific inhibitors. Here we are only using the 1755 observations --- clearly non-Gaussian, \eg see Figure \ref{fg:sachsboxplot}, corresponding to general stimulatory conditions. It is clear that using the whole dataset, \ie using specific perturbations, will produce a richer model, however handling interventional data is out of the scope of this paper mainly because handling that kind of data with a factor model is not an easy task. Thus our current order search procedure is not appropriate. Focused only on the observational data, we want to test all the possibilities of our model in this dataset, namely, standard factor models, pure DAGs, DAGs with latent variables, non-linear DAGs and quantitative model comparison using test likelihoods. The textbook DAG structure taken from \citet[see Figure 2 and Table 3,][]{sachs05} is shown in Figure \ref{fg:tsachs} and the models are estimated using the true ordering and SLIM in Figures \ref{fg:gsachs} and \ref{fg:osachs}, respectively.

The DAG found using the right ordering of the variables shown in Figure \ref{fg:gsachs} turned out to be the same structure found by the discrete Bayesian network from \citet{sachs05} without using interventional data (see supplementary material, Figure 4(a)), with one important difference: the method presented by \citet{sachs05} is not able to infer the directionality of the links in the graph without interventional data, \ie their resulting graph is undirected. SLIM in Figure \ref{fg:osachs} finds a network almost equal to the one in Figure \ref{fg:gsachs} apart from one reversed link, plc $\rightarrow$ pip3. Surprisingly this was also found reversed by \citet{sachs05} using interventional data. In addition, there is just one false positive, the pair $\{$jnk, p38$\}$, even with a dedicated latent variable in the factor model mixing matrix shown in Figure \ref{fg:sachsfacs}, thus we cannot attribute such a false positive to estimation errors. A total of 211 ordering candidates were produced during the inference out of approximately $10^7$ possible and only $m_\topc=10$ of them were used in the structure search step. Note from Figure \ref{fg:lr_sachs} that the predictive densities for the DAGs correlate well with the structural accuracy, apart from candidate 8. Candidates 3 and 8 have the same number of structural errors, however candidate 8 has 3 reversed links instead of 1 as shown in Figure \ref{fg:osachs}. The predictive densities for the best candidate, third in Figure \ref{fg:lr_sachs} are shown in Figure \ref{fg:lik_sachs} and suggest that the factor model fits the data better. This makes sense considering that estimated DAG in Figure \ref{fg:osachs} is a substructure of the ground truth. We also examined the estimated factor model in Figure \ref{fg:sachsfacs} and we found that several factors could correspond respectively to three unmeasured proteins, namely pi3k in factors 9 and 11, m$_3$ (mapkkk, mek4/7) and m$_4$ (mapkkk, mek3/6) in factor 7, ras in factors 4 and 6.

We also wanted to assess the performance of our method and several others using this dataset, including LiNGAM and those mentioned in the Bayesian network repository experiment, even knowing that this dataset contains non-Gaussian data. We found that all of them have similar results in terms of true and false positive rates when comparing them to SLIM. However the number of reversed links was not in any case less than 6, which corresponds to more than $50\%$ of the true positives found in every case. This means that they are essentially able to find the skeleton in Figure \ref{fg:gsachs}.  Besides, we do not have knowledge of any other method for DAG learning using only the observational data that also provides results substantially better than the ones shown in Figure \ref{fg:osachs}. The poor performance of LiNGAM is difficult to explain but the large amount of reversed links may be due to the FastICA based deterministic ordering search procedure.

We also tried DAG models with latent variables in this dataset. The results obtained by the DAG with 2 a priori assumed latent variables are shown in Figures \ref{fg:lsachs1} and \ref{fg:lsachs2}, corresponding to the first and second DAG candidates in terms of test likelihoods. The first option is different to the pure DAG in Figure \ref{fg:osachs} only in the reversed link, p38 $\rightarrow$ pkc, but captures some of the behavior of pik3 and ras in l$_1$ and l$_2$ respectively. It is very interesting to see how, due to the link between pik3 and ras that is not possible to model with our model, the second inferred latent variable is detecting signals pointing towards pip$_2$ and plc. We also considered a second option because l$_1$ in the top model is only connected to a single variable pip$_3$ and thus could be regarded as an estimation error since it can be easily confounded with a driving signal. Comparing Figures \ref{fg:osachs} and \ref{fg:lsachs2} reveals two differences in the observed part, a false negative pip$_3$ $\rightarrow$ plc and a new true (reversed) positive mek $\rightarrow$ pka. This candidate is particularly interesting because the first latent variable captures the connectivity of pik3 while connecting itself to plc due to the lack of connectivity between pip$_3$ and plc. Moreover, the second latent variable resembles ras and the link between pik3 and ras as a link from itself to pip$_3$. In both solutions there is a connection between l$_2$ and mek that might be explained as a link through a phosphorylation of raf different to the observed one, \ie ras$_{s259}$. In terms of median test likelihoods, the model in Figure \ref{fg:lsachs1} is only marginally better than the factor model in Figure \ref{fg:sachsfacs} and in turn marginally worse than the DAG in Figure \ref{fg:lsachs2}.

For SNIM we started from the true ordering of the variables but we could not find any improvement compared to the structure in Figure \ref{fg:osachs}. In particular there are only two differences, plc $\rightarrow$ pip$_2$ and jnk $\rightarrow$ p38 are missing, meaning that at least in this case there are no false positives in the non-linear DAG. Looking at the parameters of the covariance function used, $\bupsilon$ (not shown) with acceptance rates of approximately $\approx 20\%$ and reasonable credible intervals, we can say that our model
found almost linear functions since all the parameters of the covariance functions are rather small. Figure \ref{fg:sachsnlsig} shows two particular non-linear variables learned by the model, corresponding to pip3 and plc. In each case the uncertainty of the estimation nicely increases with the magnitude of the observed variable and although the functions are fairly linear they resemble the saturation effect we can expect in this kind of biological data. From the three non-linear methods non-requiring exhaustive order search described in the previous section (DAG search, \quotes{ideal parent} and kPC), the best result we obtained was 11 structural errors, 10 true positives, 34 true negatives, 2 reversed and 6 bidirectional links for kPC vs 12, 9, 34, 1 and 0 by SLIM and 12, 8, 35, 0 and 0 by SNIM.
\subsection{Time series data} 
\begin{figure}[p]
	\centering
	\begin{psfrags} 
	\psfrag{tfa}[c][c][0.6][0]{}\psfrag{time}[t][c][0.6][0]{Time (Minutes)}\psfrag{ArcA}[c][c][0.6][0]{ArcA}\psfrag{CRP}[c][c][0.6][0]{CRP}\psfrag{CysB}[c][c][0.6][0]{CysB}\psfrag{FadR}[c][c][0.6][0]{FadR}\psfrag{FruR}[c][c][0.6][0]{FruR}\psfrag{GatR}[c][c][0.6][0]{GatR}\psfrag{IcIR}[c][c][0.6][0]{IcIR}\psfrag{LeuO}[c][c][0.6][0]{LeuO}\psfrag{Lrp}[c][c][0.6][0]{Lrp}\psfrag{NarL}[c][c][0.6][0]{NarL}\psfrag{PhoB}[c][c][0.6][0]{PhoB}\psfrag{PurR}[c][c][0.6][0]{PurR}\psfrag{RpoE}[c][c][0.6][0]{RpoE}\psfrag{RpoS}[c][c][0.6][0]{RpoS}\psfrag{TrpR}[c][c][0.6][0]{TrpR}\psfrag{TyrR}[c][c][0.6][0]{TyrR}\psfrag{var}[c][c][0.6][0]{Gene}\psfrag{lat}[c][c][0.6][0]{Transcription factor}\psfrag{lik}[c][c][0.6][0]{Log-likelihood}\psfrag{den}[c][c][0.6][0]{Density}\psfrag{late}[c][c][0.6][0]{Factor}
		\subfigure[]{\includegraphics[scale=0.47]{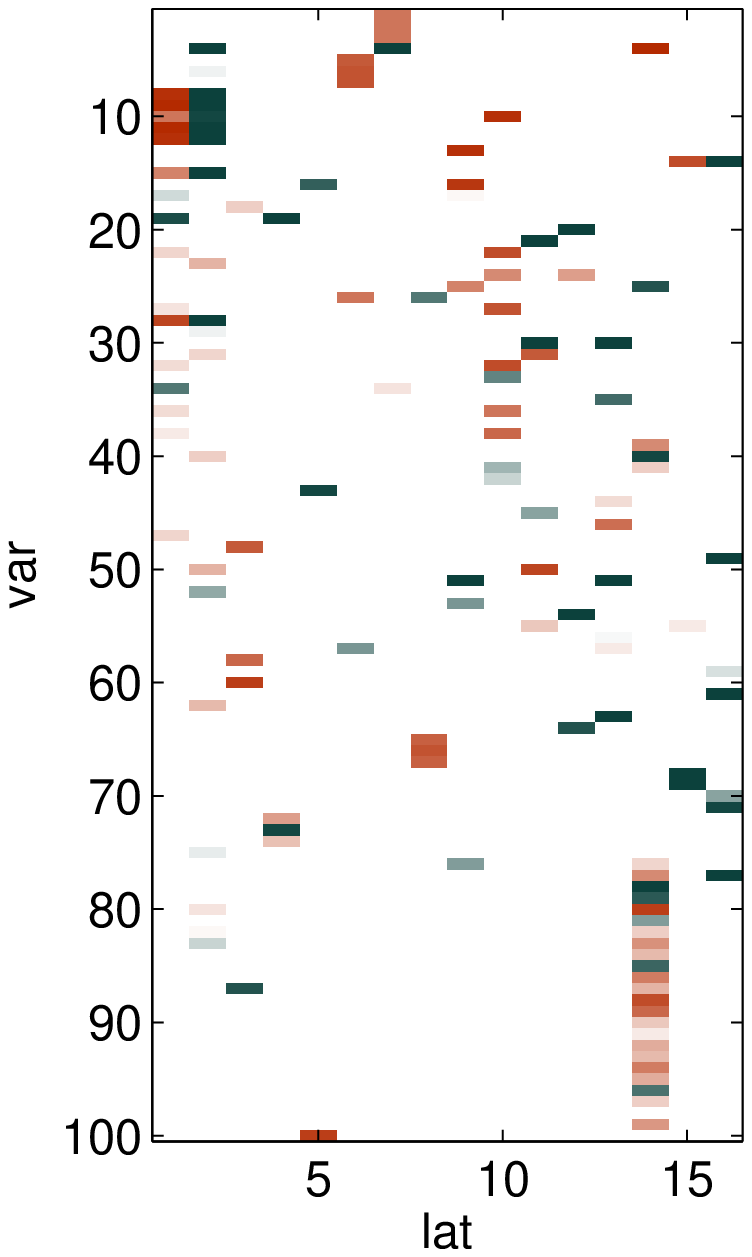}\label{fg:tecoli}}
		\subfigure[]{\includegraphics[scale=0.47]{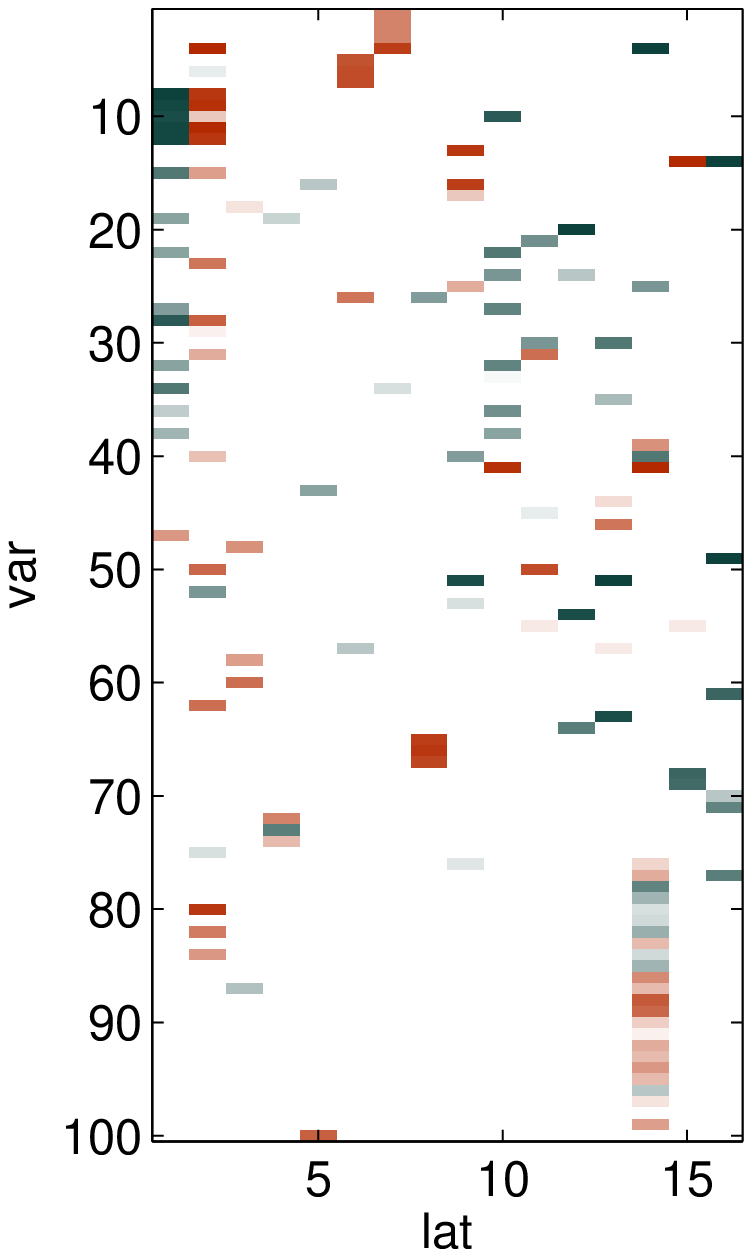}\label{fg:fRecoli}}
		\subfigure[]{\includegraphics[scale=0.47]{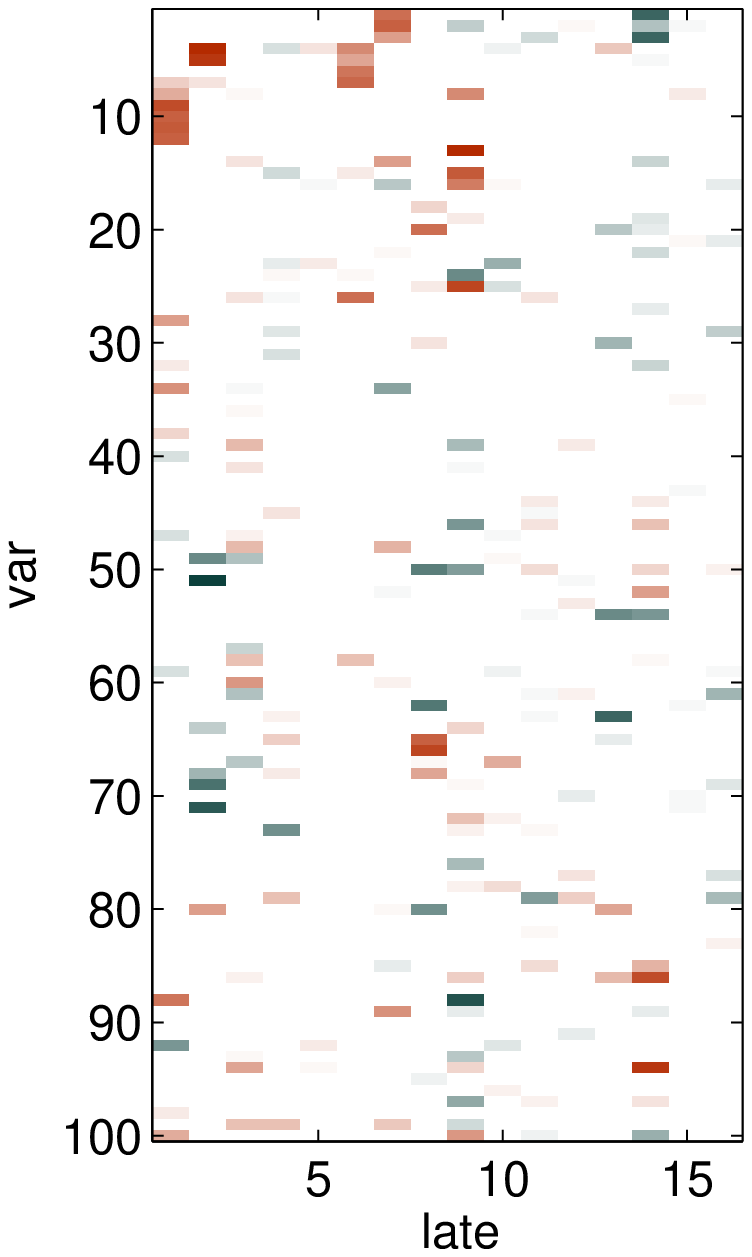}\label{fg:nfRecoli}}
		\subfigure[]{\includegraphics[scale=0.47]{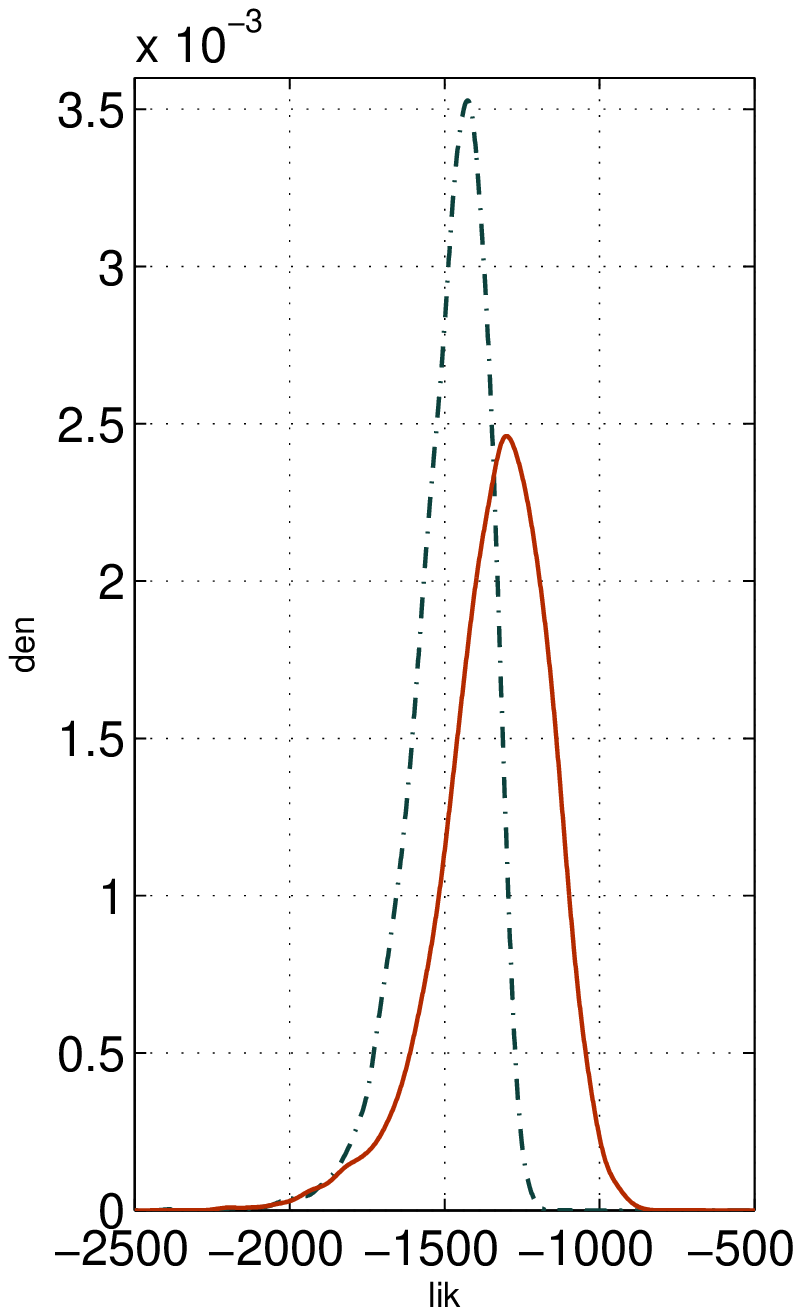}\label{fg:mcecoli}}
		\subfigure[]{\includegraphics[scale=0.54]{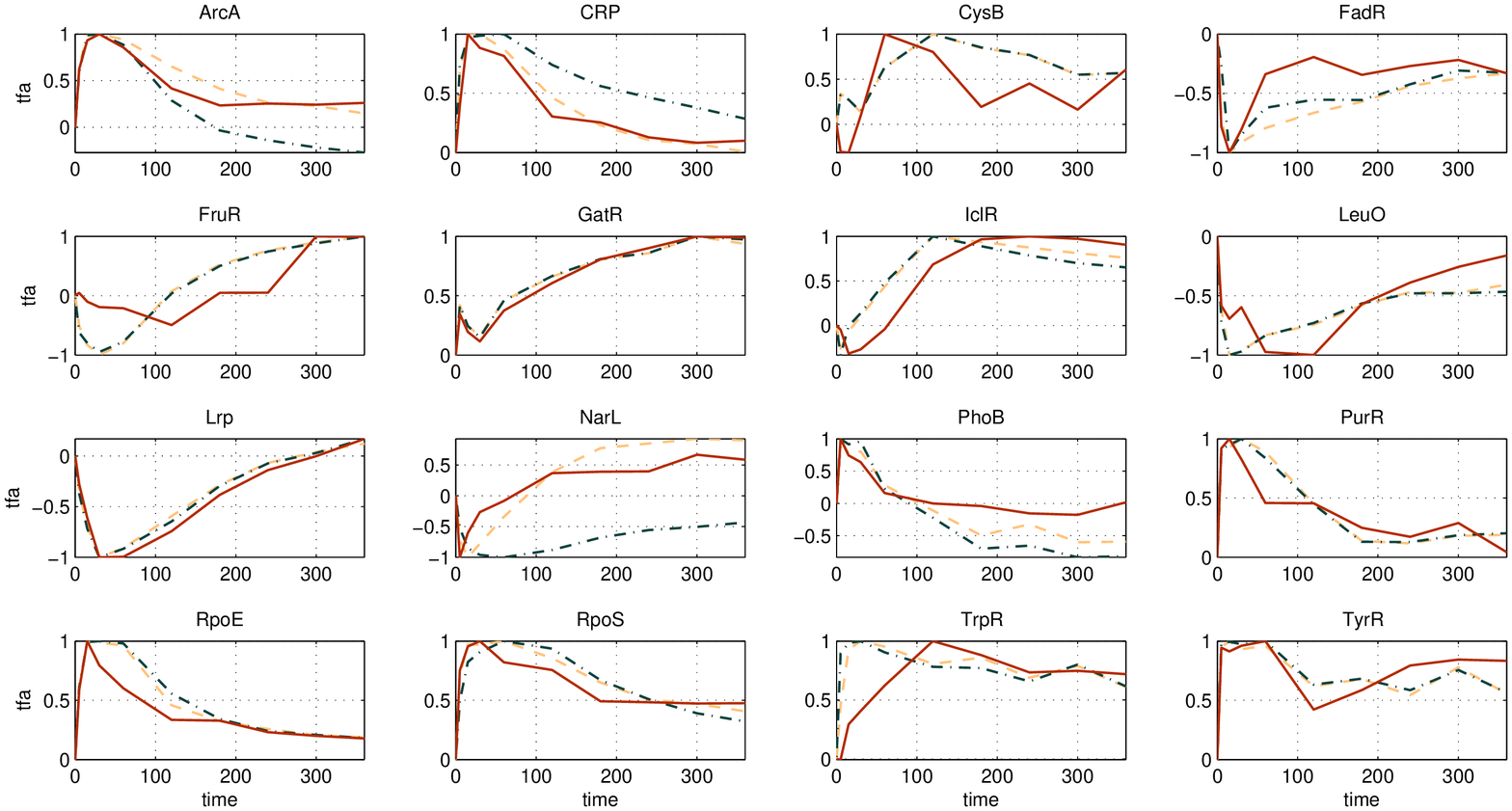}\label{fg:Zecoli}}
	\end{psfrags} \vspace{-0.3cm}
	\caption{Results for \emph{E.~coli} dataset. Mixing matrices estimated using: (a) NCA, (b) our formulation when restricting $\Q_L$ using RegulonDB information and (c) the factor model. (d) Model comparison results using test likelihoods. The restricted model (dash-dotted line) obtained a median negative log-likelihood of $1463.4$ whereas the unrestricted model (solid line) obtained $1317.1$, suggesting no significant model preferences. (e) Estimated transcription factor activities, $\Z$. Our methods (solid and dash-dotted lines for unrestricted and restricted model respectively) produce similar results to those produced by NCA (dashed line).} \label{fg:ecoli}
\end{figure}

We illustrate the use Correlated Sparse Linear Identifiable Modeling (CLSIM) on the dataset introduced by \citet{kao04} consisting of temporal gene expression profiles of \emph{E.~coli} during transition from glucose to acetate measured using DNA microarrays. Samples from 100 genes were taken at 5, 10, 15, 30, 60 minutes and every hour until 6 hours after transition\footnote{Data available at \url{http://www.seas.ucla.edu/~liaoj/NCA_module_Data}.}. The general goal is to reconstruct the unknown transcription factor activities from the expression data and some prior knowledge. In \citet{kao04} the prior knowledge consisted of taking the set of transcription factors (ArcA, CRP, CysB, FadR, FruR, GatR, IcIR, LeuO, Lrp, NarL, PhoB, PurB, RpoE, RpoS, TrpR and TyrR) controlling the observed genes and the (up-to-date) connectivity between genes and transcription factors from RegulonDB\footnote{\url{http://regulondb.ccg.unam.mx/}.} \citep{Gama-Castro08}. From this setting, we can immediately relate the transcriptions factors with $\Z$, such a connectivity with $\Q_L$, and their relative strengths with $\C_L$, hence the problem can be seen as a standard factor model. In \citet{kao04} they applied a method called Network Component Analysis (NCA), that uses a least-squares based algorithm to solve a problem similar to the one in equation \eqref{eq:PBxCz}, but assuming that the sparsity pattern (masking matrix $\Q_L$) of $\C_L$ is fixed and known. It is well-known that the information in RegulonDB is still incomplete and hard to obtain for organisms different than \emph{E.~coli}. Our goal here is thus to obtain similar transcription factor activities to those found by \citet{kao04} without using the information from RegulonDB, but taking into account that the data at hand is a time series by letting each transcription factor activity have an independent Gaussian process prior as described for CSLIM in Section \ref{sc:cslim}. We will not attempt to use $\Q_L$ to recover the ground truth connectivity information since RegulonDB is collected from a wide range of experimental conditions and not only from the transcriptional activity produced by the \emph{E.~coli} during its transition from glucose to acetate. The results are shown in Figure \ref{fg:ecoli}.

Results in Figure \ref{fg:Zecoli} show the source matrix $\Z$ recovered by our model together with those from NCA\footnote{Matlab package (v.2.3) available at \url{http://www.seas.ucla.edu/~liaoj/download.htm}.}. In this experiment we ran a single chain and collected 6000 samples after a burn-in period of 2000 samples (approximately 10 minutes in a desktop machine). Most of the profiles obtained by our method are similar to those obtained by NCA \citep{kao04}. We ran two versions of our model, one with $\Q_L$ fixed to the RegulonDB values, \ie similar in spirit to NCA, and another when we infer $\Q_L$ without any restriction. The results of NCA and our model with fixed $\Q_L$ are directly comparable (up to scaling) whereas we had to match the permutation $\PC$ of the unrestricted model to those found by NCA in order to compare, using the Hungarian algorithm. Figure \ref{fg:tecoli} shows the mixing matrices obtained by NCA and our two models. Figures \ref{fg:tecoli} and \ref{fg:fRecoli} are very similar due to the restriction imposed on $\Q_L$. The mixing matrix obtained by our unrestricted model in Figure \ref{fg:nfRecoli} is clearly denser than the other two, suggesting that there are different ways of connecting genes and transcription factors and still reconstruct the transcription factor activities given the observed gene expression data. When looking to the test log-likelihood densities obtained by our two models in Figure \ref{fg:mcecoli} they are very similar, which suggests that there is no evidence that one of the models makes a better fit on test data. In terms of Mean Squared Error (MSE), NCA obtained $0.0146$ while our model reached $0.0264$ and $0.0218$ on the restricted and unrestricted models, respectively, when using $90\%$ of the data for inference. In addition, the $95\%$ credible intervals for the MSE were $(0.0231,0.0329)$ and $(0.0164,0.0309)$ respectively. The latter shows again that there is no evidence that one of the three models is better than the other two, considering that: (i) NCA is trained on the entire dataset and (ii) our unrestricted model could, in principle, produce mixing matrices arbitrarily denser than the connectivity matrix extracted from RegulonDB, and thus, again in principle, lower MSE values.
\section{Discussion} \label{sc:dis}
We have proposed a novel approach called SLIM (Sparse Linear Identifiable Multivariate modeling) to perform inference and model comparison of general linear Bayesian networks within the same framework. The key ingredients for our Bayesian models are slab and spike priors to promote sparsity, heavy-tailed priors to ensure identifiability and predictive densities (test likelihoods) to perform the comparison. A set of candidate orderings is produced by stochastic search during the factor model inference. Subsequently, a linear DAG with or without latent variables is learned for each of the candidates. To the authors' knowledge this is the first time that a method for comparing such closely related linear models has been proposed. This setting can be very beneficial in situations where the prior evidence suggests both DAG structure and/or unmeasured variables in the data. We also show that the DAG with latent variables can be fully identifiable and that SLIM can be extended to the non-linear case (SNIM - Sparse Non-linear Identifiable Multivariate modeling), if the ordering of the variables is provided or can be tested by exhaustive enumeration. For example in the protein-signaling network \citep{sachs05}, the textbook ground truth suggests both DAG structure and a number of unmeasured proteins. The previous approach \citep{sachs05} only performed structure learning in pure DAGs but our results using observational data alone suggest that the data is better explained by a (possibly non-linear) DAG with latent variables. Our extensive results on artificial data showed one by one the features of our model in each one of its variants, and demonstrated empirically their usefulness and potential applicability. When comparing against LiNGAM, our method always performed at least as well in every case with a comparable computational cost. The presented Bayesian framework also allows easy extension of our model to match different prior beliefs about the problems at hand without significantly changing the model and its conceptual foundations, as in CSLIM and SNIM.

We believe that the priors that give raise to sparse models in the fully Bayesian inference setting, like the two-level slab (continuous) and spike (point-mass in zero) priors used are very powerful tools for simultaneous model and parameter inference. They may be useful in many settings in machine learning where sparsity of parameters is desirable. Although the posterior distributions for slab and spike priors will be non-convex, it is our experience that inference with blocked Gibbs sampling actually has very good convergence properties. In the two-level approach, one uses a hierarchy of two slab and spike priors. The first is on the parameter and the second is on the mixture parameter (i.e.\ the probability that the parameter is non-zero). Instead of letting this parameter be controlled by a single Beta-distribution (one level approach) we have a slab and spike distribution on it with a Beta-distributed slab component biased towards one. This makes the model more parsimonious, i.e.\ the probability that parameters are zero or non-zero is closer to zero and one and parameter settings are more robust.

In the following we will discuss open questions and future directions. From the Bayesian network repository experiment it is clear that we need to improve our ordering search procedure if we want to use SLIM for problems with more than say 50 variables. This basically amounts to finding proposal distributions that better exploit the particularities of the model at hand. Another option could be to provide the proposal distribution with some notion of memory to avoid permutations with low probability and/or expand the coverage of the searching procedure.

It is well studied in the literature on sparse models that for increasing number of observations any model tends to loose its sparsity capabilities. This is because the likelihood starts dominating the inference, making the prior distribution less informative. The easiest way to handle such an effect is to make the hyperparameters of the sparsity prior dependent on $N$. We have not explored this phenomenon in SLIM but it should certainly be taken into account in the specification of sparsity priors.

Directly specifying the distributions of the latent variables in order to obtain identifiability in the general DAG with latent variables requires having different distributions for the driving signals of the observed variables and latent variables. This may introduce model mismatch or be restrictive in some cases as one will not have this kind of knowledge a priori. We thus need more principled ways to specify distributions for $\z$ ensuring identifiably, without restricting some of its components to having a particular behavior, like having heavier tails than the driving signals for instance. We conjecture that providing $\z$ with a parameterization of Dirichlet process priors with appropriate base measures would be enough but we are not certain whether this would be sufficient in practice.

We set a priori that the components of $\z$ are independent. Although this is a very reasonable assumption, it does not allow for connectivity between latent variables as we see for example in the protein signaling network, see Figure \ref{fg:tsachs}. It is straight forward to specify such a model, although identifiability becomes even harder to ensure in this case.

We do not have an ordering search procedure for the non-linear version of SLIM. This is a necessary step since exhaustive enumeration of all possible orderings is not an option beyond say 10 variables. The main problem is that the non-linear DAG has no equivalent factor model representation so we cannot directly exploit the permutation candidates we find in SLIM. However, as long as the non-linearities are weak, one might in principle use the permutation candidates found in a factor model, \ie the linear effects will determine the correct ordering of the variables.
	
SLIM cannot handle experimental (interventional) data, and consequently around 80\% of the data from the \citet{sachs05} study is not used. It is well-established how to learn with interventions in DAGs \citep[see][]{sachs05}. The problem remains of how to formulate effective inference with interventional data in the factor model.
\section*{Acknowledgments}
We thank the editor and the three anonymous referees for their helpful comments and discussions that improved the presentation of this paper.
\appendix
\section{Gibbs sampling} \label{ap:inf}
Given a set of $N$ observations in $d$ dimensions, the data $\X=[\x_1,\dots,\x_N]$ and $m$ latent variables, MCMC analysis is standard and can be implemented through Gibbs sampling. Note that in the following, $\X_{i:}$ and $\X_{:i}$ are rows and columns of $\X$, respectively, and $i$, $j$, $n$ are indexes for dimensions, factors and observations, respectively. In the following we describe the conditional distributions needed to sample from the standard factor model hierarchy. Below we will briefly discus the modifications needed for the DAG.
\paragraph{Noise variance} We can sample each element of $\bPsi$ independently using
\begin{align} \label{eq:gibbsPsi}
\psi_i\inv|\X_{i:},\C_{i:},\Z,\V_i,s_s,s_r \ \sim & \ \Ga\left(\psi_i\inv\left|s_s+\frac{N+d}{2},s_r+u\right.\right) \ ,
\end{align}
where $\V_i$ is a diagonal matrix with entries $\tau_{ij}$ and
\begin{align*}
	u = \frac{1}{2}(\X_{i:}-\C_{i:}\Z)(\X_{i:}-\C_{i:}\Z)\ts+\frac{1}{2}\C_{i:}\V_i\inv\C_{i:}\ts \ .
\end{align*}
\paragraph{Factors} The conditional distribution of the latent variables $\Z$ using the scale mixtures of Gaussians representation can be computed independently for each element of $z_{jn}$ using
\begin{align} \label{eq:gibbsZ}
z_{jn}|\X_{:n},\C_{:j},\Z_{:n},\bPsi,\upsilon_{jn} \ \sim \ \DN(z_{jn}|\C_{:j}\ts\bPsi\inv\bepsilon_{\backslash jn},u_{jn}) \ ,
\end{align}
where $u_{jn} = (\C_{:j}\ts\bPsi\inv\C_{:j}+\upsilon_{jn}\inv)\inv$ and $\bepsilon_{\backslash jn}=\X_{:n}-\C\Z_{:n}|_{z_{jn}=0}$. If the latent factors are Laplace distributed the mixing variances $\upsilon_{jn}$ have exponential distribution, thus the resulting conditional is
\begin{align*}
\upsilon_{jn}\inv|z_{jn},\lambda \ \sim & \ \IG\left(\upsilon_{jn}\inv\left|\frac{\lambda}{|z_{jn}|},\lambda^2\right.\right) \ ,
\end{align*}
and for the Student's \emph{t}, with corresponding gamma densities as
\begin{align*}
\upsilon_{jn}\inv|z_{jn},\sigma^2,\theta \ \sim & \ \Ga\left(\upsilon_{jn}\inv\left|\frac{\theta+1}{2},\frac{\theta}{2}+\frac{z_{jn}^2}{2\sigma^2}\right.\right) \ ,
\end{align*}
where $\IG(\cdot|\mu,\lambda)$ is the inverse Gaussian distribution with mean $\mu$ and scale parameter $\lambda$ \citep{chhikara89}.
\paragraph{Gaussian processes} In practice, the prior distribution for each row of the matrix $\Z$ in CSLIM has the form $z_{j1},\ldots,z_{jN} \sim \DN(0,\K_j)$, where $\K_j$ is a covariance matrix of size $N\times N$ built using $k_{\upsilon_j,n}(n,n')$. The conditional distribution for $z_{j1},\ldots,z_{jN}$ can be computed using
\begin{align*}
z_{j1},\ldots,z_{jN}|\X,\C_{j:},\Z_{\backslash j},\bPsi \ \sim \ \DN(z_{j1},\ldots,z_{jN}|\C_{:j}\ts\bPsi\inv\bepsilon_{\backslash j}\V ,\V) \ ,
\end{align*}
where $\Z_{\backslash j}$ is $\Z$ without row $j$, $\V = (\U+\K_j\inv)\inv$, $\U$ is a diagonal matrix with elements $\C_{:j}\ts\bPsi\inv\C_{:j}$ and $\bepsilon_{\backslash j}=\X-\C\Z|_{z_{j1},\ldots,z_{jN}=0}$. The computation of $\V$ can be done in a numerically stable way by rewriting $\V = \K_j - \K_j (\U\inv + \K_j )\inv \K_j$ and then using Cholesky decomposition and back substitution to obtain in turn $\L\L\ts = \U\inv + \K_j$ and $\L\inv\K_j$. The hyperparameters of the covariance function in equation \eqref{eq:GPhyp} can be sampled using
\begin{align*}
	\kappa|\bupsilon,k_s,k_r \ \sim \ \Ga\left(\kappa\middle|k_s+mu_s,k_r+\sum_{j=1}^m \upsilon_j\right) \ .
\end{align*}
For the inverse length-scales we use Metropolis-Hastings updates with proposal $q(\upsilon_j^\star|\upsilon_j)=p(\upsilon_j^\star)$ and acceptance ratio
\begin{align*} 
	 \xi_{\rightarrow\star}=\frac{\DN(z_{j1},\ldots,z_{jN}|\0,\K_j^\star)}
                                 {\DN(z_{j1},\ldots,z_{jN}|\0,\K_j)} \ ,
\end{align*}
where $\K_j^\star$ is obtained using $k_{\upsilon_j^\star,n}(n,n')$. For SNIM, we only need to replace $\C$ by $\B$, $\Z$ by $\Y=[\y_1 \ \ldots \y_N]$ and $k_{\upsilon_j,n}(n,n')$ by $k_{\upsilon_i,x}(\x,\x')$.
\paragraph{Mixing matrix} In order to sample each $c_{ij}$ from the conditional distribution of the matrix $\C$ we use
\begin{align} \label{eq:gibbsD}
c_{ij}|\X_{i:},\C_{\backslash ij},\Z_{j:},\psi_i,\tau_{ij} \ \sim & \ \DN(c_{ij}|u_{ij}\bepsilon_{\backslash ij}\Z_{j:}\ts,u_{ij}\psi_i) \ ,
\end{align}
where $u_{ij} = (\Z_{j:}\Z_{j:}\ts+\tau_{ij}\inv)\inv$ and $\bepsilon_{\backslash ij}=\X_{i:}-\C_{i:}\Z|_{d_{ij}=0}$. Note that we only need to sample those $c_{ij}$ for which $r_{ij}=1$, \ie just the slab distribution. Sampling from the conditional distributions for $\tau_{ij}$ can be done using
\begin{align} \label{eq:gibbstau}
\tau_{ij}\inv|d_{jn},t_s,t_r \ \sim & \ \Ga\left(\tau_{ij}\inv\left|t_s+\frac{1}{2},t_r+\frac{d_{ij}^2}{2\psi_i}\right.\right) \ .
\end{align}
The conditional distributions for the remaining parameters in the slab and spike prior can be written first for the masking matrix $\Q$ as
\begin{align} \label{eq:gibbsq}
q_{ij}|\X_{i:},\D_{i:},\Z,\psi_i,\tau_{ij},\eta_{ij} \ \sim & \ \Ber\left(q_{ij}\middle|\frac{\xi_{\eta_{ij}}}{1+\xi_{\eta_{ij}}}\right) \ ,
\end{align}
where
\begin{align*}
\xi_{\eta_{ij}} \ = & \ \frac{\alpha_m\nu_j}{1-\alpha_m\nu_j}\frac{\psi_i^{1/2}}{(\Z_{j:}\Z_{j:}\ts+\tau_{ij}\inv)^{1/2}}\exp\left(\frac{(\bepsilon_{\backslash ij}\Z_{j:}\ts)^2}{2\psi_i(\Z_{j:}\Z_{j:}\ts+\tau_{ij}\inv)}\right) \ ,
\end{align*}
and the probability of each element of $\C$ of being non-zero as
\begin{align} \label{eq:gibbseta}
\eta_{ij}|u_{ij},q_{ij},\alpha_p,\alpha_m \ \sim & \ (1-u_{ij})\delta(\eta_{ij})+u_{ij}\Be(\eta_{ij}|\alpha_p\alpha_m+q_{ij},\alpha_p(1-\alpha_m) +1-q_{ij}) \ ,
\end{align}
where $u_{ij}\sim\Ber(h_{ij}|r_{ij}+(1-r_{ij})\nu_j(1-\alpha_m)/(1-\nu_j\alpha_m))$, \ie we set $u_{ij}=1$ if $q_{ij}=1$. Finally, for the column-wise shared sparsity rate we have
\begin{align} \label{eq:gibbsnu}
\nu_j|\u_{j},\beta_p,\beta_m \ \sim & \ \Be\left(\nu_j\middle|\beta_p\beta_m+\sum_{i=1}^d u_{ij},\beta_p(1-\beta_m)+\sum_{i=1}^d (1-u_{ij})\right) \ .
\end{align}
Sampling from the DAG model only requires minor changes in notation but the conditional posteriors are essentially the same. The changes mostly amount to replacing accordingly $\C$ by $\B$ and $\Q$ by $\R$. Note that $\Q_L$ is the identity and $\R$ is strictly lower triangular a priori, thus we only need to sample their active elements.
\paragraph{Inference with missing values} We introduce a binary masking matrix indicating whether an element of $\X$ is missing or not. For the factor model we have the following modified likelihood
\begin{align*}
	p(\X_{\rm tr}|\C,\Z,\bPsi,\M_{\rm miss}) = \DN(\M_{\rm miss}\odot\X|\M_{\rm miss}\odot(\C\Z),\bPsi) \ .
\end{align*}
Testing on the missing values, $\M_{\rm miss}^\star=\1\1\ts-\M$ requires averaging the test likelihood
\begin{align*}
	p(\X^\star|\C,\Z,\bPsi,\M_{\rm miss}^\star) = \DN(\M_{\rm miss}^\star\odot\X|\M_{\rm miss}^\star\odot(\C\Z),\bPsi) \ ,
\end{align*}
over $\C,\Z,\bPsi$ given $\X_{\rm tr}$ (training). We can approximate the predictive density $p(\X^\star|\X_{\rm tr},\cdot)$ by computing the likelihood above during sampling using the conditional posteriors of $\C$, $\Z$ and $\bPsi$ and then summarizing using for example the median. Drawing from $\C$, $\Z$, $\bPsi$ can be achieved by sampling from their respective conditional distributions as described before with some minor modifications.
\bibliography{../bib_files/mlbib}
%
\end{document}